\pdfoutput=1
\documentclass{article}

\PassOptionsToPackage{numbers,sort&compress}{natbib}
\usepackage[eandd, final]{neurips_2026}

\usepackage[utf8]{inputenc}
\usepackage[T1]{fontenc}
\usepackage{hyperref}
\usepackage{url}
\usepackage{booktabs}
\usepackage{amsfonts}
\usepackage{amsmath}
\usepackage{graphicx}
\usepackage{multirow}
\usepackage{xcolor}
\usepackage{colortbl}
\usepackage{subcaption}
\usepackage{tablefootnote}
\usepackage{array}
\usepackage{enumitem}
\usepackage{makecell}
\usepackage{tabularx}

\hypersetup{colorlinks=true, linkcolor=blue, citecolor=blue, urlcolor=blue, pdftitle={PULSE: A Synchronized Five-Modality Dataset for Sensorimotor Coordination in Long-Horizon Daily Activities}, pdfauthor={Wolin Liang, Yang Gao, Peiyu Yan, Yuexiang Hu, Yingjing Xiao, Xiongfeng Ying, Yingnian Guo, Cheng Zhang, Zhanpeng Jin}}

\title{PULSE: A Synchronized Five-Modality Dataset \\ for Sensorimotor Coordination in Long-Horizon Daily Activities}

\author{%
  Wolin Liang$^{1}$\quad
  Yang Gao$^{1}$\thanks{Corresponding author: \texttt{gaoyang2025@scut.edu.cn}}\quad
  Peiyu Yan$^{1}$\quad
  Yuexiang Hu$^{1}$\quad
  Yingjing Xiao$^{2}$%
  \AND
  Xiongfeng Ying$^{1}$\quad
  Yingnian Guo$^{1}$\quad
  Cheng Zhang$^{1}$\quad
  Zhanpeng Jin$^{1}$\\
  $^{1}$School of Future Technology, South China University of Technology, Guangzhou, China\\
  $^{2}$East China Normal University, Shanghai, China\\
  \texttt{202610196903@mail.scut.edu.cn}, \texttt{gaoyang2025@scut.edu.cn}, \texttt{zjin@scut.edu.cn}
}

\begin{document}

\maketitle

\begin{abstract}
\label{sec:abstract}
Existing human-activity datasets typically cover only one or two sensor channels
and often focus on short, isolated actions, leaving \emph{long-duration
compositional tasks} largely unsupported as benchmarks for temporal cross-modal
sensorimotor learning.
We introduce \textbf{PULSE} (\emph{$\mathbf{P}$hysiological $\mathbf{U}$nified
$\mathbf{L}$ong-duration $\mathbf{S}$ynchronized $\mathbf{E}$mbodiment}), a synchronized
five-modality dataset that enables benchmarking \emph{temporal cross-modal
sensorimotor learning}: how models recognize ongoing activities, anticipate future
contact, reconstruct unobserved signals, and remain robust when physiological,
motion, gaze, inertial, and tactile channels are partially observed.
Collected from 40 volunteers across 8 ecologically valid daily-activity scenarios
together with a dedicated motion-primitive collection, PULSE records full-body
optical motion capture with \emph{finger-level hand
articulation}, surface EMG, binocular eye tracking, wearable IMU, and a \emph{finger-level pressure array recording
quantitative grip force}, all time-aligned at 100\,Hz.
Each scenario is a long-horizon compositional task containing many
smaller sub-tasks, yielding over 7{,}300 annotated action segments. Each segment is tagged with a motor primitive, the hand involved, the manipulated object, and a natural-language
description with four paraphrased variants.
On top of the dataset, we define \textbf{multiple benchmark tasks} for temporal
cross-modal sensorimotor learning: activity-scenario and fine-grained action recognition, grasp
onset anticipation, missing-modality robustness, tactile-driven sub-second
grasp-state prediction, cross-modal pressure reconstruction, etc. Together these
tasks evaluate how information transfers across physiology, motion, gaze, and touch,
rather than reducing the dataset to an activity-label catalog. We evaluate three
backbone architectures, nine fusion strategies, seven published baselines, and
task-specific models including \textbf{SyncFuse} and \textbf{DailyActFormer}. Detailed
task definitions and results are in the Appendices. Data, annotations, and baseline
code are publicly available.
\end{abstract}

\section{Introduction}
\label{sec:introduction}

\newcommand{\yes}{\checkmark}
\newcommand{\no}{$\times$}

Human daily activities involve the seamless coordination of body posture, directed gaze,
muscle recruitment, fine-grained hand manipulation, and contact forces. Capturing this
coordination is important for applications ranging from ergonomic assessment and
rehabilitation to human--robot collaboration and cognitive load estimation.

\paragraph{Motivation.}
Several decades of motor-control and natural-behaviour research indicate that daily
activity is driven by a tightly coupled sensorimotor loop of \emph{gaze guidance},
\emph{anticipatory muscle activation}, and \emph{cutaneous force feedback}.
\textbf{(i) Gaze leads the hand.} Seminal work on activities of daily living shows
that the eyes fixate the next object of interest well before the hand arrives, and that
over 90\% of gaze is directly relevant to the ongoing manipulation
task~\cite{land1999roles,hayhoe2005eye}.
\textbf{(ii) Muscles fire before force.}
Johansson and Westling~\cite{johansson1984roles} showed that grip force is scaled in an
anticipatory, feed-forward manner, with EMG activation preceding the actual grasping
force by hundreds of milliseconds. This anticipatory coupling is only observable when
EMG is recorded synchronously with kinematics and contact.
\textbf{(iii) Cutaneous force feedback is essential for dexterity.} Digital anaesthesia,
which removes tactile force feedback, causes substantial deterioration in precision-grip
force control and object stability even when vision and proprioception remain
intact~\cite{monzee2003digital}. Yet existing activity datasets instrument only a
narrow slice of this loop: skeleton datasets discard physiological signals;
hand--object datasets omit gaze and EMG; egocentric video captures the scene without
biomechanical grounding; and contact data---when present at all---are limited to
thermal maps or tactile gloves that report raw sensor readings rather than grams-force.
Further, most HAR benchmarks consist of short, isolated clips, whereas real daily
behaviour is \emph{compositional}---a single task like ``clean up after a meal''
unfolds over minutes and chains together many primitive actions with subject-specific
ordering. These observations together motivate a 5-modality, long-horizon, 100\,Hz
time-aligned dataset covering full-body kinematics, EMG, gaze, IMU, and
quantitative finger-level pressure.

We address this gap with \textbf{PULSE}, which to our knowledge is the only
benchmark combining all five non-visual sensing modalities on a common 100\,Hz
timebase in daily-activity contexts (see Section~\ref{sec:related_work} and
Table~\ref{tab:dataset_comparison} for a detailed comparison). The 40-subject, 8-scenario
protocol yields 304 continuous task recordings ($\sim$6.8\,hours), complemented by an
additional 33 isolated motion-primitive recordings ($\sim$2.8\,hours). 282 of the 304
task recordings (from 36 volunteers) carry dense action-segment annotations totalling
\textbf{7{,}356 action segments} at two levels of granularity.
We position PULSE as a high-density, time-aligned benchmark rather
than a large-scale activity corpus: its contribution is the breadth, temporal
alignment, and dense labeling of non-visual sensorimotor signals over long
continuous tasks.

\paragraph{Contributions.}
\begin{enumerate}[leftmargin=*, nosep]
  \item \textbf{Sensor breadth and fidelity.}
        Five modalities are simultaneously captured and time-aligned at 100\,Hz:
        (i) \emph{full-body optical motion capture with finger-level hand articulation}
        via a Qualisys system (72 markers solved into a 56-joint skeleton with per-joint
        3-D positions \emph{and} quaternions, plus 3-D positions of all 10 fingertips),
        (ii) 8-channel forearm \emph{surface EMG} (Myo),
        (iii) 24-dim binocular \emph{eye tracking} (Dikablis~3; pupil position
        and size, fixations, saccades, and scene-camera gaze point),
        (iv) 160-dim \emph{wearable IMU} (10 inertial sensors), and
        (v) a \emph{50-channel finger-level pressure array} with \textbf{25 contact
        points per hand recording quantitative grip force (in grams)}---enabling not
        only binary contact detection but also force estimation.
  \item \textbf{Long-horizon compositional tasks.}
        Unlike datasets of short, isolated actions, each recording is a continuous
        0.3--4\,minute execution of a complex real-world task decomposed into dozens of
        interleaved sub-actions. This supports the study of long-horizon temporal
        structure and sub-task composition---goal-conditioned segmentation, sub-task
        anticipation, and hierarchical imitation learning---that short-clip HAR
        benchmarks cannot expose.
  \item \textbf{Densely instrumented daily-activity protocol and rich annotations.}
        \textbf{40 volunteers} perform \textbf{8
        scenarios} spanning work (office, shipping, packing) and household life
        (kitchen, dining cleanup/setup, beverage, clothes), yielding \textbf{304
        recordings ($\sim$6.8\,h)} and \textbf{7{,}356 action segments} across 282
        annotated recordings (36 volunteers). Each
        segment carries a motor primitive label (from a taxonomy of 18), a hand label
        (left / right / both), a manipulated-object label from a scene-specific
        whitelist, and a natural-language description with four paraphrased variants
        for language-grounded downstream use.
\end{enumerate}

In addition to the dataset, we release a benchmark suite covering multiple tasks:
activity-scenario recognition (T1), fine-grained action recognition (T2), grasp onset
anticipation (T3), missing-modality robustness (T4), tactile-driven grasp-state
recognition (T5), cross-modal pressure prediction (T6), and related extensions.
Baselines combine three backbones with nine fusion strategies plus seven published
baseline configurations~\cite{ordonez2016deep,furnari2019rulstm,gong2022futr,zhong2023afft,shamil2024handformer,wang2025actionllm}.
We further introduce task-specific models, including \textbf{SyncFuse} for T1
activity-scenario recognition and \textbf{DailyActFormer (DAF)} for T2 action recognition.
DAF is a multi-modal Transformer that jointly classifies verb, noun, and hand
from sensor windows. It attains the highest mean on the headline
joint top-5 metric, but margins over the strongest published baselines are small
and not significant; the same holds for SyncFuse against the strongest T1 baseline,
so we treat DAF and SyncFuse as strong baselines rather than decisive model contributions.
Results also include diagnostic negative and high-variance findings: T3 shows
that adding IMU to EMG can hurt grasp anticipation, and T6 exposes the difficulty
of pressure prediction under different input and selection settings. These tasks
are intended to define open challenges rather than solved problems. Task
definitions, baseline details, and quantitative results are in the Appendix.

\section{Related Work}
\label{sec:related_work}

\paragraph{Multi-modal activity datasets.}
Prior activity datasets provide complementary but partial views of human behavior.
Large-scale RGB/depth/skeleton datasets such as NTU~RGB+D~120~\cite{liu2019ntu120}
lack physiological and contact sensing; hand--object benchmarks
(GRAB~\cite{taheri2020grab}, ARCTIC~\cite{fan2023arctic},
H2O~\cite{kwon2021h2o}, ContactPose~\cite{brahmbhatt2020contactpose}, and the
KIT Bimanual dataset~\cite{krebs2021kit} with whole-body motion capture, data gloves,
RGB-D and IMUs) capture detailed hand kinematics but omit synchronized EMG and gaze; egocentric or
procedural video datasets such as EgoExo4D~\cite{grauman2024egoexo4d} and
Assembly101~\cite{sener2022assembly101} scale visual coverage without biomechanical
instrumentation; and wearable-only benchmarks
(Opportunity~\cite{roggen2010opportunity}, PAMAP2~\cite{reiss2012pamap2},
USC-HAD~\cite{zhang2012uschad}) primarily record IMU streams; free-living accelerometer
datasets (CAPTURE-24~\cite{chan2024capture24}, PAAWS~\cite{potter2025paaws}) trade
modality breadth for cohort size and recording hours. Large unlabeled
wrist-accelerometer cohorts (UK Biobank, NHANES) serve instead for self-supervised
pretraining~\cite{yuan2024harnet,tarale2026bioinspired}; PULSE can serve as a densely
labeled evaluation set for such models, and we include the pretrained harnet5
encoder~\cite{yuan2024harnet} as a T1 baseline.
The closest contemporary work is ActionSense~\cite{delpreto2022actionsense}, which
combines IMUs, EMG, eye tracking, open-fingered tactile gloves, and RGB in kitchen
activities. In contrast, PULSE provides full-body optical motion capture rather than
IMU-only kinematics, a 50-channel finger-level pressure array in grams-force,
including the fingertips, and eight daily-activity scenarios rather than a kitchen-only
setting.

Recent multimodal datasets further broaden sensor coverage but target different
questions. OctoNet~\cite{yuan2025octonet} introduces 12 modalities and MoCap-derived
3D pose labels for ambient-sensing-based pose estimation and sensor foundation models,
whereas PULSE treats full-body and finger-level MoCap as a first-class modality and
adds surface EMG, eye tracking, and quantitative finger pressure to study
sensorimotor coordination from muscle to motion to contact. MM-Fi~\cite{yang2023mmfi}
and HA-ViD~\cite{zheng2023havid} are also recent multimodal datasets, but rely mainly
on RF/video and lack physiological sensing. Table~\ref{tab:dataset_comparison}
summarises these distinctions. To our knowledge, PULSE is the only benchmark that
combines full-body optical motion capture, surface EMG, eye tracking, IMU, and a
50-channel finger-level pressure array on a common 100\,Hz timebase in a
daily-activity setting.

\begin{table}[t]
  \centering
  \caption{Comprehensive comparison of PULSE with representative human activity datasets.
    Modality columns: \textbf{V} = RGB video, \textbf{S} = full-body skeleton, \textbf{H} = finger-level
    hand tracking, \textbf{E} = surface EMG, \textbf{G} = gaze/eye tracking, \textbf{I} = inertial
    sensors, \textbf{P} = finger-level pressure. ``Sync'' denotes hardware-level time synchronization.
    PULSE is the \emph{only} dataset combining all five non-visual sensing modalities
    in ecologically valid daily-activity contexts.}
  \label{tab:dataset_comparison}
  \scriptsize
  \setlength{\tabcolsep}{2.5pt}
  \renewcommand{\arraystretch}{1.1}
  \begin{tabular}{@{}l c r r c ccccccc c r@{}}
    \toprule
    & & & & & \multicolumn{7}{c}{\textbf{Modalities}} & & \\
    \cmidrule(lr){6-12}
    Dataset & Year & \#Sub & Hrs & Set &
      V & S & H & E & G & I & P &
      Sync & Hz \\
    \midrule
    \multicolumn{14}{@{}l@{}}{\emph{Video / Skeleton datasets}} \\
    NTU RGB+D 120~\cite{liu2019ntu120}       & 2020 & 106 & ---   & Lab  & \yes & \yes & L & \no & \no & \no & \no & \yes & 30 \\
    Assembly101~\cite{sener2022assembly101}  & 2022 & 53  & 513   & Lab  & \yes & \no  & L & \no & \no & \no & \no & \yes & 60 \\
    EgoExo4D~\cite{grauman2024egoexo4d}      & 2024 & 740 & 1286  & Real & \yes & \no  & L & \no & \yes & \yes & \no & \yes & 30 \\
    \midrule
    \multicolumn{14}{@{}l@{}}{\emph{Hand/Object interaction (motion capture)}} \\
    GRAB~\cite{taheri2020grab}               & 2020 & 10  & 3.8   & Lab  & \no  & \yes & \yes & \no & \no & \no & \no & \yes & 120 \\
    ContactPose~\cite{brahmbhatt2020contactpose} & 2020 & 50 & ---  & Lab  & \yes & \no  & \yes & \no & \no & \no & L & \no   & 30 \\
    H2O~\cite{kwon2021h2o}                   & 2021 & 4   & ---   & Lab  & \yes & \no  & \yes & \no & \no & \no & \no & \yes & 30 \\
    ARCTIC~\cite{fan2023arctic}              & 2023 & 10  & 1.4   & Lab  & \yes & \yes & \yes & \no & \no & \no & \no & \yes & 30 \\
    KIT Bimanual~\cite{krebs2021kit}         & 2021 & 2   & ---   & Lab  & \yes & \yes & \yes & \no & \no & \yes & \no & L    & mixed \\
    \midrule
    \multicolumn{14}{@{}l@{}}{\emph{Wearable sensor datasets}} \\
    Opportunity~\cite{roggen2010opportunity} & 2010 & 4   & 25    & Lab  & \no  & \no  & \no  & \no & \no & \yes & \no & L    & 30 \\
    PAMAP2~\cite{reiss2012pamap2}            & 2012 & 9   & 10    & Lab  & \no  & \no  & \no  & \no & \no & \yes & \no & \yes & 100 \\
    USC-HAD~\cite{zhang2012uschad}           & 2012 & 14  & 7     & Real & \no  & \no  & \no  & \no & \no & \yes & \no & ---  & 100 \\
    CAPTURE-24~\cite{chan2024capture24}      & 2024 & 151 & 2562  & Real & \no  & \no  & \no  & \no & \no & \yes & \no & ---  & 100 \\
    PAAWS~\cite{potter2025paaws}             & 2025 & 272 & 2221  & Both & \no  & \no  & \no  & L   & \no & \yes & \no & ---  & 80 \\
    \midrule
    \multicolumn{14}{@{}l@{}}{\emph{Multimodal sensor (recent)}} \\
    ActionSense~\cite{delpreto2022actionsense}    & 2022 & 10  & 9.1   & Lab  & \yes & L    & L    & \yes & \yes & \yes & L & \yes & mixed \\
    MM-Fi~\cite{yang2023mmfi}                     & 2023 & 40  & 3.0   & Lab  & \yes & \yes & \no  & \no & \no & \no & \no & \yes & 30 \\
    HA-ViD~\cite{zheng2023havid}                  & 2023 & 30  & 86.9  & Lab  & \yes & \no  & \no  & \no & \no & \no & \no & \yes & 30 \\
    OctoNet~\cite{yuan2025octonet}                & 2025 & 41  & 41    & Lab  & \yes & \yes & L    & \no & \no & \yes & \no & \yes & mixed \\
    \midrule
    \rowcolor{blue!8}
    \textbf{PULSE (Ours)}              & \textbf{2026} & \textbf{40} & \textbf{9.6} & \textbf{Lab} &
      \yes & \textbf{\yes} & \textbf{\yes} & \textbf{\yes} & \textbf{\yes} & \textbf{\yes} & \textbf{\yes} &
      \textbf{\no} & \textbf{100} \\
    \bottomrule
  \end{tabular}

  \vspace{0.3em}
  {\scriptsize ``\yes'' = full support, ``L'' = limited/partial support, ``\no'' = not
    provided, ``---'' = single-device or single-modality dataset where sync is not applicable.
    ContactPose uses post-hoc thermal contact maps. PAMAP2 provides IMU and heart-rate streams, but no EMG.
    KIT Bimanual: 588 demonstrations of 12 bimanual kitchen actions; its sensors are started by a software trigger and aligned by timestamps.
    CAPTURE-24 and PAAWS hours are annotated hours; PAAWS combines a semi-naturalistic laboratory protocol (252 participants, 657\,h)
    with a 7-day free-living protocol (20 participants, 1{,}564\,h), and its EMG is limited to polysomnography of a sleep protocol.
    PULSE duration includes $\sim$6.8\,h of daily-activity scenarios plus $\sim$2.8\,h of motion-primitive recordings.
    PULSE streams are time-aligned on a common 100\,Hz timebase from each device's recorded start timestamp.
    PULSE provides MoCap (56 joints, 620d), finger-level hand, 8-ch EMG, 24d gaze, 160d IMU, and 50-ch pressure.}
\end{table}

\paragraph{Grasp and contact sensing.}
Physical contact during manipulation has been studied from complementary angles.
ContactDB~\cite{brahmbhatt2019contactdb} shows that post-hoc thermal imprints can
reconstruct where a hand touched an object during a grasp; Sundaram et
al.~\cite{sundaram2019tactile} demonstrate that a high-resolution hand pressure array
alone carries enough information to classify grasp type and object identity. Both lines
of work take \emph{the grasp itself} as the unit of analysis. PULSE takes a
different perspective: rather than studying grasps as isolated events, we embed a
50-channel finger-level pressure array inside continuous long-horizon activities, co-registered
with whole-body kinematics, EMG, and gaze. This shifts the research question from
\emph{what} is being grasped to \emph{when, why, and how} a contact event unfolds within
a multi-minute task---for instance, whether contact onset can be predicted from the
preceding motion and muscle activation, as we illustrate in Appendix~\ref{sec:grasp-phase-main}.

\paragraph{Cross-modal force and pressure prediction.}

Two regimes have been studied in isolation. (i) \emph{EMG\,$\to$\,force} models regress isometric grip force from sEMG with CNN/LSTM/Transformer backbones, reaching Pearson $r$ of 0.91–0.99 on single-target laboratory data~\cite{hajian2021generalized,ghorbani2023grip}, but target a single scalar under constrained contractions rather than distributed pressure during naturalistic manipulation. (ii) \emph{Kinematics\,$\to$\,force} methods recover forces from hand and object motion via dynamics plus learned residuals~\cite{pham2015force,pham2018hand} or from end-effector deformation~\cite{zhu2024forcesforfree}, while omitting muscle activation. PULSE is the first dataset to pair time-aligned MoCap (with finger articulation), sEMG, IMU, and gaze with a 50-channel quantitative finger-level pressure array in long-horizon daily activities, providing a unified testbed for EMG-only, kinematics-only, and fused predictors of distributed finger-level pressure (T6).

\section{Dataset}
\label{sec:dataset}

We introduce \textbf{PULSE}, a multi-modal daily-activity dataset collected in a
controlled laboratory environment. The dataset provides five time-synchronized
sensor modalities from 40 volunteers performing 8 ecologically valid activity
scenarios.

\subsection{Data Collection Protocol}
\label{sec:protocol}

We recruited 40 healthy adult volunteers from a university campus. The cohort
was 18--26 years old (mean 21.2, SD 2.1), with 31 male and 9 female
participants; mean height was 174.6\,cm (SD 7.2, range 162--190) and mean
weight was 65.1\,kg (SD 10.1, range 49.0--86.3). Self-reported handedness was
right for 26 participants and left for 14. Each volunteer
performed a set of predefined daily-activity scenarios in an
instrumented laboratory room, whose layout was designed to mimic common
real-world settings, including a work desk, kitchen counter, dining table,
wardrobe area, and package-preparation station. Before each recording,
volunteers heard a short verbal description of the target scenario; beyond this
high-level goal, they were allowed to complete the task in their own style and
at their own pace.

\paragraph{Participant safety and informed consent.}
Throughout data collection, we prioritized participant well-being and privacy.
The wearable sensors were selected so that they did not restrict free movement
within the designated recording area, and commercially available CE/FDA-certified
sensors were preferred whenever possible. Before each session, researchers
explained the potential risks, including risks associated with tools and props
used in the scenarios and a modest reduction in hand dexterity from wearing the
sensor suite, and obtained written informed consent. Each session lasted
approximately 2--3 hours, and participants received CNY 100--150 (approximately USD 14--21) in compensation;
participants who paused or withdrew during the session still received full
compensation. In addition, all participants signed a media-release agreement for
the released scene-camera and eye-tracking videos. The released first-person videos contain no
audio track and do not show the participant's own face; laboratory members who appear
incidentally consented to the release, and their faces are pixelated. Participants were informed that
first-person video and biomechanical signals would be released as research data.
The full protocol was reviewed and approved by the Ethics Committee of South China
University of Technology (approval no.\ \mbox{S-2025-026-01}, 4~March 2025).

Each recording corresponds to one complete execution of a single activity
scenario and lasts 0.3--4.0 minutes (mean 1.3 minutes). The public release
contains \textbf{304 task recordings} from the 8 daily-activity scenarios
(S1--S8), totaling approximately 6.8 hours of time-aligned multi-modal data,
plus 33 isolated motion-primitive recordings (scenario S9, approximately
2.8 hours). Not every volunteer completed every scenario, and some recordings
miss one or more sensor streams due to device availability, sensor dropouts, or
task executions that deviated from the expected protocol; this produces the
modality-availability distribution reported in Section~\ref{sec:stats}. Among
the 304 task recordings, \textbf{282 recordings from 36 volunteers} have dense
action-segment annotations, yielding \textbf{7{,}356 segments}. The benchmark
experiments in this paper are evaluated on this annotated pool, while the
remaining 22 recordings are released as raw aligned sensor streams (except
v4/s1 and v4/s4, which contain no data and appear only in \texttt{metadata/recordings.csv}).

\paragraph{Synchronization.}
All five sensor streams are resampled onto a unified \textbf{100\,Hz} time base
and placed on it from each device's recorded start timestamp; the aligned window
of each recording is the overlap of all available streams (per-stream start and
end times are given in each recording's \texttt{alignment\_metadata.json}).

\subsection{Sensor Specifications}
\label{sec:sensors}

\begin{figure}[t]
  \centering
  \includegraphics[width=0.95\linewidth]{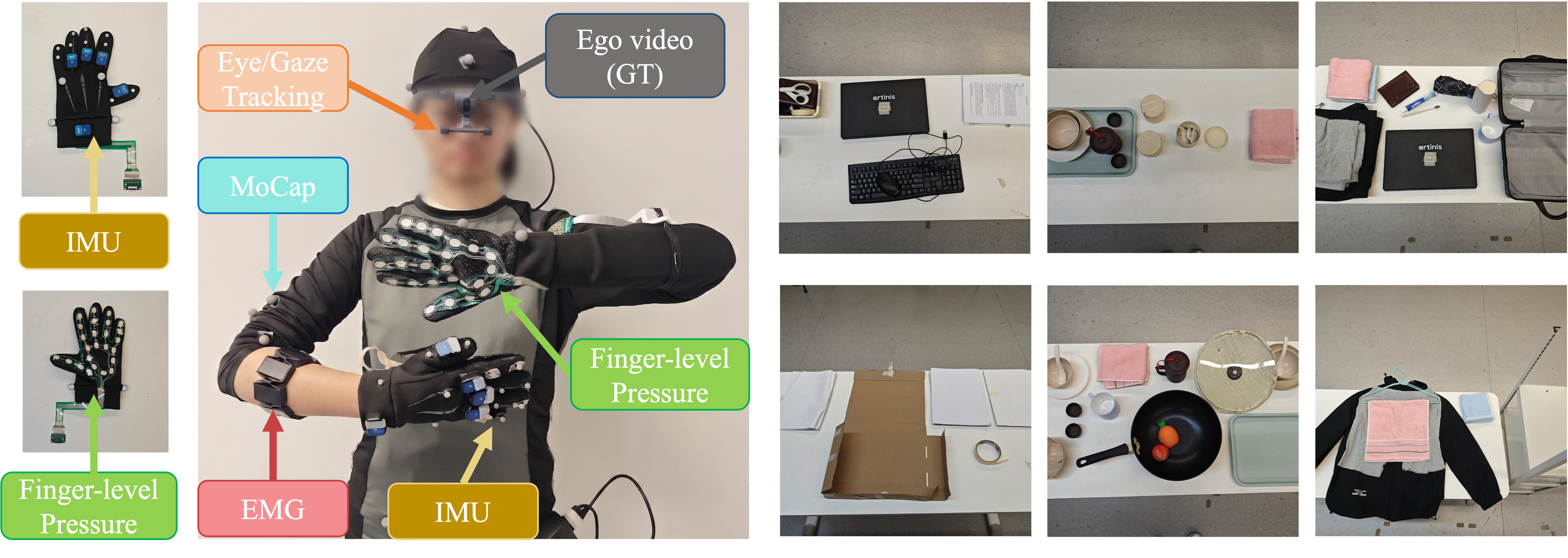}
  \caption{Left: A volunteer wearing the full PULSE sensor suite: optical
    motion-capture markers across the body and fingers, a Myo Armband for
    surface EMG, Dikablis Glasses~3 for binocular eye tracking, 10 wearable IMUs,
    and finger-level pressure arrays on both hands. Right: Representative photos of the recording environments, covering
    \emph{work-oriented} tasks (desk organization, package preparation, luggage
    packing) and \emph{home-oriented} tasks (kitchen organization, table
    cleaning, and clothing handling). Each photo shows the props in their
    initial state.}
  \label{fig:volunteer-setup}
\end{figure}

PULSE provides five time-synchronized modalities at 100\,Hz (Figure~\ref{fig:volunteer-setup}). We summarize their
raw and processed dimensionalities below; a compact summary is provided in
Appendix Table~\ref{tab:sensors}.

\noindent\textbf{Motion capture (MoCap).}
Whole-body kinematics are recorded with a Qualisys optical marker system that
tracks 56 joints. The raw output contains 3D joint positions and per-joint
quaternions (422 dimensions). After preprocessing
(Section~\ref{sec:app_annot_preproc}), hip-relative coordinates and per-joint
velocity estimates are appended, expanding the feature dimensionality to
620 dimensions.

\noindent\textbf{Surface EMG.}
Muscle activation is recorded by a Myo Armband worn on the right forearm,
providing 8 channels of surface EMG at 100\,Hz. The raw signed Myo signal ($\sim$200\,Hz native) is linearly interpolated to 100\,Hz without band-pass filtering or rectification.

\noindent\textbf{Eye/gaze tracking.}
Binocular gaze is captured by Dikablis Glasses~3, producing a 24-dim binocular feature vector covering 2D pupil positions, per-eye pupil size, fixation/saccade events, and scene-camera gaze point.

\noindent\textbf{Inertial measurement unit (IMU).}
Ten wearable inertial measurement units are mounted on both hands:
one on the wrist and one on each of the thumb, index, middle, and ring fingers.
Each unit records triaxial accelerometer, gyroscope, and magnetometer readings,
Euler angles, and an orientation quaternion (16 signals), yielding a 160-dimensional signal
(the released files add each unit's temperature and battery level, 180 columns).

\noindent\textbf{Finger-level Pressure.}
A pressure-sensor array customised by its manufacturer (GT-G0505M) with 25 sensing points per hand (50 total) records
contact force during manipulation.

\noindent\textbf{Scope of the benchmark protocol.}
Scene-camera video synchronized with the five sensor streams (Dikablis
Glasses~3, $1920 \times 1080$, 30\,fps) and a gaze-overlay rendering of the same view (kept at its recorded frame rate) are released with the dataset (622 files covering 312 recordings, 63.1\,GB), \emph{but excluded from the T1–T7 protocol}. We position
PULSE as a non-visual sensor benchmark because its core 100\,Hz anticipatory
EMG--motion signal is not recoverable from 30\,fps RGB video, and this scope
matches related wearable-sensor benchmarks that release video only as auxiliary
material. Per-modality availability over the 304 recordings is reported in
Section~\ref{sec:stats}.

\subsection{Activity Scenarios}
\label{sec:scenes}

The dataset covers 8 daily-activity scenarios designed to preserve real-life
representativeness while spanning a wide range of manipulation and locomotion
behaviors (Figure~\ref{fig:volunteer-setup}). The scenarios (S1--S8) are:
\textbf{S1} office desk organization,
\textbf{S2} package shipping preparation,
\textbf{S3} kitchen spice organization,
\textbf{S4} post-meal table cleaning,
\textbf{S5} pre-meal table setting,
\textbf{S6} business-travel luggage packing,
\textbf{S7} coffee and beverage preparation, and
\textbf{S8} clothes hanging and folding. They are familiar tasks that can be performed
without instruction in one instrumented session, and together they span four
manipulation regimes: bimanual coordination, tool use, deformable objects, and fine
in-hand manipulation. Detailed task descriptions for each
scenario are provided in Appendix~\ref{sec:app_scenarios}.

\subsection{Long-Horizon Compositional Tasks}
\label{sec:composition}

A defining property of PULSE is that each recording captures a \emph{complete,
long-horizon} execution of a realistic task, rather than a short isolated
action. Volunteers are given only a high-level goal, such as ``tidy the desk and
prepare for work'', ``pack for a three-day business trip'', or ``clean up after
a meal'', and choose their own action order and sub-actions. A single recording
therefore contains dozens of smaller subtasks with natural transitions between
them.

\paragraph{Two-level annotation structure.}
The dataset is organized at two levels: (1) Scenario (L1, 8 classes, 0.3--4 minutes per recording): the high-level goal of a recording, such as \emph{post-meal table cleaning}. (2) Action segment (L2, mean 2.2\,s): a dense temporal segment annotated with a motion primitive from an 18-class vocabulary (Appendix~\ref{sec:app_annot_preproc}), a hand label, the manipulated object from a scenario-specific whitelist, and one natural-language description plus 4 paraphrased variants.

On average, each annotated recording contains approximately 26 action segments,
and the empirical transition distribution preserves the strong temporal
structure of real task execution. This two-level design supports long-horizon
research problems, including goal-conditioned action segmentation, subtask
prediction, hierarchical imitation learning, and procedure-level evaluation,
which are not naturally expressed in short-clip HAR benchmarks. Downstream
benchmarks that require coarser label sets, such as segmentation tasks, use
task-specific explicit rollups from the 18 primitives to coarser classes; the
corresponding mappings are given in the task appendices.

\paragraph{Isolated motion-primitive recordings.}
In addition to the 8 task scenarios, 33 volunteers completed a dedicated
\emph{motion-primitive protocol} (S9): they performed each atomic motion unit in
isolation, including \emph{reach, grasp, release, wrist rotation, press, pull,
push, slide, pinch, spread, lift, lower, insert, extract, fold, tear, wipe}, and
additional upper-body, whole-body, and fine-grained primitives. They then
performed a library of common task compositions, including pick-and-place,
opening and closing containers, pouring, stirring, folding clothes, wiping
surfaces, and applying labels or stickers. This subset totals approximately
2.8 hours and is annotated at the primitive level, providing a clean and
segmentable counterpart to the in-context long-horizon scenarios. It can be used
for supervised motion-primitive classification, primitive pretraining followed
by transfer to in-context recordings, or as a motion library for imitation
learning. The complete primitive and task taxonomy is provided in
Appendix~\ref{sec:app_primitives}.

\subsection{Annotations and Preprocessing}
\label{sec:annotations}

Each recording is assigned one of the 8 scenario labels (S1--S8) and densely
annotated at the segment level. The 282 annotated recordings contain
\textbf{7{,}356 action segments}. Each segment provides start and end times
(1\,s resolution), one action class from \textbf{18 motion primitives} (e.g.,
\textit{grasp, move, rotate, open, close, place, push, pull}), the acting hand
(\textit{left}, \textit{right}, or \textit{both}), the primary manipulated
object selected from a scenario-specific vocabulary of 50$+$ objects, and one natural-language description with 4
paraphrased variants for language-guided downstream use. Segments are generated
by a \textbf{VLM-assisted annotation pipeline}: compressed first-person
eye-tracker video is passed to a vision-language model under a scene-constrained
prompt that enforces the object and primitive vocabularies. The full prompt and
pipeline are provided in Appendix~\ref{sec:app_annot_preproc}. Every annotated
recording was then \textbf{reviewed by a human} segment by segment, and each segment
was confirmed or corrected, so all released segments have been confirmed or
corrected by a human (Appendix~\ref{sec:app_human_validation}). The 1\,s
boundary resolution should be interpreted as semantic action-interval
annotation rather than a 100\,Hz event timestamp: sub-second benchmark labels
are derived from the synchronized sensor streams, not from manually specified
sub-second boundaries.

For preprocessing, MoCap data are transformed into a hip-relative coordinate
system and augmented with velocity features (422$\to$620 dimensions). All
modalities are subsampled to 20\,Hz (every fifth frame). Channel-wise z-score
normalization uses training-set statistics only. Variable-length sequences are
zero-padded and masked at the batch level. The complete annotation structure,
taxonomy, and preprocessing details are provided in
Appendix~\ref{sec:app_annot_preproc}.

\subsection{Dataset Statistics}
\label{sec:stats}

\paragraph{Modality availability.}
EMG ($\approx$99\%) and IMU ($\approx$98\%) together form a near-complete
wearable backbone. Eye tracking ($\approx$93\%), MoCap ($\approx$81\%), and
finger-level pressure ($\approx$70\%) also maintain substantial coverage. The
per-modality availability over all 304 recordings is reported in Appendix
Table~\ref{tab:sensors}. The T4 experiment in Appendix~\ref{sec:results}
evaluates inference robustness under this realistic deployment pattern of
modality availability.

\begin{figure}[t]
  \centering
  \includegraphics[width=\linewidth]{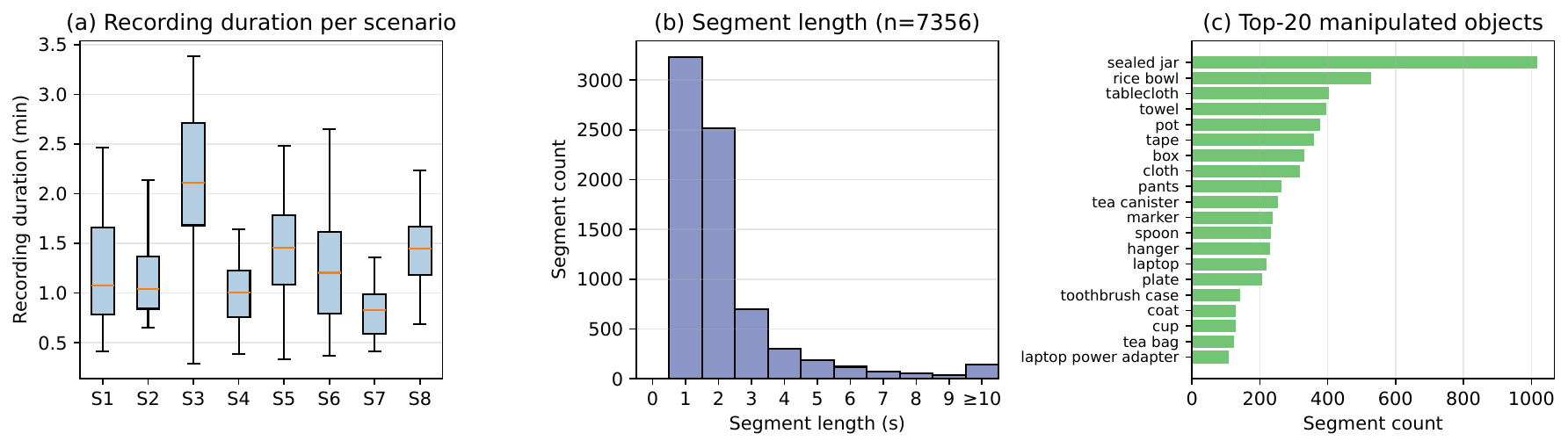}
  \caption{Dataset statistics: (a) recording durations by scenario; (b)
    action-segment lengths, with a median of approximately 2\,s; and (c) the top
    20 manipulated objects by segment count.}
  \label{fig:dataset-stats}
\end{figure}

\paragraph{Train/test split.}
All benchmark experiments in this paper use a unified
\textbf{volunteer-stratified 5-fold cross-validation} protocol over the
annotated volunteers. Each fold rotates 5--7 volunteers through the
test set, so every volunteer of a task's pool appears in the test set exactly once;
training and test volunteers are always disjoint within a fold. Per-task
pools differ by modality requirement: T1, T4, and T7 evaluate on the
33 volunteers with EMG and skeleton MoCap (test folds of 7/7/7/6/6 volunteers),
and T2, T3, T5, and T6 on the 27 annotated volunteers with pressure data
(6/6/5/5/5); modality-availability filtering is applied to held-out
recordings inside each fold (per-configuration counts in Appendix
Table~\ref{tab:pools}). From each fold's training volunteers, 15\% are held out
as inner-validation volunteers for model selection and early stopping, and each
test fold is evaluated once. Headline metrics average three seeds per fold and
are reported as mean $\pm$ std (ddof\,=\,1) across the five fold means; methods
are compared with paired $t$-tests over the fold means. All
reported numbers therefore measure generalization to unseen individuals.

\section{Benchmarks: Multi-axis Generalization Probes}
\label{sec:benchmarks}

The PULSE benchmark is organized as a multi-axis generalization probe: rather than reducing the dataset to a list of activity-recognition tasks, we identify six distinct axes along which training and deployment conditions can diverge, and define one probe task per axis to measure how methods generalize when the corresponding distribution shifts. This framing is motivated by two observations from prior work. First, real-world wearable AI deployment rarely fails along a single axis—a model trained on one cohort, with full sensor availability, on isolated demonstrations, will face shifts on all of these dimensions simultaneously when deployed. Second, single-task benchmarks systematically underestimate this difficulty: a method that achieves strong cross-subject performance may fail under modality dropout, and a method robust to modality dropout may fail under cross-context transfer. PULSE makes these axes individually measurable on a single time-aligned dataset, enabling controlled comparison of where methods generalize and where they fail.

Table \ref{tab:multi_axis} summarizes the six benchmark themes. F0 is a descriptive case study that examines the anticipatory sensorimotor structure in the time-aligned 100\,Hz streams, and motivates the five generalization probes that follow. F1--F5 each pair a probe task with an explicit train-to-deployment transfer setting and report a headline finding from our baseline evaluation. Full task definitions, baselines, and results are reported in Appendix~\ref{sec:tasks} and Appendix~\ref{sec:app_results_new}.

\begin{table*}[t]
\centering
\footnotesize
\setlength{\tabcolsep}{4pt}
\renewcommand{\arraystretch}{1.18}
\caption{ PULSE benchmark suite. F0 reports an anticipatory-coordination case study that motivates the framework; F1–F5 are generalization probes covering subject, modality, temporal, cross-modal, and context shifts. F1–F5 headline results are reported under volunteer-stratified 5-fold cross-validation (mean ± std over the five fold means; three seeds per fold); F0 values are mean ± SD over the 458 events; see Appendix \ref{sec:tasks} and \ref{sec:results} for full protocol.}
\label{tab:multi_axis}
\begin{tabularx}{\linewidth}{
>{\raggedright\arraybackslash}p{0.23\linewidth}
>{\centering\arraybackslash}p{0.10\linewidth}
>{\raggedright\arraybackslash}p{0.2\linewidth}
>{\raggedright\arraybackslash}X}
\toprule
\textbf{Theme} & \textbf{Probe} & \textbf{Transfer Setting} & \textbf{Headline Results} \\
\midrule

F0: Anticipatory coordination (case study)
& \ref{sec:case-study}
& --
& EMG onset $-580 \pm 380$ ms; motion onset $-627 \pm 377$ ms before pressure contact ($n = 458$ clean grasp events)\\

F1: Activity recognition under cross-subject
& T1--T7
& train subjects $\rightarrow$ unseen subjects
& T1 Macro F1 $0.440 \pm 0.063$ (SyncFuse; n.s.\ vs.\ CTR-GCN); adding pressure: $+0.053$ ($p = 0.034$)\\

F2: Robustness to missing modalities
& T4
& full sensors $\rightarrow$ missing sensors
& F1 $0.480 \pm 0.063$ (4-mod, modality-dropout training) $\rightarrow 0.338 \pm 0.045$ (no MoCap) $\rightarrow 0.153 \pm 0.009$ (EMG only). \\

F3: Predictive Contact dynamics
& T3, T5
& past window $\rightarrow$ future state
& T3: AUC \textbf{0.737} $\pm$ 0.055 (anticipation; n.s.\ vs.\ EMG); T5: SustainedGrasp F1 lift \textbf{+0.194} (adding pressure).  \\

F4: Cross-modal physical-signal regression
& T6, T7
& EMG/MoCap $\rightarrow$ pressure or hand pose
& T6: $r_{\max}$ \textbf{0.305$\pm$0.083} (models n.s.); T7: MPJPE 19.71$\pm$0.99~mm vs.\ 20.57$\pm$1.14~mm for the mean pose (n.s.) \\

F5: Few-shot transfer from isolated primitives
& T2 (S9)
& isolated primitives $\rightarrow$ daily actions
& 5--100\% labels: $|\Delta| \leq 0.8$ pp, n.s.\ (inconclusive with a 90-segment pretraining pool) \\
\bottomrule
\end{tabularx}
\end{table*}

\section{Discussion and Limitations}
\label{sec:discussion}

\paragraph{Dataset scale and variance.}
PULSE comprises 40 volunteers, 304 task recordings, and 33 isolated
motion-primitive recordings. This is sufficient for baseline construction but
modest for large-scale activity modeling, so we frame PULSE as a high-density,
time-aligned benchmark rather than a large-scale corpus. Consistent
with this scale, stochastic training has high variance, so we adopt a
unified \textbf{volunteer-stratified 5-fold cross-validation} protocol
across all headline tasks (Section~\ref{sec:dataset}) and report mean
$\pm$ std across the 5 fold means, averaging three seeds per fold.

\paragraph{Cohort, EMG hardware, and scope.}
The task corpus (S1--S8) is $\sim$6.8\,h, and the cohort is young adults (18--26
years) with no older or clinical participants and limited body-size variability; for a
dataset positioned toward prosthetics, rehabilitation, and ergonomics this is a real
generalization limit. A single Myo on the right forearm gives no bimanual EMG, covers
only a coarse subset of forearm musculature, and records the non-dominant arm of the 14
left-handed participants; the EMG-only ceilings in T3, T5, and T6a partly reflect this
choice. Together, these bound the benchmark to evidence about which modalities carry
which information under controlled conditions, not evidence of transfer to older,
clinical, or free-living cohorts. We did not run a bare-hand control, so the recordings
do not show whether the hand-mounted sensor suite alters hand kinematics.

\paragraph{Modality coverage and 3-modality subset.}
At the task-recording level, EMG and IMU each exceed 98\% coverage, eye tracking
is approximately 93\%, MoCap approximately 81\%, and pressure approximately
70\%. Hardware-heavy experiments (MoCap, pressure) therefore use smaller and
potentially less representative subject subsets. The \emph{three-modality core}
(IMU + EMG + EyeTrack) is the most deployment-friendly setting; 5-modality
results are included for completeness but should be interpreted with this
caveat.

\paragraph{MoCap and deployment caveats.}
Raw absolute joint coordinates are highly subject-dependent, so all MoCap inputs use
hip-relative coordinates with velocity features. Optical MoCap also requires marker placement and lab
calibration, making the wearable subset (EMG + IMU + EyeTrack) more practical,
though occlusion, lighting, and electrode-placement shifts remain open.

\paragraph{Fusion on small datasets.}
In the T1 backbone and fusion ablations, the best single-modality model (BiLSTM on
MoCap, $0.449 \pm 0.057$) and the best fusion strategy (weighted late fusion,
$0.438 \pm 0.078$) reach comparable Macro-F1, and no fusion strategy differs
significantly from plain late fusion, reflecting the difficulty of
jointly optimizing heterogeneous modalities with limited data. This points to
optimization---not raw modality count---as the dominant constraint at our scale.

\paragraph{RGB video as auxiliary material.}
We release synchronized scene-camera video, but exclude visual features from the
headline benchmark because PULSE is designed to test 100\,Hz non-visual
sensorimotor signals---EMG, finger-level pressure, and gaze dynamics---that 30\,fps
RGB video cannot capture. The video remains useful for annotation and future
vision-language or gaze-conditioned visual extensions.

\paragraph{Published baselines validate the benchmark.}
On T2, we benchmark DAF against seven published baselines (HandFormer,
FUTR, AFFT, RU-LSTM, DeepConvLSTM (IMU and 3-modality), ActionLLM) under
volunteer-stratified 5-fold cross-validation. DAF obtains the highest mean
on the three headline metrics, but its act\_vn@5 margin over HandFormer is not significant,
and absolute act\_vn@5 still sits at $0.468$
in this small-subject regime, so off-the-shelf architectures do not
transfer trivially. The same interpretation applies to the extended
tasks: on T3 the 4-modality input has the highest mean AUC but is not
significantly better than EMG alone ($p = 0.217$), and on T6 no input
modality or architecture is significantly better than the others. We therefore view these results primarily as
benchmark diagnostics and open challenge definitions.

\paragraph{Intended use cases and research opportunities.}
Intended use cases. Beyond the T1–T7 benchmark, PULSE supports 100\,Hz motor-control analysis, EMG–kinematics–pressure modelling for prosthetics and rehabilitation, long-horizon robot imitation learning from hierarchical action labels, self-supervised representation learning for wearable sensing, gaze-based attention and task-switching analysis, and MoCap-to-pressure prediction for VR/AR haptics.

\paragraph{Future directions.}
Immediate extensions include pressure-based refinement of action boundaries from
1\,s to frame-level precision, self-supervised pretraining on unlabeled wearable
recordings, and inference-time robustness to missing modalities, a realistic
failure mode for wearable sensor arrays.

\section{Broader Impact}
\label{sec:broader-impact}

PULSE may support research on myoelectric prosthetics, rehabilitation,
human--robot collaboration, VR/AR haptics, ergonomics, and wearable activity
sensing. At the same time, fine-grained behavioural and biometric-class signals
such as EMG, eye tracking, pressure, and first-person video can create privacy
and misuse risks. The released first-person videos contain no audio track and do not show the
participant's own face, and faces of laboratory members who appear incidentally are pixelated, but they may still reveal task context, hand appearance, and manipulation
style. All participants gave informed consent under a protocol approved by the Ethics Committee of South China University of Technology (\mbox{S-2025-026-01}).
The CC BY-NC license and additional use restrictions disallow commercial
redeployment, re-identification attempts, covert worker surveillance, and
biometric identification without freely given consent. Models trained on PULSE
should not be deployed on new populations without additional validation and
ethics review.

\section{Conclusion}
\label{sec:conclusion}

We presented \textbf{PULSE}, a time-aligned 100\,Hz benchmark of full-body MoCap, sEMG, binocular eye tracking, IMU, and finger-level pressure across eight long-horizon daily-activity scenarios, together with seven tasks (T1–T7) that probe cross-subject generalization, modality dropout, contact anticipation, tactile grasp-state recognition, cross-modal pressure prediction, and EMG-to-hand-pose reconstruction. Across three backbones, nine fusion strategies, and seven published baselines, no modality dominates across tasks: EMG and MoCap are comparably predictive for grasp anticipation, MoCap is the strongest single modality for activity-scenario recognition, and adding finger-level pressure improves grasp-state recognition and, for SyncFuse, activity-scenario recognition; fusion remains non-trivial at this data scale. Substantial headroom across every task positions PULSE as an open benchmark for sensorimotor learning rather than a solved one. Data, code, and a representative sample are publicly available;\footnote{Dataset: \url{https://huggingface.co/datasets/velvet-pine-22/PULSE}; code: \url{https://huggingface.co/velvet-pine-22/PULSE-code}; sample: \url{https://huggingface.co/datasets/velvet-pine-22/PULSE-sample}.} the dataset is archived under the permanent DOI \href{https://doi.org/10.57967/hf/11314}{10.57967/hf/11314}.

\begin{ack}
This work was supported in part by the National Natural Science Foundation of China (62302168, W2532047), the Guangdong S\&T Programme (2025B0101120004), the Guangdong Provincial Department of Science and Technology (2023CX10X070), the Guangdong Basic and Applied Basic Research Foundation (2026A1515010254), the Guangdong Provincial Key Laboratory of Human Digital Twin (2022B1212010004), the Guangzhou Basic and Applied Basic Research Foundation (2024A04J6311), and the Shenzhen Holdfound Foundation Endowed Professorship. This work is partially supported by the High Performance Computing Platform of South China University of Technology.

The authors declare no competing interests.
\end{ack}

\bibliographystyle{unsrtnat}
\bibliography{references}

\newpage
\appendix
\section{Benchmark Tasks (Detailed Definitions)}
\label{sec:tasks}
This appendix expands on the brief overview in Section~\ref{sec:benchmarks} with
full definitions for each task (input format, label taxonomy, evaluation protocol).

\paragraph{T1: Activity-Scenario Recognition.}
Sequence classification into one label per recording from the eight daily-activity scenarios.
Models aggregate evidence across the full sequence; we evaluate single
modalities (IMU, EMG, MoCap), nine fusion strategies on IMU+MoCap+EMG, and published methods with their original inputs under
volunteer-stratified 5-fold cross-validation (3 seeds per fold; 1 seed for Tables~\ref{tab:s_backbone}--\ref{tab:s_fusion}); published methods are
also compared under matched inputs (Table~\ref{tab:t1-matched}).

\paragraph{T2: Action Recognition.}
Recognition of a four-tuple label
($\textit{verb\_fine}, \textit{verb\_composite}, \textit{noun}, \textit{hand}$)
from 20\,Hz multi-modal sensor windows: 4\,s windows at 1\,s stride are cut from
each recording, and a window takes the labels of the annotated segment that
contains its center (a segment-aligned variant with one sample per segment is
reported in Table~\ref{tab:action-seg}). The model predicts
four heads: 17 fine-grained verbs, a 6-class verb-composite rollup,
33 retained nouns (at least 50 segments over all 36 annotated volunteers), and 3 hand labels
(left/right/bimanual); 7{,}076 of the 7{,}356 segments (all 36 annotated volunteers) pass this label filtering; within the
27-volunteer T2 pool, 5{,}312 of 5{,}492 segments pass it before modality-availability filtering
(effectively 26 volunteers, since v5 has neither usable IMU nor MoCap);
volunteers rotate through the test partition under volunteer-stratified
5-fold cross-validation. The headline metric is joint top-5 action
accuracy for $(\textit{verb\_fine}, \textit{noun})$, $\mathrm{act\_vn@5}$;
we also report \texttt{verb\_fine@5} and \texttt{noun@5}.
The verb-composite and hand heads are retained in the released labels but are
not headline metrics; they are reported in Table~\ref{tab:action-recognition-full}
together with the accuracy of the training-majority class.

\paragraph{T3: Grasp Onset Anticipation.}
Window-based binary classification derived from anticipatory-coordination findings.
Given 1.0\,s of sensor data that ends 0.5\,s before an anchor time $a$, predict whether
hand contact starts at $a$, i.e., within the next $500\,\mathrm{ms}$.
Contact means that the sum of the 50 pressure cells exceeds 5\,g; positive anchors are
contact onsets with no contact during the preceding 1.5\,s, and negative anchors are
contact-free windows from the same recordings at least 2\,s from every onset.
Pressure is never an input.
Headline metric is AUC and Average Precision (AP).

\paragraph{T4: Missing-Modality Robustness.}
The same late-fusion Transformer is trained with or without per-sample random modality dropout
($p = 0.3$) on all four core modalities (MoCap + EMG + EyeTrack + IMU),
and evaluated on eight test-time configurations:
full 4 modalities, all leave-one-out (drop MoCap, drop EMG, drop EyeTrack, drop IMU),
and three singleton motion modalities (MoCap, EMG, IMU; EyeTrack omitted by design).
Probes graceful degradation when modalities drop at deployment-realistic availability pattern.
We report macro F1 for each of the 8 eval configs and accuracy for the full configuration (3 seeds per fold); a five-modality
variant that adds pressure is reported in Table~\ref{tab:missing-mod-5}.

\paragraph{T5: Tactile-driven Grasp State Recognition (TGSR).}
Frame-level 3-class classification of object-interaction state in a
future window $(t, t+T_\mathrm{fut}]$ given only the past observation
window $[t-T_\mathrm{obs}, t]$ at 20\,Hz (downsampled from 100\,Hz), with
$T_\mathrm{obs}=1$\,s, $T_\mathrm{fut}=0.5$\,s, and anchors every 0.25\,s.
Tests whether the tactile (Pressure) channel can resolve proprioceptive ambiguities
in grasp-state discrimination that are invisible to motor-only signals.
The label space separates three semantic object-interaction states:
\textit{NoGrasp} denotes no physical contact with the target object;
\textit{AttemptedGrasp} denotes contact without stable object control
(e.g., touching or briefly grasping without lifting, transport, or sustained manipulation);
and \textit{SustainedGrasp} denotes stable object control
(e.g., successful lifting, holding, or transport).
Operationally, an anchor is NoGrasp unless at least 50\% of its future window lies in
grasp-verb segments; otherwise, future-window contact lasting $<\!0.3$\,s is AttemptedGrasp and
contiguous contact $\geq\!0.3$\,s is SustainedGrasp.
The \textit{Sustained / Attempted} distinction is fundamentally
pressure-only-visible; Pressure is the contact side-channel.
To avoid target leakage, pressure is used offline to derive labels in the
future window, but any pressure input to the model is restricted to the
past observation window and never overlaps the label window.
To isolate pressure's marginal contribution and avoid an
``position proxy'' shortcut (finger pose $\to$ contact), we employ
a leave-target-out cross-modal ablation:
baseline uses \emph{forearm EMG} (motor intent) only; experiment group adds \emph{Pressure} (actual contact) as a side-channel.
A second comparison adds pressure to the full non-tactile stack (Table~\ref{tab:tactile-full}).
For analysis, anchors are further stratified by past/future contact occupancy
into four event types
(non-contact, pre-contact, steady-grip, release).
Headline metric is 3-class macro-F1 reported under volunteer-stratified
5-fold cross-validation; we use 3 backbones (DAF, FUTR, DeepConvLSTM)
$\times$ 5 folds (3 seeds per fold).
Configuration matrix: 3 sequence backbones (DAF, FUTR, DeepConvLSTM)
$\times$ 2 input configurations (EMG only, EMG\,$+$\,Pressure)
$\times$ 5 folds.
We additionally report per-class F1, with SustainedGrasp lift as the primary
class-specific analysis.

\paragraph{T6: Cross-Modal Pressure Prediction.}
From non-tactile modalities (sEMG and/or full-body MoCap with finger articulation),
we regress two granularities of finger-level pressure:
(a) \emph{T6a (scalar grip force)} predicts the right-hand summed pressure
$y_t \in \mathbb{R}^{1}\,\text{(g)}$, matching the right-arm Myo placement and literature setup of EMG/kinematics $\to$ scalar force regression~\cite{ghorbani2023grip,pham2018hand}.
(b) \emph{T6b (50-channel spatial pressure)} regresses full per-cell pressure distribution
$y_t \in \mathbb{R}^{50}\,\text{(g)}$, exploiting PULSE's unique array-level spatial signal.

\textbf{Inputs and protocol.}
Five-modality 100\,Hz data are downsampled to 20\,Hz.
Following literature pre-processing convention, sEMG is replaced by its rectified moving-RMS envelope
(200\,ms window) and all input features are \emph{per-subject z-scored}
to remove subject-specific gain differences, using only a calibration window (the
first 30\,s of the subject's recordings), whose frames are not scored.
We use the same volunteer-stratified 5-fold cross-validation protocol as
the other benchmark tasks (27-volunteer T6 pool with pressure availability),
with pressure-availability filtering applied inside each fold.
Active manipulation segments are extracted by thresholding the total pressure trace
(at 100\,g, dilating by 1.0\,s, adding 1.0\,s of context, and merging overlapping segments)
to filter away long quiescent stretches that dominate naturalistic recordings.
For T6a, the target is the right-hand pressure sum (matching the right-forearm Myo placement);
for T6b, the target is all 50 cells, and loss is weighted by per-cell active rate
so capacity concentrates on cells with non-trivial signal.
Loss is a mixed Huber + (1 - Pearson) on \emph{log1p(g)} targets; metrics are reported in raw grams after $\exp(\cdot) - 1$.

\textbf{Backbones and fusion.} Four sequence backbones are evaluated:
three literature force-regression architectures
(\textbf{Ghorbani 2023} GRU,
 \textbf{Pham 2018} BiLSTM~\cite{ghorbani2023grip,pham2018hand}, and a multi-branch
1-D CNN after Hajian et al.~\cite{hajian2021generalized})
+ a PULSE Transformer baseline.
Beyond naive feature concatenation, we also introduce a \emph{LateFusion} model with
modality-specialist branches (1-D CNN branch for sEMG, Pham-style BiLSTM for hand MoCap),
per-branch projection to a shared hidden dimension, training-time per-sample modality dropout,
and a fused output head---designed to address dimension imbalance
(8-channel sEMG vs $\sim$460-channel hand MoCap pose) that can affect naive concat.

\textbf{Evaluation protocol.} Headline metrics are Pearson correlation $r$ between
predicted and ground-truth pressure on the held-out test split:
$r$ for T6a (scalar) and per-cell $r_{\max}$ for T6b (50-channel; the maximum
Pearson correlation across 50 finger-level cells). Because $r_{\max}$
is an extremum across cells and varies across seeds, we report both the
single-seed mean (averaged over 3 seeds per fold) and a 3-seed prediction-ensemble $r$;
details in
Section~\ref{sec:app_results_t6}.

\paragraph{T7: EMG-to-Hand-Pose Reconstruction.}
From forearm sEMG alone, we reconstruct fine-grained right-hand
kinematics. Given a synchronized EMG sequence, the model predicts a
63-D pose vector at each frame, corresponding to 21 three-dimensional
right-hand joints ($21 \times 3$ coordinates, in the wrist frame) derived from the
finger-level MoCap stream. This task complements T6: whereas T6 asks
whether non-tactile signals can recover contact pressure, T7 asks whether
muscle activity can recover the hand configuration that precedes and
accompanies contact.

\textbf{Inputs and protocol.}
T7 uses the 33-volunteer pool with aligned EMG and right-hand MoCap. The target hand follows the Myo armband side, which is the right forearm for every participant. As in the other benchmark tasks, we use volunteer-stratified 5-fold
cross-validation, with training and test volunteers disjoint in every
fold. The 100\,Hz streams are pooled to 20\,Hz (per-channel EMG RMS over blocks of 5 frames;
the target is the pose at the block center). EMG
inputs are z-scored using training-set statistics. Target hand poses are
normalized using training-set pose statistics, and predictions are
de-normalized before metric computation.

\textbf{Backbones.}
We evaluate four sequence-to-sequence regressors under the same protocol:
NeuroPose~\cite{10.1145/3442381.3449890}, the TDS featurizer of
emg2pose~\cite{salter2024emg2pose}, a multi-scale CNN followed by a Transformer
encoder (MS-CNN+Transformer), and a generic Transformer encoder; as a reference,
we also report the training-set mean pose predicted at every frame. All models use the same input features, target
representation, normalization, train/test folds, and evaluation code;
they differ only in the temporal feature extractor.

\textbf{Evaluation protocol.}
Models are trained with masked MPJPE loss in de-normalized physical
coordinates, with padding frames ignored and fingertip joints assigned
higher loss weight. Headline metrics are MPJPE over the 21 predicted hand
joints, fingertip MPJPE over the five fingertip joints, and bone angle
error over the 20 hand-skeleton bones. MPJPE and fingertip MPJPE are
reported in millimeters; angle error is reported in degrees. Results are
reported as mean\,$\pm$\,std across the five volunteer-disjoint folds, each
fold averaged over 3 seeds.

\section{Baselines and Results}
\label{sec:results}

\paragraph{Baselines.}
For each of the benchmark tasks defined in Section~\ref{sec:tasks} we evaluate three sequence
backbones (1D~CNN, BiLSTM, Transformer) combined with nine multi-modal fusion strategies
(early, late, weighted/gated late, attention, stacking, product/mixture-of-experts, and
feature concatenation). To position our dataset against the broader
community we further evaluate published baselines on T1, T2, T5, T6, and T7.
All architectures, hyperparameters, and training details are documented in
Appendices~\ref{sec:app_fusion}--\ref{sec:app_training} and our released code.
Unless otherwise stated, we use the Transformer backbone and report macro F1
as the primary metric. All headline experiments are evaluated
under volunteer-stratified 5-fold cross-validation (Section~\ref{sec:dataset});
results are mean $\pm$ std across the 5 fold means with 3 seeds per fold (Tables~\ref{tab:s_backbone}
and~\ref{tab:s_fusion}: 1 seed), with model selection on inner-validation volunteers.
Paired differences $\Delta$ are computed on the test recordings both compared rows share;
$p$ is a paired $t$-test over the five fold means, and we call $p \geq 0.05$ not significant.

\subsection{Activity-Scenario Recognition}
\label{subsec:scene-results}

\paragraph{Backbone comparison.}
We compare CNN1D, BiLSTM, and Transformer backbones across all single modalities
(Appendix Table~\ref{tab:s_backbone}). The Transformer achieves the best F1 on IMU (0.332)
and EMG (0.202), and BiLSTM on MoCap (0.449); BiLSTM outperforms CNN1D on IMU and MoCap
but not on EMG.

\paragraph{Fusion strategy comparison.}
We evaluate nine fusion strategies on the IMU$+$MoCap$+$EMG combination
(Appendix Table~\ref{tab:s_fusion}). \textbf{Late-family fusion} achieves the best F1 (weighted
late 0.438, late 0.430), ahead of Attention (0.353) and Feature Concatenation (0.351).
Early fusion performs worst (F1$=$0.337), but no strategy differs significantly from
late fusion (paired $p \geq 0.158$).

\paragraph{Comparison against published baselines and our SyncFuse.}
Table~\ref{tab:scene-published-ext} compares all evaluated methods---spanning
\emph{all modality families} in our benchmark: MoCap skeleton
(ST-GCN~\cite{yan2018stgcn}, CTR-GCN~\cite{chen2021ctrgcn}),
IMU (LIMU-BERT~\cite{xu2021limubert}),
sEMG (CNN baseline from the sEMG gesture-recognition literature),
and multi-modal fusion (ActionSense LSTM baseline~\cite{delpreto2022actionsense},
MulT~\cite{tsai2019multimodal}, Perceiver IO~\cite{jaegle2022perceiverio}),
plus a plain per-modality Transformer with late fusion trained from scratch
(\textit{Transformer+Late})---under
a single \emph{unified} volunteer-stratified 5-fold cross-validation protocol
(33-volunteer pool, 3 seeds per fold, mean $\pm$ std across fold means). All methods
are reproduced with their original architectures; modalities reflect each
paper's primary design (ST-GCN/CTR-GCN/LIMU-BERT/EMG-CNN each use only the
modality their original paper targets); Table~\ref{tab:t1-matched} compares methods on identical inputs.
We additionally propose \textbf{SyncFuse}, a simple modality-late-fusion
architecture tailored to our dataset. Each of the five modalities
(MoCap, EMG, EyeTrack, IMU, Pressure) is processed by an independent
two-layer MLP $\to$ BiGRU stem with LayerNorm; the resulting per-frame
features are concatenated across modalities and passed through a 2-layer
BiLSTM, with attention pooling over time replacing simple mean pooling.
Two regularisers stabilise training under our small-subject regime:
(i) \emph{per-sample modality dropout} during training ($p = 0.3$), which
randomly zeroes entire modality streams to enforce graceful degradation, and
(ii) layer normalisation between stem and fusion. SyncFuse uses 3.1M
parameters (2.6M with four modalities) and adds Pressure
as a fifth modality (the other published baselines consume four modalities
or fewer), so it is evaluated on the 192 test recordings that have all five.

\begin{table}[t]
  \centering
  \caption{Comprehensive comparison on T1 activity-scenario recognition under a
    \emph{volunteer-stratified 5-fold cross-validation} protocol
    (33-volunteer pool, 3 seeds per fold, Macro F1 on 8 scenarios; mean
    $\pm$ std across fold means). All methods reproduced with their original
    architectures; modalities reflect each paper's primary design. Our
    Among the methods in this table, SyncFuse attains the highest mean Macro F1 but is not
    significantly better than CTR-GCN (paired $p = 0.849$), which has a slightly higher mean accuracy.}
  \label{tab:scene-published-ext}
  \footnotesize
  \setlength{\tabcolsep}{3pt}
  \resizebox{\textwidth}{!}{%
  \begin{tabular}{@{}l l l c c r@{}}
    \toprule
    Method & Year & Modalities & Macro F1 & Accuracy & Params \\
    \midrule
    EMG-CNN (sEMG literature)                            & 2021 & EMG only                  & $0.149 \pm 0.028$ & $0.194 \pm 0.027$ & 0.1M \\
    LIMU-BERT~\cite{xu2021limubert}                      & 2021 & IMU only                  & $0.294 \pm 0.063$ & $0.321 \pm 0.049$ & 1.9M \\
    Perceiver IO~\cite{jaegle2022perceiverio}            & 2022 & 4-mod                     & $0.339 \pm 0.079$ & $0.366 \pm 0.069$ & 2.0M \\
    ST-GCN~\cite{yan2018stgcn}                           & 2018 & MoCap skeleton            & $0.391 \pm 0.029$ & $0.432 \pm 0.044$ & 7.0M \\
    Transformer+Late (ours)                              & 2026 & 4-mod                     & $0.392 \pm 0.076$ & $0.417 \pm 0.075$ & 5.9M \\
    ActionSense (LSTM)~\cite{delpreto2022actionsense}    & 2022 & 4-mod (MoCap+EMG+Eye+IMU) & $0.396 \pm 0.029$ & $0.427 \pm 0.049$ & 1.2M \\
    MulT~\cite{tsai2019multimodal}                       & 2019 & 3-mod (MoCap+EMG+IMU)     & $0.410 \pm 0.069$ & $0.423 \pm 0.065$ & 3.9M \\
    CTR-GCN~\cite{chen2021ctrgcn}                        & 2021 & MoCap skeleton            & $0.434 \pm 0.099$ & $0.466 \pm 0.100$ & 3.8M \\
    \midrule
    \textbf{SyncFuse (Ours)}                             & \textbf{2026} & \textbf{5-mod (incl.~Pressure)} & $0.440 \pm 0.063$ & $0.463 \pm 0.071$ & 3.1M \\
    \bottomrule
  \end{tabular}%
  }
\end{table}

\begin{figure}[t]
  \centering
  \includegraphics[width=0.82\linewidth]{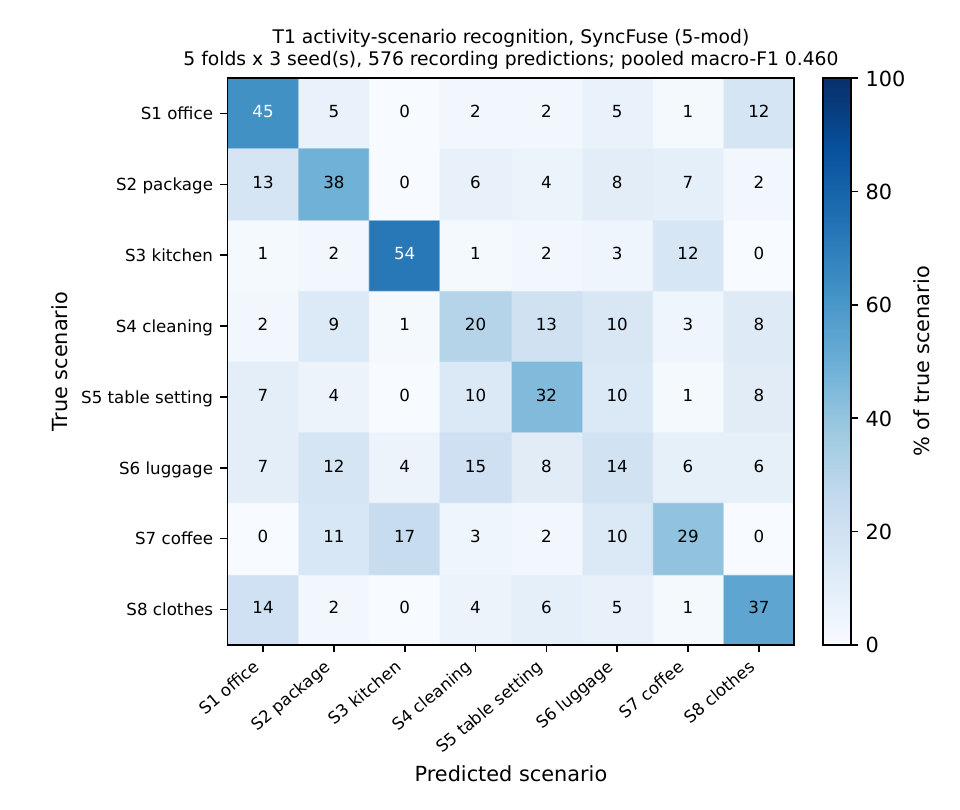}
  \caption{T1 activity-scenario confusion matrix for SyncFuse, aggregated
    across the 5 cross-validation folds $\times$ 3 seeds in
    Table~\ref{tab:scene-published-ext} (576 predictions). Cell values are raw counts and
    colors are row-normalized percentages. The matrix shows that most
    residual errors are structured confusions between scenarios with similar
    object-interaction patterns, rather than uniform mistakes. The macro-F1 printed in the
    figure (0.460) is pooled over all 576 predictions; Table~\ref{tab:scene-published-ext}
    reports the mean over fold means (0.440).}
  \label{fig:t1-scene-confusion}
\end{figure}

\paragraph{Discussion.}
A few observations:
(i) Single-modality baselines split sharply: EMG-only ($0.149$) and
LIMU-BERT (IMU, $0.294$) lag the MoCap-skeleton methods (ST-GCN $0.391$,
CTR-GCN $0.434$), which themselves rival the strongest multi-modal fusion
methods (ActionSense $0.396$, MulT $0.410$). Modality \emph{which} matters
more than modality \emph{count} at this scale: scenario-discriminative
information is concentrated in body motion and object-interaction
signatures rather than spread uniformly across all five modalities.
(ii) Naive multi-modal fusion is not universally beneficial: Perceiver IO
($0.339$, 4-mod) underperforms even single-modality MoCap baselines,
reinforcing the dataset's broader observation that fusion architectures
must be designed for small-subject regimes rather than borrowed wholesale
from large-scale benchmarks.
(iii) SyncFuse is not significantly better than the strongest published baseline
(CTR-GCN, $p = 0.849$), MulT, or Transformer+Late, and under matched inputs not than
ActionSense (Table~\ref{tab:t1-matched}). We therefore do not attribute a
SyncFuse advantage to its architecture; the only ingredient with measurable support is
the added pressure modality, which raises SyncFuse Macro F1 by $+0.053$ on the same
recordings ($p = 0.034$). A differently configured four-modality late-fusion
Transformer, the T4 model of Table~\ref{tab:missing-mod} (1.1M parameters, selection on
inner-validation loss), reaches $0.460$ ($0.480$ with modality-dropout training) on its 235
test recordings, above the Transformer+Late row on the same recordings ($0.392$,
$p = 0.019$); on the 192 recordings with pressure it reaches $0.415$ vs.\ $0.440$ for
SyncFuse (n.s.). T1 scores are therefore as sensitive to model size and selection rule as
to architecture, and no T1 model is clearly ahead.

\paragraph{Matched inputs and pretrained encoders.}
Table~\ref{tab:t1-matched} gives the published methods identical inputs, and
Table~\ref{tab:t1-harnet} evaluates the self-supervised harnet5
encoder~\cite{yuan2024harnet} on the right-wrist accelerometer.

\begin{table}[h]
  \centering
  \caption{T1 activity-scenario recognition under matched inputs (pool33). Within
    each block all methods receive the same modalities; $\Delta$ is the Macro-F1
    difference to the first row of the block (ref.), with the paired $p$ over five
    folds. The last block scores the 4-modality and 5-modality versions of the same
    model on the same 192 test recordings (26 volunteers), so
    $\Delta$ is the gain from adding finger-level pressure. Results with each
    method's original modalities are in Table~\ref{tab:scene-published-ext}.}
  \label{tab:t1-matched}
  \footnotesize
  \setlength{\tabcolsep}{4pt}
  \begin{tabular}{@{}l l c c c@{}}
    \toprule
    Inputs & Method & Macro-F1 & Accuracy & $\Delta$ Macro-F1 ($p$) \\
    \midrule
    IMU only            & LIMU-BERT~\cite{xu2021limubert}         & 0.294 $\pm$ 0.063 & 0.321 $\pm$ 0.049 & ref. \\
                        & TinyHAR~\cite{zhou2022tinyhar}          & 0.316 $\pm$ 0.073 & 0.340 $\pm$ 0.063 & $+0.022$ (0.132) \\
                        & InceptionTime~\cite{fawaz2020inceptiontime} & 0.331 $\pm$ 0.067 & 0.346 $\pm$ 0.060 & $+0.036$ (0.427) \\
                        & Transformer                             & 0.320 $\pm$ 0.055 & 0.342 $\pm$ 0.050 & $+0.025$ (0.141) \\
                        & harnet5, fine-tuned$^{\dagger}$~\cite{yuan2024harnet} & 0.374 $\pm$ 0.033 & 0.424 $\pm$ 0.028 & $+0.080$ (0.030) \\
    \midrule
    MoCap+EMG           & SyncFuse (ours)                         & 0.425 $\pm$ 0.063 & 0.457 $\pm$ 0.068 & ref. \\
    +Eye+IMU            & ActionSense~\cite{delpreto2022actionsense} & 0.396 $\pm$ 0.029 & 0.427 $\pm$ 0.049 & $-0.029$ (0.209) \\
                        & Transformer+Late                        & 0.392 $\pm$ 0.076 & 0.417 $\pm$ 0.075 & $-0.033$ (0.324) \\
                        & Perceiver IO~\cite{jaegle2022perceiverio} & 0.339 $\pm$ 0.079 & 0.366 $\pm$ 0.069 & $-0.086$ (0.023) \\
    \midrule
    All five            & SyncFuse (ours)                         & 0.440 $\pm$ 0.063 & 0.463 $\pm$ 0.071 & ref. \\
                        & ActionSense                             & 0.426 $\pm$ 0.081 & 0.461 $\pm$ 0.083 & $-0.014$ (0.565) \\
                        & Perceiver IO                            & 0.316 $\pm$ 0.065 & 0.352 $\pm$ 0.072 & $-0.124$ (0.015) \\
    \midrule
    4-mod $\to$ 5-mod   & SyncFuse                                & 0.387 $\to$ 0.440 & 0.427 $\to$ 0.463 & $\mathbf{+0.053}$ (\textbf{0.034}) \\
    (same recordings)   & ActionSense                             & 0.382 $\to$ 0.426 & 0.416 $\to$ 0.461 & $+0.043$ (0.083) \\
    \bottomrule
  \end{tabular}

  \vspace{0.2em}
  {\scriptsize $^{\dagger}$Right-wrist accelerometer only, 5\,s windows; recording prediction = mean of window logits (Appendix Table~\ref{tab:t1-harnet}).}
\end{table}

\begin{table}[h]
  \centering
  \caption{T1 on the right-wrist accelerometer (pool33, 248
    test recordings; 5 folds $\times$ 3 seeds; mean $\pm$ SD over fold means).
    harnet5~\cite{yuan2024harnet}: pretrained checkpoint (OxWearables
    \texttt{ssl-wearables}) fine-tuned, or the same architecture from random
    initialization; AdamW, lr $10^{-3}$, batch 64, at most 80 epochs, patience 15 on
    the inner-validation volunteers. Window: 5\,s windows; recording: mean of window
    logits.}
  \label{tab:t1-harnet}
  \footnotesize
  \setlength{\tabcolsep}{4pt}
  \begin{tabular}{@{}l r c c c c@{}}
    \toprule
    & & \multicolumn{2}{c}{Recording level} & \multicolumn{2}{c}{5\,s window level} \\
    \cmidrule(lr){3-4}\cmidrule(lr){5-6}
    Method & Params & Macro-F1 & Accuracy & Macro-F1 & Accuracy \\
    \midrule
    harnet5, fine-tuned     & 4.2M & 0.374 $\pm$ 0.033 & 0.424 $\pm$ 0.028 & 0.252 $\pm$ 0.017 & 0.299 $\pm$ 0.019 \\
    harnet5, random init    & 4.2M & 0.330 $\pm$ 0.080 & 0.394 $\pm$ 0.072 & 0.229 $\pm$ 0.028 & 0.295 $\pm$ 0.025 \\
    TinyHAR~\cite{zhou2022tinyhar} & 0.2M & 0.378 $\pm$ 0.049 & 0.434 $\pm$ 0.043 & 0.254 $\pm$ 0.014 & 0.295 $\pm$ 0.018 \\
    \bottomrule
  \end{tabular}
\end{table}

\subsection{Action Recognition}
\label{subsec:action-results}

T2 evaluates action recognition on 4\,s multi-modal sensor windows at 1\,s stride, each
labeled by the annotated segment that contains its center; models predict
a four-tuple of labels (17-class verb fine, 6-class verb composite, 33-class noun, 3-class hand).
All models are evaluated under \emph{volunteer-stratified 5-fold cross-validation}
on a 27-volunteer pool (after modality-availability and noun filtering); within each
fold, modality-availability and noun filtering are applied to the
recordings of the held-out volunteers.

We additionally introduce \textbf{DailyActFormer (DAF)}, a multi-modal action
classifier built on top of HandFormer's multi-scale conv backbone with four
modifications targeting cross-subject generalisation:
(i) \emph{per-modality} multi-scale 1-D conv stems (kernels $\{3,5,9\}$) so the
8-channel EMG stream is not drowned by the high-dimensional MoCap features,
(ii) whole-modality dropout ($p=0.3$, drawn once per batch) during training,
(iii) concat$\to$$1{\times}1$-conv fusion across modalities (instead of
compressing the $M$ modality streams with attention), and
(iv) a 3-layer Transformer with learnable-query pooling over time, feeding a
joint verb--noun head ($17 \times 33$).
We evaluate DAF against seven published baselines spanning all design families
relevant to action recognition on sensor data: DeepConvLSTM (IMU and 3-modality)~\cite{ordonez2016deep},
RU-LSTM (rolling/unrolling LSTM)~\cite{furnari2019rulstm}, FUTR (Transformer encoder
with cross-attention pool)~\cite{gong2022futr}, AFFT (token-interleaved cross-modal
Transformer)~\cite{zhong2023afft}, HandFormer (multi-scale conv on
MoCap)~\cite{shamil2024handformer}, and a from-scratch ActionLLM
surrogate~\cite{wang2025actionllm} (a convolutional stem and a 2-layer Transformer; no
language model).

All eight models are trained under a unified protocol: AdamW, lr $3\mathrm{e}{-4}$ with
5-epoch linear warmup and cosine decay over at most 40 epochs, weight decay $1\mathrm{e}{-4}$, label smoothing
$0.05$, no class weighting, patience 25 on inner-validation $\mathrm{act\_vn@5}$, batch size 32, 3 seeds per fold. Each baseline uses
the modality subset recommended by its original publication; DAF uses all five modalities.

\begin{table}[t]
  \centering
  \caption{Action recognition (T2) under volunteer-stratified 5-fold
    cross-validation (27-volunteer pool, 3 seeds per fold), mean $\pm$ std
    across fold means. Headline metric is \emph{joint top-5 action accuracy}
    $\mathrm{act\_vn@5}$---the probability that the verb-fine and noun
    ground-truth labels both fall within the model's top-5 predictions on
    each head---reported because the trainer logs only top-1 and top-5;
    \texttt{verb\_fine@5} and \texttt{noun@5} report per-head top-5 accuracy.
    Class counts: 17 verbs, 33 nouns; 17$\times$33$=$561 joint cells, chance
    $\mathrm{act\_vn@5}\!\approx\!25/561\!=\!4.5\%$.}
  \label{tab:action-recognition}
  \footnotesize
  \setlength{\tabcolsep}{3pt}
  \resizebox{\textwidth}{!}{%
  \begin{tabular}{@{}llllccc@{}}
    \toprule
    \# & Method & Year & Modalities & act\_vn@5 $\uparrow$ & verb\_fine@5 $\uparrow$ & noun@5 $\uparrow$ \\
    \midrule
    8 & DeepConvLSTM (IMU)~\cite{ordonez2016deep}              & 2016 & IMU                   & $0.343 \pm 0.034$ & $0.701 \pm 0.018$ & $0.476 \pm 0.044$ \\
    7 & ActionLLM (surrogate)~\cite{wang2025actionllm}         & 2025 & IMU+EMG+Eye           & $0.366 \pm 0.022$ & $0.701 \pm 0.016$ & $0.501 \pm 0.034$ \\
    6 & RU-LSTM~\cite{furnari2019rulstm}                       & 2019 & IMU+MoCap             & $0.408 \pm 0.033$ & $0.730 \pm 0.019$ & $0.526 \pm 0.055$ \\
    5 & DeepConvLSTM (3 mod)~\cite{ordonez2016deep}            & 2016 & IMU+MoCap+EMG         & $0.412 \pm 0.052$ & $0.731 \pm 0.029$ & $0.520 \pm 0.069$ \\
    4 & FUTR~\cite{gong2022futr}                               & 2022 & MoCap+IMU+EMG         & $0.423 \pm 0.048$ & $0.734 \pm 0.026$ & $0.550 \pm 0.070$ \\
    3 & AFFT~\cite{zhong2023afft}                              & 2023 & IMU+EMG+Eye+MoCap     & $0.430 \pm 0.017$ & $0.734 \pm 0.015$ & $0.553 \pm 0.037$ \\
    2 & HandFormer~\cite{shamil2024handformer}                 & 2024 & MoCap                 & $0.463 \pm 0.057$ & $0.755 \pm 0.023$ & $0.582 \pm 0.072$ \\
    1 & \textbf{DailyActFormer (Ours)} & \textbf{2026}  & IMU+EMG+Eye+MoCap+Pressure & $\mathbf{0.468 \pm 0.026}$ & $\mathbf{0.762 \pm 0.013}$ & $\mathbf{0.597 \pm 0.034}$ \\
    \bottomrule
  \end{tabular}}
\end{table}

\paragraph{Findings.}
\textbf{(i)} DAF achieves the highest mean on every reported metric---joint
$\mathrm{act\_vn@5}$ as well as per-head verb\_fine@5 and noun@5.
\textbf{(ii)} DAF's margin over HandFormer is small and not significant
(act\_vn@5 difference $0.003$ on the shared test recordings, $p = 0.892$), so we treat
DAF as a strong baseline rather than a decisive model contribution; DAF is significantly
better than the six other baselines, including FUTR ($+4.3$\,pp, $p = 0.039$).
\textbf{(iii)} Modality access matters more than model family below the top two:
AFFT (4-mod) and FUTR (3-mod) are tightly clustered, and the two rows without MoCap
(DeepConvLSTM on IMU and ActionLLM) trail.
\textbf{(iv)} Per-head verb\_composite (6 classes) and hand (3 classes) are not headline
metrics: verb-composite top-1 accuracy is clearly above the training-majority class
(DAF $0.388$ vs.\ $0.237$), whereas the hand head is closer to it ($0.655$ vs.\ $0.597$);
full per-head numbers are in Appendix~\ref{sec:app_action_full}.

\paragraph{Additional T2 protocols.}
Table~\ref{tab:action-seg} reports the segment-aligned protocol (one sample per annotated
segment), and Table~\ref{tab:t2-matched} gives nine architectures identical inputs; in
neither is any architecture significantly better than DAF.

\begin{table}[h]
  \centering
  \caption{T2 action recognition, secondary segment-aligned protocol (one sample per
    annotated segment; same models, folds, seeds, and recordings as
    Table~\ref{tab:action-recognition}). $\Delta$: $\mathrm{act\_vn@5}$ difference to
    DAF on the shared test recordings, with the paired $p$ over five folds.}
  \label{tab:action-seg}
  \footnotesize
  \setlength{\tabcolsep}{3pt}
  \resizebox{\textwidth}{!}{%
  \begin{tabular}{@{}clcccc@{}}
    \toprule
    \# & Method & act\_vn@5 $\uparrow$ & verb\_fine@5 $\uparrow$ & noun@5 $\uparrow$ & $\Delta$ act\_vn@5 ($p$) \\
    \midrule
    8 & DeepConvLSTM (IMU)      & 0.305 $\pm$ 0.033 & 0.751 $\pm$ 0.019 & 0.406 $\pm$ 0.038 & $-0.136$ (0.0002) \\
    7 & ActionLLM (surrogate)   & 0.340 $\pm$ 0.045 & 0.762 $\pm$ 0.021 & 0.436 $\pm$ 0.051 & $-0.098$ (0.005) \\
    6 & RU-LSTM                 & 0.385 $\pm$ 0.045 & 0.782 $\pm$ 0.021 & 0.484 $\pm$ 0.054 & $-0.057$ (0.004) \\
    5 & DeepConvLSTM (3 mod)    & 0.397 $\pm$ 0.048 & 0.776 $\pm$ 0.028 & 0.490 $\pm$ 0.054 & $-0.045$ (0.012) \\
    4 & FUTR                    & 0.403 $\pm$ 0.042 & 0.796 $\pm$ 0.021 & 0.500 $\pm$ 0.050 & $-0.038$ (0.010) \\
    3 & AFFT                    & 0.412 $\pm$ 0.024 & 0.803 $\pm$ 0.022 & 0.497 $\pm$ 0.026 & $-0.031$ (0.065) \\
    2 & HandFormer              & 0.426 $\pm$ 0.046 & 0.798 $\pm$ 0.021 & 0.519 $\pm$ 0.053 & $-0.014$ (0.351) \\
    1 & DAF (ours)              & 0.443 $\pm$ 0.030 & 0.807 $\pm$ 0.019 & 0.538 $\pm$ 0.040 & ref. \\
    \bottomrule
  \end{tabular}}
\end{table}

\begin{table}[h]
  \centering
  \caption{T2 under matched inputs (sliding windows; pool27; 5 folds $\times$ 3 seeds).
    All rows receive the same modalities: all five (192 test
    recordings) or IMU only (198). $\Delta$:
    $\mathrm{act\_vn@5}$ difference to DAF with the same inputs, paired $p$ over five
    folds. Params are for the all-five input.}
  \label{tab:t2-matched}
  \footnotesize
  \setlength{\tabcolsep}{3pt}
  \resizebox{\textwidth}{!}{%
  \begin{tabular}{@{}l r c c c c@{}}
    \toprule
    & & \multicolumn{2}{c}{All five modalities} & \multicolumn{2}{c}{IMU only} \\
    \cmidrule(lr){3-4}\cmidrule(lr){5-6}
    Method & Params & act\_vn@5 & $\Delta$ ($p$) & act\_vn@5 & $\Delta$ ($p$) \\
    \midrule
    DeepConvLSTM~\cite{ordonez2016deep} & 0.6M & 0.412 $\pm$ 0.038 & $-0.056$ (0.001) & 0.343 $\pm$ 0.034 & $-0.024$ (0.017) \\
    RU-LSTM~\cite{furnari2019rulstm}    & 1.5M & 0.426 $\pm$ 0.030 & $-0.042$ (0.026) & 0.348 $\pm$ 0.033 & $-0.020$ (0.008) \\
    FUTR~\cite{gong2022futr}            & 0.9M & 0.432 $\pm$ 0.048 & $-0.036$ (0.067) & 0.362 $\pm$ 0.026 & $-0.006$ (0.184) \\
    AFFT~\cite{zhong2023afft}           & 0.5M & 0.439 $\pm$ 0.021 & $-0.029$ (0.008) & 0.354 $\pm$ 0.035 & $-0.013$ (0.104) \\
    TinyHAR~\cite{zhou2022tinyhar}      & 0.9M & 0.439 $\pm$ 0.046 & $-0.029$ (0.041) & 0.359 $\pm$ 0.042 & $-0.009$ (0.370) \\
    InceptionTime~\cite{fawaz2020inceptiontime} & 0.7M & 0.450 $\pm$ 0.039 & $-0.018$ (0.120) & 0.362 $\pm$ 0.035 & $-0.005$ (0.531) \\
    ActionLLM (surrogate)~\cite{wang2025actionllm} & 2.0M & 0.453 $\pm$ 0.046 & $-0.015$ (0.241) & 0.377 $\pm$ 0.037 & $+0.010$ (0.190) \\
    HandFormer~\cite{shamil2024handformer} & 2.6M & 0.458 $\pm$ 0.042 & $-0.010$ (0.443) & 0.359 $\pm$ 0.035 & $-0.008$ (0.179) \\
    DAF (ours)                          & 3.0M & 0.468 $\pm$ 0.026 & ref. & 0.367 $\pm$ 0.025 & ref. \\
    \bottomrule
  \end{tabular}}
\end{table}

\paragraph{Motion-primitive pretraining transfer (F5).}
\label{para:s9_t2_transfer}
The S9 isolated motion-primitive recordings (33 volunteers, $\sim$2.8\,h)
provide a complementary supervised signal aligned with T2's verb axis.
We test whether pretraining a DAF encoder on the motion-primitive labels
and then fine-tuning on T2 yields a measurable prior, and how any such
prior scales with the amount of T2 supervision available downstream. The
motion-primitive recordings were collected with EMG and IMU only, so we
run this analysis with a 2-modality DAF (EMG\,$+$\,IMU) on the T2 task;
this matches the encoder modalities exactly between motion-primitive
pretraining and T2 fine-tuning, ensuring the pretrained weights cover
100\% of the downstream encoder. The pretraining pool is restricted to
the seven volunteers (\emph{v9, v16--v21}) whose motion-primitive
recordings are disjoint from all T2 cross-validation folds, eliminating
subject leakage; one of these recordings (\emph{v20}) fails the IMU
quality check and is dropped, leaving six recordings and 90 labelled
segments (72 for training, 18 held out for epoch selection). Pretraining
optimises a joint verb $+$ noun head over the motion-primitive labels.
After pretraining, the encoder is loaded into a fresh DAF and the network
is fine-tuned on T2 with the encoder learning rate scaled by $0.1\times$
relative to the heads. The downstream metric is
$\mathrm{verb\_composite@1}$ (6-class verb rollup), the granularity at
which motion-primitive labels and T2 verb labels share semantic
structure. We sweep the T2 training fraction over
$\{5\%, 10\%, 25\%, 50\%, 100\%\}$ under the same volunteer-stratified
5-fold CV protocol used elsewhere, with three seeds (42, 43, 44) per fold.

\begin{table}[h]
  \centering
  \caption{Motion-primitive pretraining transfer to T2 under 5-fold CV
    (DAF with EMG\,$+$\,IMU; mean $\pm$ std over the five fold means,
    three seeds per fold). The fine-tuned model is statistically
    indistinguishable from the scratch baseline: all $|\Delta| \leq 0.008$
    and no fraction reaches significance in a paired $t$-test over folds
    ($p \geq 0.15$). The pretraining set (90 segments) is also smaller than
    the T2 training set at every fraction (157--183 segments per fold at 5\%), so we read the probe as
    inconclusive at this scale rather than as evidence against the
    transfer direction.}
  \label{tab:s9-pretrain-t2}
  \footnotesize
  \setlength{\tabcolsep}{6pt}
  \resizebox{\textwidth}{!}{%
  \begin{tabular}{@{}lcccc@{}}
    \toprule
    T2 train fraction & Baseline verb\_comp@1 & + Motion-primitive pretrain & $\Delta$ & $p$ (paired) \\
    \midrule
    5\%   & $0.257 \pm 0.015$ & $0.265 \pm 0.012$ & $+0.008$ & $0.26$ \\
    10\%  & $0.284 \pm 0.016$ & $0.280 \pm 0.008$ & $-0.004$ & $0.61$ \\
    25\%           & $0.295 \pm 0.012$ & $0.295 \pm 0.013$ & $0.000$  & $0.93$ \\
    50\%           & $0.309 \pm 0.027$ & $0.317 \pm 0.019$ & $+0.008$ & $0.15$ \\
    100\%          & $0.317 \pm 0.019$ & $0.318 \pm 0.028$ & $+0.001$ & $0.91$ \\
    \bottomrule
  \end{tabular}%
  }
\end{table}

\begin{figure}[h]
  \centering
  \includegraphics[width=0.78\linewidth]{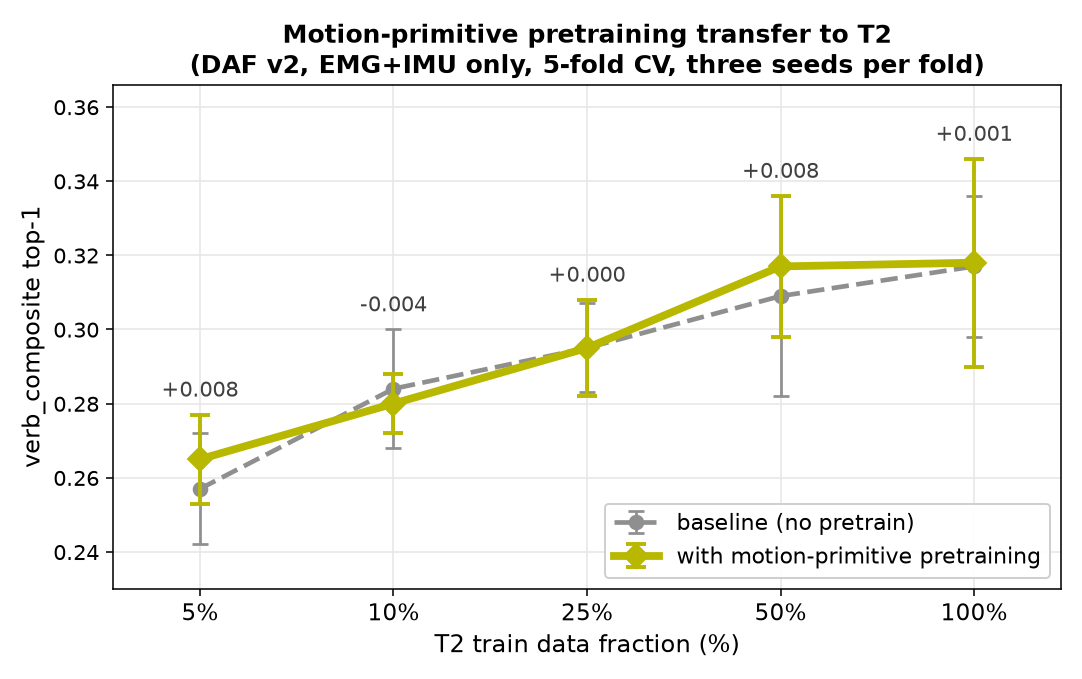}
  \caption{Motion-primitive pretraining transfer on T2 as a function of
    downstream supervision (DAF, EMG\,$+$\,IMU, 5-fold CV, three seeds
    per fold; same runs as Table~\ref{tab:s9-pretrain-t2}). Error bars are
    1\,std across the five fold means. The two curves coincide at every
    fraction and the paired difference stays within the seed-to-seed
    spread, so the probe does not resolve a pretraining benefit at this
    scale.}
  \label{fig:s9-pretrain-t2}
\end{figure}

With the reviewed labels and three seeds per fold, the motion-primitive
prior does not change T2 transfer measurably: the paired difference is at
most $+0.008$ verb\_composite@1 and stays within the seed-to-seed spread
at every fraction (paired $t$-test over folds, $p \geq 0.15$). We
therefore report F5 as a secondary, inconclusive probe rather than as a
positive transfer result; the headline T2 number in
Table~\ref{tab:action-recognition} is trained on the full 100\% T2 data
without pretraining, and no other result in this paper depends on the
pretraining arm. The S9 recordings remain useful in their own right as a
densely annotated set of isolated primitives, and we report the F5
outcome for transparency rather than as a general claim about primitive
transfer.

\subsection{Cross-Task Analysis}
\label{subsec:cross-task}

Several patterns emerge across the benchmark tasks that collectively characterize the
design space.

\paragraph{Modality utility is task-dependent.}
MoCap is the strongest single modality for activity-scenario recognition (T4 singletons under
dropout: MoCap F1$=$0.394, IMU 0.318, EMG 0.153), EMG and MoCap alone are comparably
predictive for grasp anticipation (T3 AUC $=0.696$ and $0.686$), and finger-level
pressure adds information for grasp-state recognition (T5) and, for SyncFuse, activity-scenario recognition (T1; $+0.053$, $p = 0.034$; not significant for ActionSense or the T4 late-fusion model). This diversity motivates the
multi-task, multi-modal design of the benchmark: no single modality is
universally optimal, and meaningful baselines require evaluating each
combination per task.

\paragraph{Modality count is not monotonic.}
Adding modalities helps non-monotonically. On T3 the full 4-modality
configuration has the highest mean AUC ($+4.6$\,pp over EMG-only, not significant), but the
intermediate EMG+IMU input is significantly worse than EMG alone. On T1 single-modality
skeleton methods (CTR-GCN $0.434$) match strong 4-modality fusion
baselines (ActionSense $0.396$). On T6b, concatenating EMG with hand MoCap raises the
mean $r_{\max}$ of the Pham BiLSTM, but the four T6b models do not differ significantly. Which modalities are added matters more than how many,
and small-subject data regimes amplify these design sensitivities.

\paragraph{Cross-fold variance is substantial.}
Cross-fold standard deviations reach $0.06$--$0.10$ for the strongest
T1 methods (SyncFuse $\pm 0.063$, CTR-GCN $\pm 0.099$), representing
$14$--$23\%$ of the mean. This level of variance underscores the
importance of cross-validated evaluation on small-subject datasets and
cautions against single-split comparisons, particularly for methods
reporting improvements of a few percentage points.

\subsection{Case Study: Anticipatory Coordination in Grasp Events}
\label{sec:case-study}
\label{sec:grasp-phase-main}
\label{sec:app_grasp_phase}

To illustrate the class of research questions that PULSE is positioned to support,
we present a short analysis of \emph{anticipatory sensorimotor coordination} around
grasping: the well-documented phenomenon whereby gaze, muscle activation, and hand
motion precede physical contact in a stereotyped temporal order~\cite{johansson1984roles,
land1999roles}. Recovering this ordering requires muscle, motion, and contact signals to
share a single high-frequency timebase, and therefore cannot be addressed by
video-only or by unsynchronised multi-sensor datasets.

\paragraph{Event detection and filtering.}
We define a \emph{grasp onset} as the moment when the summed pressure of the 25
right-hand cells (the Myo is on the right forearm) crosses $5$\,g
from below. From the initial set of $4{,}384$ grasp onsets detected across 198
recordings, we retain only ``clean'' grasps that transition from a quiescent pre-grasp
period: specifically, events where both EMG envelope and hand speed are low in a rest
window and rise clearly before contact, both onsets lie within 1.5\,s before contact, and
the 2\,s before contact contain no untracked MoCap frames. This filtering yields $458$
high-quality events from 160 recordings of 26 volunteers. For each retained event we
align the EMG envelope, right-hand speed, and pressure trace to the contact onset
($t=0$) and average across events.

\begin{figure}[t]
  \centering
  \includegraphics[width=0.78\linewidth]{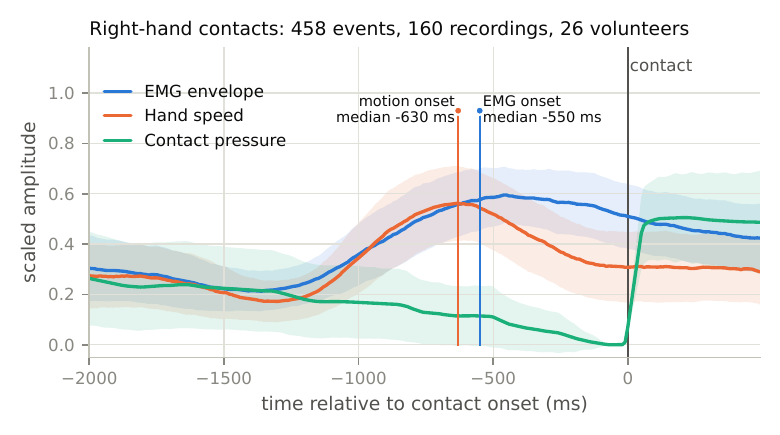}
  \caption{\textbf{Anticipatory coordination revealed by 100\,Hz synchronised recording.}
    EMG envelope, hand speed, and pressure signals averaged across 458 clean grasp
    events from 26 volunteers, aligned to the pressure contact onset ($t=0$); shaded bands are
    $\pm0.5\,\sigma$. Both EMG activation and hand motion precede contact by
    approximately half a second---an anticipatory temporal structure consistent with
    Johansson \& Westling~\cite{johansson1984roles}, observable here because the three
    modalities share a common 100\,Hz timebase.}
  \label{fig:grasp-phase}
\end{figure}

\paragraph{Finding.}
Figure~\ref{fig:grasp-phase} shows that EMG envelope and hand velocity both begin to
rise approximately $0.55$--$0.63$\,s before physical contact---consistent with the classical anticipatory
force-control picture~\cite{johansson1984roles}. The two onsets are close
(median paired difference $-35$\,ms), and their order varies from event to event:
motion comes first in 52\% of events.

\begin{figure}[h]
  \centering
  \includegraphics[width=0.95\linewidth]{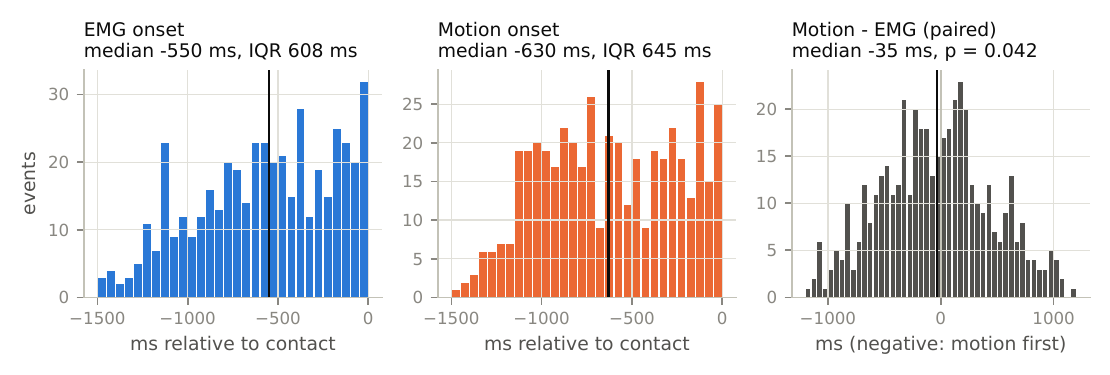}
  \caption{Distribution of per-event onset delays relative to contact ($n = 458$). Left: EMG
    activation onset leads contact by $580 \pm 380$\,ms on average (IQR 608\,ms). Middle: hand motion
    onset leads contact by $627 \pm 377$\,ms on average (IQR 645\,ms). Right: paired
    motion $-$ EMG difference per event (negative: motion first; median $-35$\,ms; the $p$ in the panel is from a Wilcoxon signed-rank test). The
    anticipatory lead is a population-level tendency with large event-to-event variability.}
  \label{fig:grasp-delays}
\end{figure}

\paragraph{Per-event onset delays.}
Figure~\ref{fig:grasp-delays} reports the distributions of per-event onset delays, with
mean EMG onset at $-580$\,ms ($\mathrm{median}=-550$\,ms) and mean motion onset at
$-627$\,ms ($\mathrm{median}=-630$\,ms) relative to contact. These figures are
consistent with established motor-control literature on pre-grasp muscle activation and
reach dynamics.

\paragraph{Implication.}
Both signals lead contact by roughly half a second, while their relative order varies
from event to event. Measuring such inter-modality timing requires the
three signals to be co-registered at a shared high-frequency timebase, as in
PULSE. Beyond reproducing a known motor-control
phenomenon, the same analysis pattern applies to open research questions---how gaze
leads muscle, how anticipatory coordination degrades with task complexity, how
prosthetic-relevant EMG-to-motion mappings vary across subjects---that the dataset is
designed to support.

\subsection{Extended Benchmarks: T3--T7}
\label{sec:app_results_new}

Beyond the headline classification tasks (T1, T2) and the grasp-phase case study,
we provide baseline results for T3--T7, which directly exploit
the dataset's unique signals (synchronised EMG, anticipatory coordination,
finger-level pressure, cross-modal force estimation) and realistic deployment
constraints (missing modalities, unseen scenarios). T3 and T4 use Transformer
backbones with early or late fusion, and T5--T7 the task-specific backbones of their subsections; all experiments use \emph{volunteer-stratified 5-fold
cross-validation} (T4, T7: 33-volunteer pool; T3, T5, T6: 27) with three seeds per fold;
statistics are first averaged within each fold and then
aggregated across folds.

\subsubsection{T3: Grasp Onset Anticipation}

Table~\ref{tab:anticipation} reports binary classification performance of a
Transformer over a 1.0\,s sensor window that ends 0.5\,s before the anchor. Negatives
require long contact-free stretches, which are rarer than onsets, giving a
$0.645$ positive rate at test time on average (0.59--0.69 per fold).

\begin{table}[h]
  \centering
  \caption{Grasp onset anticipation (T3). Binary classification:
    ``will a contact start in the next 500\,ms, given 1\,s of sensor data ending
    0.5\,s before the anchor?''. Pool27, 5 folds $\times$ 3 seeds, mean $\pm$ SD over
    fold means. F1 is the positive-class F1; AUC and AP are the robust metrics under
    the positive fraction shown in the last data column. Test rec.: test recordings
    over the five folds. $\Delta$: AUC difference to EMG on the 192
    test recordings shared by all rows, with the paired $p$ over five folds.}
  \label{tab:anticipation}
  \footnotesize
  \setlength{\tabcolsep}{3pt}
  \resizebox{\textwidth}{!}{%
  \begin{tabular}{@{}l c c c c c r c@{}}
    \toprule
    Modalities              & Acc & F1 & AUC & AP & Pos.\ frac. & Test rec. & $\Delta$ AUC ($p$) \\
    \midrule
    EMG                     & 0.666 $\pm$ 0.030 & 0.753 $\pm$ 0.028 & 0.696 $\pm$ 0.038 & 0.791 $\pm$ 0.045 & 0.645 & 210 & ref. \\
    IMU                     & 0.627 $\pm$ 0.034 & 0.736 $\pm$ 0.019 & 0.607 $\pm$ 0.061 & 0.732 $\pm$ 0.087 & 0.645 & 198 & $-0.085$ (0.021) \\
    MoCap                   & 0.694 $\pm$ 0.047 & 0.784 $\pm$ 0.044 & 0.686 $\pm$ 0.057 & 0.776 $\pm$ 0.035 & 0.645 & 198 & $-0.002$ (0.950) \\
    EyeTrack                & 0.642 $\pm$ 0.038 & 0.767 $\pm$ 0.030 & 0.569 $\pm$ 0.044 & 0.706 $\pm$ 0.035 & 0.645 & 209 & $-0.112$ (0.003) \\
    EMG+IMU                 & 0.626 $\pm$ 0.032 & 0.735 $\pm$ 0.017 & 0.614 $\pm$ 0.067 & 0.734 $\pm$ 0.083 & 0.645 & 198 & $-0.076$ (0.035) \\
    MoCap+EMG+IMU+EyeTrack  & 0.707 $\pm$ 0.043 & 0.787 $\pm$ 0.047 & 0.737 $\pm$ 0.055 & 0.807 $\pm$ 0.068 & 0.643 & 192 & $+0.046$ (0.217) \\
    \bottomrule
  \end{tabular}%
  }
\end{table}

\paragraph{Key finding.}
\textbf{Under cross-subject evaluation, the full 4-modality input has the
highest mean on every metric} (AUC $=0.737$), but its gain over EMG-only
($0.696$; $+4.6$\,pp AUC on shared recordings) is not significant ($p = 0.217$). EMG alone
and MoCap alone are comparably predictive (AUC $0.696$ and $0.686$, $p = 0.950$)---consistent
with the case study, in which EMG and hand-motion onsets both precede contact by about
half a second (Section~\ref{sec:app_grasp_phase})---while IMU alone and EyeTrack alone are
significantly lower. The intermediate EMG+IMU
configuration is significantly worse than EMG alone ($p = 0.035$), so the lift from going to
the full 4-modality set is not monotonic in modality count: \emph{which}
modalities are added matters more than \emph{how many}.

\subsubsection{T4: Missing-Modality Robustness}

Table~\ref{tab:missing-mod} compares a baseline 4-modality late-fusion Transformer
(trained with full modalities) against the same architecture trained with per-sample
random modality dropout ($p = 0.3$). Both models are evaluated on eight test-time
configurations: full, leave-one-out, and motion-modality singletons (MoCap, EMG, IMU),
all on the same 235 test recordings.
Training with modality dropout helps most when MoCap or IMU is missing at test time and
does not hurt clean-test performance.

\begin{table}[h]
  \centering
  \caption{Missing-modality robustness (T4). Test macro F1 on the 8-class
    activity-scenario recognition task under eight evaluation configurations
    (pool33, 235 test recordings for every configuration; removed
    modalities are zeroed; 5 folds $\times$ 3 seeds for both models; mean $\pm$ SD over
    fold means). $\Delta$: dropout minus no dropout, number of folds where dropout is
    better, and paired $p$ over five folds. EyeTrack is reported only in leave-one-out
    and full configurations: it is intended as a complementary stream paired
    with at least one motion modality, never used standalone.}
  \label{tab:missing-mod}
  \footnotesize
  \setlength{\tabcolsep}{4pt}
  \begin{tabular}{@{}l l c c c@{}}
    \toprule
    Eval configuration & Active modalities & No dropout & Dropout $p=0.3$ & $\Delta$ (folds, $p$) \\
    \midrule
    Full                  & MoCap, EMG, Eye, IMU  & 0.460 $\pm$ 0.079 & 0.480 $\pm$ 0.063 & $+0.019$ (4/5, 0.306) \\
    \midrule
    \multicolumn{5}{@{}l@{}}{\emph{Leave-one-out (test only)}} \\
    drop MoCap            & EMG, Eye, IMU         & 0.282 $\pm$ 0.037 & 0.338 $\pm$ 0.045 & $+0.056$ (4/5, 0.040) \\
    drop EMG              & MoCap, Eye, IMU       & 0.469 $\pm$ 0.073 & 0.466 $\pm$ 0.063 & $-0.003$ (3/5, 0.776) \\
    drop EyeTrack         & MoCap, EMG, IMU       & 0.452 $\pm$ 0.089 & 0.453 $\pm$ 0.064 & $+0.001$ (2/5, 0.975) \\
    drop IMU              & MoCap, EMG, Eye       & 0.364 $\pm$ 0.028 & 0.438 $\pm$ 0.076 & $+0.074$ (4/5, 0.093) \\
    \midrule
    \multicolumn{5}{@{}l@{}}{\emph{Singleton (test only; EyeTrack omitted by design)}} \\
    only MoCap            & MoCap                 & 0.347 $\pm$ 0.043 & 0.394 $\pm$ 0.071 & $+0.047$ (4/5, 0.189) \\
    only EMG              & EMG                   & 0.115 $\pm$ 0.026 & 0.153 $\pm$ 0.009 & $+0.038$ (5/5, 0.036) \\
    only IMU              & IMU                   & 0.271 $\pm$ 0.025 & 0.318 $\pm$ 0.074 & $+0.047$ (4/5, 0.153) \\
    \midrule
    Accuracy (full)       & MoCap, EMG, Eye, IMU  & 0.483 $\pm$ 0.069 & 0.499 $\pm$ 0.057 & $+0.016$ (3/5, 0.293) \\
    \bottomrule
  \end{tabular}
\end{table}

\paragraph{Key findings.}
(i) Dropout training helps most when MoCap or IMU is missing (drop MoCap $+0.056$,
$p = 0.040$; drop IMU $+0.074$, $p = 0.093$), raises all three singletons ($+0.038$ to
$+0.047$; significant only for EMG alone), and does not hurt clean input (full $+0.019$,
not significant).
(ii) MoCap and IMU carry most of the scenario-discriminative signal; removing
MoCap drops dropout-trained F1 by $14.2$\,pp ($0.480 \to 0.338$) and removing
IMU drops it by $4.2$\,pp ($9.6$\,pp without dropout).
(iii) EMG and EyeTrack are largely redundant with motion signals for scenario
identification at this data scale: leave-one-out F1 changes by $\leq 2.7$\,pp.
(iv) Volunteer-stratified CV is a strict cross-subject protocol: every
volunteer of the 33-volunteer pool is held out exactly once, so absolute F1 values reflect
worst-case generalization rather than a single favourable split.

\paragraph{Five modalities.}
Table~\ref{tab:missing-mod-5} repeats T4 with pressure as a fifth modality on the 192 test
recordings that have it; pressure does not improve T4 (0.408 vs.\ 0.415 for the 4-modality
model on the same recordings, $p = 0.683$).

\begin{table}[h]
  \centering
  \caption{T4 with five modalities (pool33, 192 test recordings
    with pressure; 5 folds $\times$ 3 seeds; test macro F1, mean $\pm$ SD over fold
    means).}
  \label{tab:missing-mod-5}
  \footnotesize
  \setlength{\tabcolsep}{4pt}
  \begin{tabular}{@{}l c c@{}}
    \toprule
    Eval configuration & No dropout & Dropout $p=0.3$ \\
    \midrule
    Full (5 modalities)  & 0.393 $\pm$ 0.076 & 0.408 $\pm$ 0.062 \\
    drop MoCap           & 0.275 $\pm$ 0.050 & 0.374 $\pm$ 0.052 \\
    drop EMG             & 0.401 $\pm$ 0.075 & 0.405 $\pm$ 0.052 \\
    drop EyeTrack        & 0.384 $\pm$ 0.072 & 0.406 $\pm$ 0.053 \\
    drop IMU             & 0.327 $\pm$ 0.044 & 0.368 $\pm$ 0.068 \\
    drop Pressure        & 0.361 $\pm$ 0.064 & 0.397 $\pm$ 0.095 \\
    only MoCap           & 0.273 $\pm$ 0.027 & 0.328 $\pm$ 0.066 \\
    only EMG             & 0.156 $\pm$ 0.048 & 0.174 $\pm$ 0.042 \\
    only IMU             & 0.246 $\pm$ 0.063 & 0.310 $\pm$ 0.050 \\
    only Pressure        & 0.155 $\pm$ 0.046 & 0.209 $\pm$ 0.029 \\
    \midrule
    4 modalities, same recordings & -- & 0.415 $\pm$ 0.058 \\
    \bottomrule
  \end{tabular}
\end{table}

\paragraph{Split diagnostics.}
Volunteer-stratified CV folds rotate every volunteer through the test set
exactly once, so per-volunteer variance is implicitly reflected in the
cross-fold standard deviations reported here. The released results
files store per-recording statistics, from which per-volunteer metrics can be recomputed for fine-grained diagnosis.

\subsubsection{T5: Tactile-driven Grasp State Recognition (TGSR)}
\label{sec:app_t5_tactile}

\begin{figure}[h]
  \centering
  \includegraphics[width=0.96\linewidth]{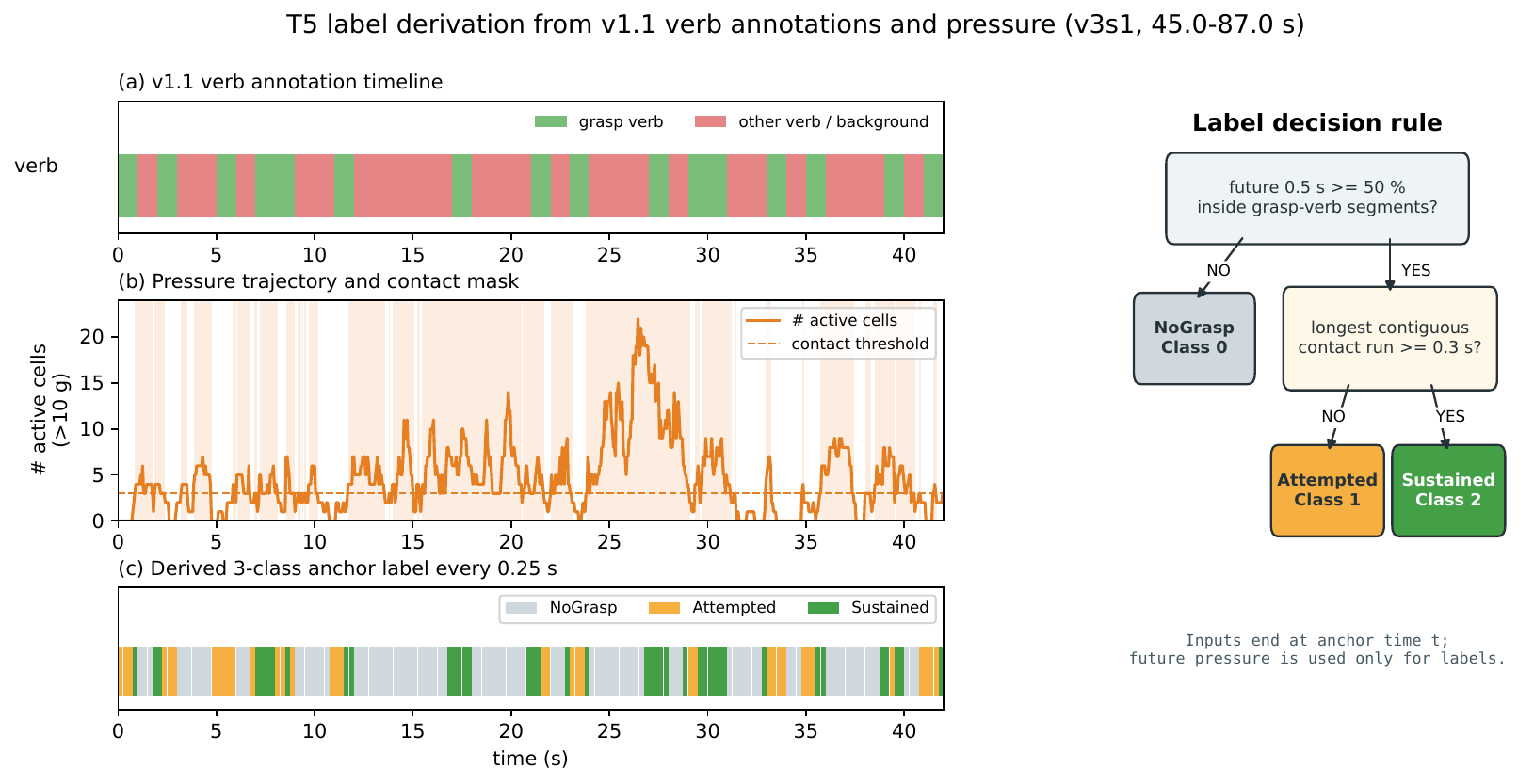}
  \caption{T5 (TGSR) label derivation. \textbf{(a)} Verb annotation timeline
    (green = grasp-verb segment, red = non-grasp verb/background). \textbf{(b)} Pressure
    trajectory ($n_\mathrm{active}$ cells over time, threshold $\geq\!3$
    active cells with per-cell pressure $>\!10$\,g) with contact frames
    shaded orange. \textbf{(c)} Derived 3-class anchor label every 0.25\,s.
    \textbf{Right panel:} Decision tree -- if less than 50\% of the future 0.5\,s window
    lies inside grasp-verb segments, the anchor is NoGrasp; otherwise it is
    SustainedGrasp iff the longest contiguous contact run in that window is
    $\geq\!0.3$\,s, else AttemptedGrasp. Model inputs stop at the anchor
    time $t$; the future pressure window is used only for label derivation.
    The Sustained/Attempted split is
    pressure-only-visible: kinematic motion alone cannot distinguish
    ``contact without object control'' from ``stable object control.''}
  \label{fig:tactile-method}
\end{figure}

Tables~\ref{tab:tactile-online}, \ref{tab:tactile-priv}, and~\ref{tab:tactile-by-event}
report the cross-modal grasp-recognition lift from adding tactile pressure to a
forearm-EMG kinematic baseline. We adopt this leave-target-out setup
(no MoCap, no IMU) to remove the autoregressive shortcut whereby kinematic
position would proxy for ``in contact'' via hand trajectory; the resulting
baseline isolates motor \emph{intent} (EMG) so that pressure's contribution can
only come from the contact signal itself. Table~\ref{tab:tactile-full} repeats the
comparison with the full non-tactile stack. Inputs are causal: for an anchor
at time $t$, both EMG and optional Pressure are drawn only from the past
observation window $[t-T_\mathrm{obs}, t]$, while labels are derived from
the future window $(t,t+T_\mathrm{fut}]$. Thus the pressure side-channel never
contains the same frames used to define the target label. The 3-class label space represents
three semantic object-interaction states and is operationalized by crossing
human verb annotations with the pressure trajectory:
\emph{NoGrasp} denotes no physical contact between the hand and the target object;
\emph{AttemptedGrasp} denotes contact without stable object control, such as
touching or briefly grasping an object without lifting, transporting, or sustained
manipulation; and
\emph{SustainedGrasp} denotes stable object control, such as successfully lifting,
holding, or transporting the object. Operationally, an anchor is labeled NoGrasp unless
at least 50\% of its future window lies in grasp-verb segments (released v1.1 labels); otherwise, contact lasting less
than $0.3$\,s in the future window is labeled AttemptedGrasp, while a contiguous
contact run of at least $0.3$\,s is labeled SustainedGrasp.
Contact at a frame is $\geq\!3$ active cells with per-cell pressure $>\!10$\,g
on the 50-channel pad. All T5 numbers below are mean $\pm$ std across the
5 cross-validation fold means (3 backbones $\times$ 5 folds $\times$ 3 seeds unless stated;
210 test recordings from 27 volunteers); $p$ is a paired $t$-test over the fold means.
Figure~\ref{fig:tactile-method} illustrates the label-derivation pipeline on
a representative session.

\begin{table}[h]
  \centering
  \caption{T5 (TGSR-headline): macro F1 with vs.\ without pressure as a
    side-channel under volunteer-stratified 5-fold CV
    (pool27, 5 folds $\times$ 3 seeds, mean $\pm$ SD over the 5 fold means).
    The kinematic baseline is forearm EMG only (motor intent); pressure
    adds the contact side-channel. All input modalities are restricted to
    the past observation window, with no overlap with the future label
    window.
    Lift $=$ $F1$(EMG\,$+$\,Pressure) $-$ $F1$(EMG only), paired per fold and seed;
    $p$: paired $t$-test over the five fold means.}
  \label{tab:tactile-online}
  \small
  \setlength{\tabcolsep}{6pt}
  \begin{tabular}{@{}lcccc@{}}
    \toprule
    Backbone & $F1$ (EMG only) & $F1$ (EMG $+$ Pressure) & Lift & $p$ \\
    \midrule
    DAF              & 0.383 $\pm$ 0.024 & 0.473 $\pm$ 0.010 & $+0.090 \pm 0.024$ & 0.001 \\
    FUTR             & 0.382 $\pm$ 0.028 & 0.474 $\pm$ 0.011 & $+0.092 \pm 0.026$ & 0.001 \\
    DeepConvLSTM     & 0.374 $\pm$ 0.013 & 0.448 $\pm$ 0.021 & $+0.074 \pm 0.029$ & 0.005 \\
    \midrule
    3-backbone mean  & 0.380 $\pm$ 0.021 & 0.465 $\pm$ 0.012 & {\boldmath$+0.086 \pm 0.024$} & 0.001 \\
    \bottomrule
  \end{tabular}
\end{table}

\begin{table}[h]
  \centering
  \caption{T5 per-class F1, mean over the three backbones (5 folds $\times$ 3 seeds;
    mean $\pm$ SD over the 5 fold means; $p$ over the fold means).
    The Sustained-vs-Attempted distinction is the pressure-only-visible
    axis: kinematic motion alone cannot disambiguate ``contact without
    object control'' from ``stable object control''
    (near-identical EMG signatures),
    and the SustainedGrasp lift is the largest single-class evidence that contact
    pressure encodes grasp-state information unavailable to forearm myoelectric
    activity. NoGrasp F1 decreases on average, but not consistently (lift positive in
    2/5 folds; not significant).}
  \label{tab:tactile-priv}
  \small
  \setlength{\tabcolsep}{6pt}
  \begin{tabular}{@{}lcccc@{}}
    \toprule
    Class & $F1$ (EMG only) & $F1$ (EMG $+$ Pressure) & Lift & $p$ \\
    \midrule
    NoGrasp                 & 0.447 $\pm$ 0.019 & 0.406 $\pm$ 0.056 & $-0.041 \pm 0.072$ & 0.265 \\
    AttemptedGrasp          & 0.371 $\pm$ 0.028 & 0.475 $\pm$ 0.032 & $+0.104 \pm 0.032$ & 0.002 \\
    \textbf{SustainedGrasp} & 0.321 $\pm$ 0.059 & 0.515 $\pm$ 0.047 & {\boldmath$+0.194 \pm 0.018$} & $1.6\times 10^{-5}$ \\
    \bottomrule
  \end{tabular}
\end{table}

\begin{table}[h]
  \centering
  \caption{T5 per-event SustainedGrasp F1, mean over the three backbones
    (5 folds $\times$ 3 seeds; $n_\mathrm{test}$: anchors per fold, averaged over
    folds). Anchors are
    stratified by past/future contact transitions (per-cell threshold
    $>\!10$\,g, $\geq\!3$ active cells). Sustained-class samples exist only in
    the pre-contact and steady-grip events (where the future window contains
    sustained contact); pressure's lift is largest in the steady-grip
    event, where stable object control must be
    distinguished from transient contact without control despite near-identical
    kinematic signatures. Cells with no sustained-class samples (non-contact,
    release) are structurally undefined.}
  \label{tab:tactile-by-event}
  \small
  \setlength{\tabcolsep}{4pt}
  \begin{tabular}{@{}lccccc@{}}
    \toprule
    Event & $n_\mathrm{test}$ & $F1_\mathrm{sust}$ (EMG) & $F1_\mathrm{sust}$ (EMG $+$ Pres) & Lift & $p$ \\
    \midrule
    non-contact          & 8086 & --- & --- & --- & --- \\
    pre-contact          & 1191 & 0.437 $\pm$ 0.042 & 0.500 $\pm$ 0.028 & $+0.062 \pm 0.048$ & 0.043 \\
    \textbf{steady-grip} & 2973 & 0.485 $\pm$ 0.030 & 0.646 $\pm$ 0.011 & {\boldmath$+0.161 \pm 0.039$} & 0.0008 \\
    release              & 1182 & --- & --- & --- & --- \\
    \bottomrule
  \end{tabular}
\end{table}

\begin{figure}[h]
  \centering
  \includegraphics[width=0.92\linewidth]{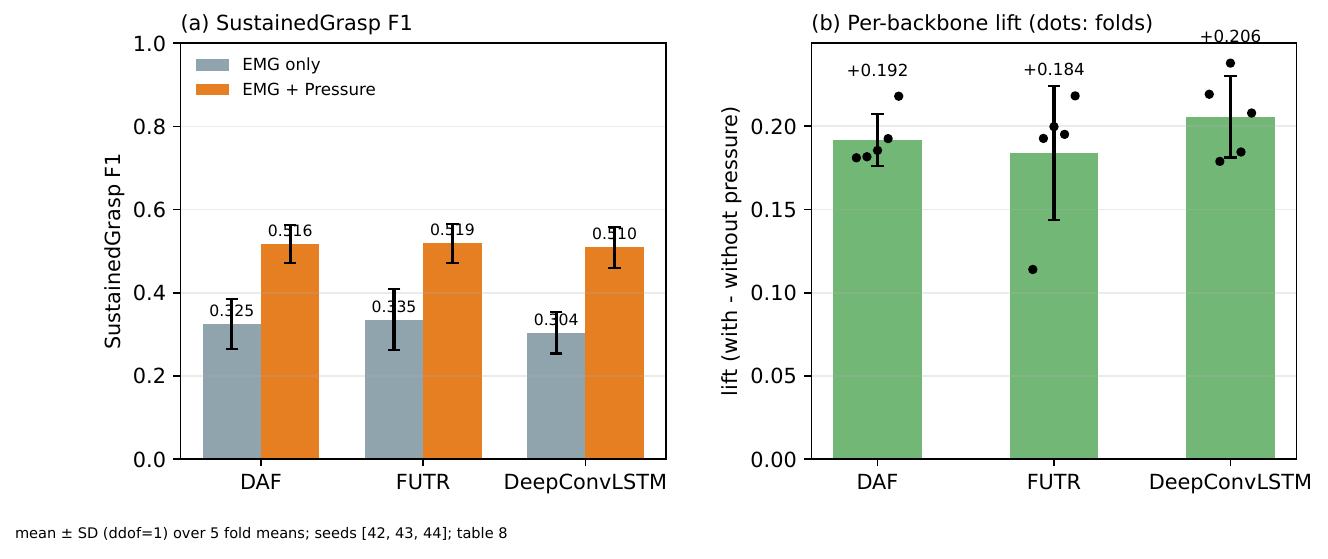}
  \caption{T5 (TGSR) SustainedGrasp F1 lift from adding tactile pressure to a
    forearm-EMG kinematic baseline under 5-fold CV (3 seeds per fold).
    \textbf{(a)} Side-by-side
    comparison of SustainedGrasp F1 with vs.\ without pressure for each
    backbone (mean over the 5 fold means; error bars are 1\,SD). Pressure
    raises SustainedGrasp F1 by about $1.6\times$ for each backbone.
    \textbf{(b)} Per-backbone lift, mean $\pm$ SD over 5 folds (dots: folds).
    The three architecturally diverse backbones agree to within about $0.02$ F1
    (DAF $+0.192$, FUTR $+0.184$,
    DeepConvLSTM $+0.206$), and every fold
    of every backbone gives a positive lift.}
  \label{fig:tactile-lift}
\end{figure}

\paragraph{Key finding.}
Under volunteer-stratified 5-fold cross-validation
(27 volunteers, 3 backbones $\times$ 5 folds $\times$ 3 seeds),
adding pressure to a forearm-EMG kinematic baseline yields an
overall macro F1 lift of $+0.086$ ($p = 0.001$; Table~\ref{tab:tactile-online}) and a
SustainedGrasp F1 lift of $+0.194 \pm 0.018$ ($p = 1.6\times 10^{-5}$; positive in all 45 runs;
Table~\ref{tab:tactile-priv}). The lift is consistent across the three
architecturally diverse backbones (DAF $+0.192$, FUTR $+0.184$,
DeepConvLSTM $+0.206$ SustainedGrasp F1; Appendix~\ref{sec:app_t5_cv})
and event-specific: pressure most amplifies SustainedGrasp recognition
during steady-grip events ($+0.161$ within-event F1, versus $+0.062$ for pre-contact;
Table~\ref{tab:tactile-by-event}), where stable object control must be
distinguished from transient contact without control despite
near-identical kinematic signatures. This empirically confirms the
predictive-tactile component of Johansson 1984~\cite{johansson1984roles} at
the algorithmic level: in a setup where the kinematic baseline encodes
\emph{intent to grasp}, pressure provides the only direct signal of
\emph{sustained contact required for stable object control}---precisely the
information that distinguishes contact without control from stable object
control.

\paragraph{Full non-tactile stack.}
With all four non-tactile modalities as the baseline (Table~\ref{tab:tactile-full}),
pressure still raises macro F1 by $+0.053 \pm 0.009$ ($p = 0.0002$) and SustainedGrasp F1
by $+0.103 \pm 0.023$ ($p = 0.0005$).

\begin{table}[h]
  \centering
  \caption{T5 with the full non-tactile stack (EMG, IMU, eye tracking, MoCap) with
    vs.\ without pressure (pool27, 192 test recordings from 26
    volunteers; 5 folds $\times$ 3 seeds; mean $\pm$ SD over the 5 fold means; lift
    paired per fold and seed).}
  \label{tab:tactile-full}
  \footnotesize
  \setlength{\tabcolsep}{3pt}
  \resizebox{\textwidth}{!}{%
  \begin{tabular}{@{}lcccccc@{}}
    \toprule
    Backbone & macro F1 (full stack) & macro F1 (+ Pressure) & macro lift &
    sust F1 (full stack) & sust F1 (+ Pressure) & sust lift \\
    \midrule
    DAF              & 0.423 $\pm$ 0.023 & 0.474 $\pm$ 0.020 & $+0.051 \pm 0.018$ & 0.344 $\pm$ 0.041 & 0.441 $\pm$ 0.053 & $+0.097 \pm 0.025$ \\
    FUTR             & 0.423 $\pm$ 0.023 & 0.481 $\pm$ 0.022 & $+0.058 \pm 0.007$ & 0.353 $\pm$ 0.035 & 0.460 $\pm$ 0.052 & $+0.107 \pm 0.028$ \\
    DeepConvLSTM     & 0.409 $\pm$ 0.031 & 0.460 $\pm$ 0.027 & $+0.051 \pm 0.010$ & 0.321 $\pm$ 0.067 & 0.428 $\pm$ 0.050 & $+0.107 \pm 0.034$ \\
    \midrule
    3-backbone mean  & 0.419 $\pm$ 0.025 & 0.472 $\pm$ 0.022 & $+0.053 \pm 0.009$ & 0.340 $\pm$ 0.047 & 0.443 $\pm$ 0.051 & $+0.103 \pm 0.023$ \\
    \bottomrule
  \end{tabular}%
  }
\end{table}

\subsubsection{T6: Cross-Modal Pressure Prediction}
\label{sec:app_results_t6}

Table~\ref{tab:t6-summary} reports cross-modal regression performance for the
two T6 sub-tasks: T6a (scalar right-hand grip force) and T6b (50-channel
spatial finger-level pressure). The task definition, input pre-processing, and
backbones are detailed in Appendix~\ref{sec:tasks}.

\paragraph{Honest reporting.}
50-cell distributed regression single-seed Pearson $r$ is highly variable
because $r$ is an extremum across cells. Under volunteer-stratified 5-fold
cross-validation we report two complementary aggregations:
(i) \emph{single-seed CV mean}---per-seed peak $r_{\max}$ averaged across
3 seeds within each fold and then across folds;
(ii) \emph{within-fold prediction-ensemble}---predictions averaged across
seeds within a fold \emph{before} computing $r_{\max}$, then averaged
across folds. The ensemble protocol matches the seed-ensembling used in
the original Ghorbani-style reporting but stays honest under cross-subject
evaluation. Single-seed CV mean is the conservative lower bound; the
within-fold ensemble is the headline because it captures the variance
reduction expected when multiple training runs are available.

\begin{table}[h]
  \centering
  \caption{T6 cross-modal pressure prediction (pool27; 5 folds $\times$ 3 seeds;
    mean $\pm$ SD over fold means). Models predict the right-hand pressure sum
    (T6a, 1-D) or the full 50-channel distribution (T6b). T6b $r_{\max}$ is the
    maximum Pearson correlation across 50 finger-level cells. Test recordings: T6a
    210 (27 volunteers), T6b 198 (26 volunteers).
    The within-fold ensemble averages the predictions of seeds 42, 43, and 44 before
    computing $r$. The highest T6b means are in bold; no T6b difference is significant.}
  \label{tab:t6-summary}
  \footnotesize
  \setlength{\tabcolsep}{5pt}
  \renewcommand{\arraystretch}{1.1}
  \begin{tabular}{@{}l c l c c@{}}
    \toprule
    Method & Year & Modality & Single-seed CV $r$ & Within-fold ensemble $r$ \\
    \midrule
    \multicolumn{5}{@{}l@{}}{\emph{T6a: scalar grip force (right-hand sum, 1-D regression)}} \\
    Pham BiLSTM~\cite{pham2018hand}             & 2018 & EMG only & 0.146 $\pm$ 0.209 & 0.157 $\pm$ 0.224 \\
    Transformer (PULSE)                         & 2026 & EMG only & 0.088 $\pm$ 0.108 & 0.104 $\pm$ 0.128 \\
    Multi-branch CNN~\cite{hajian2021generalized} & 2021 & EMG only & 0.083 $\pm$ 0.140 & 0.093 $\pm$ 0.156 \\
    \midrule
    \multicolumn{5}{@{}l@{}}{\emph{T6b: 50-channel spatial pressure (per-cell $r_{\max}$)}} \\
    Pham BiLSTM~\cite{pham2018hand}             & 2018 & hand MoCap         & 0.235 $\pm$ 0.060 & 0.273 $\pm$ 0.066 \\
    LateFusion (CNN+BiLSTM)                     & 2026 & EMG+MoCap          & 0.248 $\pm$ 0.049 & 0.294 $\pm$ 0.047 \\
    \rowcolor{blue!8}
    Pham BiLSTM~\cite{pham2018hand}             & 2018 & EMG+MoCap (concat) & \textbf{0.272 $\pm$ 0.059} & \textbf{0.305 $\pm$ 0.083} \\
    Ghorbani GRU~\cite{ghorbani2023grip}        & 2023 & hand MoCap         & 0.227 $\pm$ 0.036 & 0.253 $\pm$ 0.051 \\
    \midrule
    Ghorbani GRU$^{\dagger}$~\cite{ghorbani2023grip} & 2023 & hand MoCap+EMG+Eye+IMU & 0.274 $\pm$ 0.053 & -- \\
    \bottomrule
  \end{tabular}

  \vspace{0.2em}
  {\scriptsize $^{\dagger}$192 test recordings with all four inputs; no control row on the same recordings, so this row is not compared with the others.}
\end{table}

\paragraph{Key findings.}
\textbf{(i)} Under the single-seed CV protocol, the four T6b methods lie within a
0.045 band (0.227--0.272);
even the best and the lowest row do not differ significantly
($\Delta = +0.045$,
$p =$~0.113).
\textbf{(ii)} The within-fold prediction ensemble raises every row but does not
separate them. Pham BiLSTM with concatenated EMG and hand MoCap has the highest mean
(0.305 $\pm$ 0.083); none of the three other rows differs
significantly from the lowest row, Ghorbani GRU with hand MoCap
($p \geq$~0.083). The four T6b models
therefore do not differ significantly at this sample size.
\textbf{(iii)} All models stay far below fixed-test-set literature numbers reported on
related benchmarks ($r \approx 0.91$--$0.99$ for isometric single-grasp force
regression). T6 thus remains an open challenge for cross-subject force estimation
during everyday manipulation on PULSE.

\paragraph{Modality analysis.}
For scalar grip force (T6a), all three EMG models reach low correlations
($r \leq 0.16$). A budget-matched comparison trains each backbone on identical
recordings with EMG only or with MoCap, EMG, eye tracking, and IMU
(Table~\ref{tab:t6a-matched}). Whether EMG alone or the 4-modality input is better
depends on the model: EMG has the higher mean for the Pham BiLSTM, the two are equal for
the Transformer, and the 4-modality input has the higher mean for the multi-branch
CNN; none of these differences in $r$ is significant, and all correlations stay at or
below 0.16. For the 50-channel target (T6b), concatenating EMG with hand MoCap raises the
Pham BiLSTM's mean $r_{\max}$ (0.235 $\to$
0.272 single-seed,
0.273 $\to$
0.305 ensemble), but the four T6b models
are not significantly different, so the data do not single out a dominant modality or
architecture for spatial pressure.

\begin{table}[h]
  \centering
  \caption{T6a budget-matched comparison: each backbone is trained and tested on the
    same 192 recordings (26 volunteers) with EMG only or with
    MoCap+EMG+Eye+IMU (single-seed CV mean over 3 seeds and 5 folds). $\Delta r$:
    4-modality minus EMG, folds where the 4-modality input is better, and paired $p$
    over five folds.}
  \label{tab:t6a-matched}
  \footnotesize
  \setlength{\tabcolsep}{4pt}
  \begin{tabular}{@{}l l c c c@{}}
    \toprule
    Backbone & Input & Pearson $r$ & MAE (g) & $\Delta r$ (folds, $p$) \\
    \midrule
    Pham BiLSTM~\cite{pham2018hand}               & EMG     & 0.140 $\pm$ 0.164 & 181.5 $\pm$ 49.9 & ref. \\
                                                  & 4-mod   & 0.062 $\pm$ 0.095 & 204.4 $\pm$ 41.3 & $-0.077$ (0/5, 0.072) \\
    Transformer (PULSE)                           & EMG     & 0.088 $\pm$ 0.093 & 192.2 $\pm$ 43.3 & ref. \\
                                                  & 4-mod   & 0.089 $\pm$ 0.145 & 198.6 $\pm$ 45.1 & $+0.001$ (2/5, 0.971) \\
    Multi-branch CNN~\cite{hajian2021generalized} & EMG     & 0.041 $\pm$ 0.051 & 257.1 $\pm$ 33.7 & ref. \\
                                                  & 4-mod   & 0.069 $\pm$ 0.104 & 453.9 $\pm$ 50.4 & $+0.028$ (3/5, 0.316) \\
    \bottomrule
  \end{tabular}
\end{table}

\subsubsection{T7: EMG-to-Hand-Pose Reconstruction}
\label{sec:app_results_t7}

Table~\ref{tab:t7-emg-pose} reports T7 results for reconstructing
21-joint right-hand pose from forearm EMG. Given a synchronized EMG sequence,
models predict a 63-dimensional pose vector at each frame, corresponding
to the 3D coordinates of 21 hand joints in the wrist frame. T7 is intended as
a motor-regression benchmark rather than a new model contribution: we
evaluate NeuroPose, the TDS featurizer of emg2pose, a multi-scale
CNN+Transformer, and a generic Transformer encoder under the same
volunteer-stratified cross-subject protocol, and compare every model with the
training-set mean pose, which uses no EMG.

\begin{table}[h]
  \centering
  \caption{T7 EMG-to-hand-pose reconstruction (33-volunteer pool,
    volunteer-stratified 5-fold CV, 3 seeds per fold; mean $\pm$ std over
    folds). Models predict 21-joint right-hand pose from EMG only.
    ``Mean pose'' predicts the training-set mean pose at every frame and
    ignores the EMG. $\Delta$: full-hand MPJPE minus that of the mean pose
    (negative = better), with the number of folds in which the model is
    better. MPJPE and fingertip MPJPE are reported in millimeters; bone angle
    error is reported in degrees. Lower is better for all metrics.}
  \label{tab:t7-emg-pose}
  \footnotesize
  \setlength{\tabcolsep}{5pt}
  \renewcommand{\arraystretch}{1.1}
  \begin{tabular}{@{}l c c c c@{}}
    \toprule
    Method & MPJPE $\downarrow$ & Fingertip MPJPE $\downarrow$ & Angle Error $\downarrow$ & $\Delta$ MPJPE (folds) \\
    \midrule
    TDS (emg2pose)~\cite{salter2024emg2pose}
      & \textbf{19.71 $\pm$ 0.99}
      & \textbf{31.32 $\pm$ 1.96}
      & \textbf{21.56 $\pm$ 0.96}
      & $-0.86$ (4/5) \\
    NeuroPose~\cite{10.1145/3442381.3449890}
      & 20.09 $\pm$ 1.33
      & 31.92 $\pm$ 2.42
      & 21.83 $\pm$ 2.01
      & $-0.48$ (4/5) \\
    MS-CNN+Transformer
      & 20.79 $\pm$ 1.33
      & 33.18 $\pm$ 2.39
      & 22.43 $\pm$ 1.23
      & $+0.22$ (3/5) \\
    Transformer
      & 20.80 $\pm$ 1.51
      & 33.23 $\pm$ 2.50
      & 23.01 $\pm$ 1.44
      & $+0.23$ (3/5) \\
    \midrule
    Mean pose (training set, no EMG)
      & 20.57 $\pm$ 1.14
      & 33.21 $\pm$ 2.02
      & 21.92 $\pm$ 1.33
      & -- \\
    \bottomrule
  \end{tabular}
\end{table}

\paragraph{Key findings.}
\textbf{(i)} Under cross-subject evaluation, current EMG-to-pose models
recover little hand-pose information beyond the population mean pose. The
best model, the TDS featurizer of emg2pose, reaches a full-hand MPJPE of
19.71 $\pm$ 0.99\,mm against
20.57 $\pm$ 1.14\,mm for the mean pose
($\Delta = -0.86 \pm 0.77$\,mm,
better in 4/5 folds); the difference is not
significant over the five folds (paired $t$-test, $p =$~0.066).
\textbf{(ii)} The clearest gain is at the fingertips, where the error is
largest: fingertip MPJPE changes by
$\Delta = -1.89$\,mm for TDS
(4/5 folds,
$p =$~0.036) and
$\Delta = -1.28$\,mm for NeuroPose
($p =$~0.069).
\textbf{(iii)} The two attention-based models, MS-CNN+Transformer and the
plain Transformer, are slightly worse than the mean pose in full-hand MPJPE
($\Delta = +0.22$ and
$+0.23$\,mm) and in angle error.

\paragraph{Discussion.}
T7 is therefore an open problem on PULSE rather than a solved regression
task. Forearm EMG differs considerably between people (electrode placement,
anatomy, skin impedance); with volunteer-disjoint folds and no per-user
calibration, the models stay close to a typical hand configuration.
We report the mean-pose reference so that future methods, for example
user-adaptive or calibration-based EMG decoders, can be judged by how far
they move away from it. We treat T7 as a complementary motor-regression
probe to T6: T6 asks whether motion and EMG can reconstruct physical
contact pressure, whereas T7 asks whether EMG alone can recover the
kinematic hand configuration that precedes and accompanies contact.

\section{Dataset Card}
\label{sec:app_datasheet}

Following the \emph{Datasheets for Datasets} framework~\cite{gebru2021datasheets},
Table~\ref{tab:datasheet} summarises the key metadata of PULSE.

\begin{table}[h]
  \centering
  \caption{PULSE dataset card.}
  \label{tab:datasheet}
  \footnotesize
  \setlength{\tabcolsep}{4pt}
  \renewcommand{\arraystretch}{1.15}
  \begin{tabular}{@{}p{0.24\linewidth}p{0.7\linewidth}@{}}
    \toprule
    \textbf{Field} & \textbf{Value} \\
    \midrule
    Name            & PULSE \\
    Motivation      & Study sensorimotor coordination (gaze$\to$muscle$\to$motion$\to$contact) during long-horizon daily activities \\
    Modalities      & Motion capture (MoCap, full-body + fingers), surface EMG, binocular eye tracking, inertial sensors (IMU), finger-level pressure \\
    Sampling rate   & 100\,Hz; all modalities time-aligned on a common timebase from each device's recorded start timestamp \\
    Subjects        & 40 healthy adult volunteers; age 21.2$\pm$2.1 years (18--26), 31 male / 9 female, height 174.6$\pm$7.2\,cm, weight 65.1$\pm$10.1\,kg, handedness 26 right and 14 left \\
    Scenarios       & 8 daily activity scenarios (S1--S8) + 1 isolated motion-primitive protocol (S9, from 33 of the 40 subjects; 32 usable recordings) \\
    Recordings      & 304 task recordings (S1--S8; v4/s1 and v4/s4 contain no data and are listed only in \texttt{recordings.csv}) + 33 motion-primitive recordings (S9); 198 with all five modalities; 282 annotated at segment level (from 36 of the 40 volunteers) \\
    Total duration  & $\sim$6.8\,hours task scenarios + $\sim$2.8\,hours motion-primitive protocol (per-recording: 0.3--4.0\,min, mean 1.3\,min for S1--S8) \\
Annotations     & 7{,}356 action segments across 282 annotated recordings (VLM annotation followed by human review; every segment confirmed or corrected by a human); 18 motor-primitive classes, left/right/bimanual hand labels, object names, natural-language captions with 4 paraphrases \\
    Train/test split & Subject-independent volunteer-stratified 5-fold CV over two pools: 33 volunteers with EMG and skeleton MoCap (T1, T4, T7) and 27 annotated volunteers with pressure (T2, T3, T5, T6); inner-validation volunteers for model selection (Section~\ref{sec:dataset}) \\
    Format          & CSV (per-modality aligned at 100\,Hz) + TSV (solved skeleton) + MP4 (first-person video) + JSON/CSV/XLSX (metadata and annotations) \\
    Distribution    & Public Hugging Face dataset repository + companion code and representative sample; archived under the permanent DOI \href{https://doi.org/10.57967/hf/11314}{10.57967/hf/11314} \\
    License (data)  & CC BY-NC 4.0 \\
    License (code)  & MIT \\
    Ethics review   & Approved by the Ethics Committee of South China University of Technology (\mbox{S-2025-026-01}, 4~March 2025); all subjects signed informed-consent and media-release forms \\
    Intended use    & Academic research on multi-modal sensorimotor prediction, activity understanding, motor control, rehabilitation, imitation learning, and wearable sensing \\
    Maintenance     & Maintained by the authors for at least five years after publication; corrections are released as versioned updates \\
    Contact         & Wolin Liang, \texttt{202610196903@mail.scut.edu.cn} \\
    \bottomrule
  \end{tabular}
\end{table}

\section{Dataset License, Availability, and Ethics}
\label{sec:app_license}

\paragraph{License.}
The PULSE dataset is available under the \textbf{Creative Commons
Attribution--NonCommercial 4.0 International (CC BY-NC 4.0)} license, which
permits sharing and adaptation for non-commercial research purposes with
attribution. The accompanying benchmark code (data loaders, training pipelines,
baseline implementations, and evaluation scripts) is available
under the \textbf{MIT License}.

\paragraph{Hosting and long-term availability.}
The full dataset artifact (3,653 files,
85.1\,GB, of which 63.1\,GB are first-person videos) is publicly available at
\url{https://huggingface.co/datasets/velvet-pine-22/PULSE}, including the
\texttt{croissant.json} (Croissant 1.0 with RAI fields), \texttt{DATASHEET.md}
(Gebru-style datasheet), \texttt{LICENSE} (CC BY-NC 4.0), and
\texttt{CODE\_LICENSE} (MIT). The companion code is available at
\url{https://huggingface.co/velvet-pine-22/PULSE-code}, and a small representative
sample for quick schema inspection is available at
\url{https://huggingface.co/datasets/velvet-pine-22/PULSE-sample}.
The dataset is archived under the permanent DOI
\href{https://doi.org/10.57967/hf/11314}{10.57967/hf/11314} and is committed to
remain available for at least five years after publication.

\paragraph{Ethics and privacy.}
All participants signed an informed-consent form and a media-release agreement
explicitly authorising the public release of synchronised scene-camera,
eye-tracking video, and biomechanical data (Section~\ref{sec:protocol}). The
released first-person videos contain no audio track and do not show the participant's
own face; laboratory members who appear incidentally consented to the release, and their faces are pixelated. Participant names,
contact details, and
any other directly identifying metadata are kept confidential and are not
distributed. The recording protocol was reviewed and approved by the
Ethics Committee of South China University of Technology (approval no.\ \mbox{S-2025-026-01}, 4~March 2025). Users of the dataset must respect these
conditions: in particular, they must not attempt to re-identify participants
beyond what is already present in the released data, and any downstream use
involving additional human subjects must undergo its own ethics review.

\paragraph{Author responsibility.}
The authors bear responsibility for any rights violations arising from data
collection and will take corrective action---including removal of affected
records from public releases---should any issue be identified.

\section{Dataset Details}
\label{sec:app_dataset}

\begin{table}[h]
  \centering
  \caption{Sensor modalities in PULSE. ``Coverage'' denotes the fraction of
    the 304 task recordings (S1--S8) for which the modality is available. IMU: 16
    signal channels per unit plus temperature and battery level, i.e., 160 signal
    channels and 180 released columns; the benchmark models use all 180 columns.}
  \label{tab:sensors}
  \setlength{\tabcolsep}{5pt}
  \begin{tabular}{@{}llccc@{}}
    \toprule
    \textbf{Modality} & \textbf{Sensor} & \textbf{Raw Dim.} & \textbf{Processed Dim.} & \textbf{Coverage} \\
    \midrule
    Motion capture (MoCap) & Qualisys system      & 422 & 620 & $\approx$81\% \\
    EMG                    & Myo Armband          & 8   & 8   & $\approx$99\% \\
    Eye tracking           & Dikablis Glasses~3   & 24  & 24  & $\approx$93\% \\
    IMU                    & 10 wearable sensors  & 180 & 180 & $\approx$98\% \\
    finger-level pressure     & Pressure sensor array & 50 & 50  & $\approx$70\% \\
    \bottomrule
  \end{tabular}
\end{table}

\begin{table}[h]
  \centering
  \caption{Hardware specifications of the sensors used to record PULSE.
    Body and hand motion capture share the same Qualisys optical camera system
    under two different marker protocols (Animation Marker Set, 42 body markers;
    Full Fingers Marker Set, 2$\times$15 finger markers).}
  \label{tab:device_specs}
  \scriptsize
  \setlength{\tabcolsep}{3pt}
  \resizebox{\textwidth}{!}{%
  \begin{tabular}{@{}lllll@{}}
    \toprule
    \textbf{Modality} & \textbf{Model} & \textbf{Manufacturer} & \textbf{Channels / Protocol} & \textbf{Interface} \\
    \midrule
    Body MoCap        & Animation Marker Set     & Qualisys AB (Sweden)         & 42 passive reflective markers       & Optical (QTM) \\
    Hand MoCap        & Full Fingers Marker Set  & Qualisys AB (Sweden)         & 2$\times$15 finger markers          & Optical (QTM) \\
    Eye tracking      & Dikablis Glasses~3       & Ergoneers GmbH (Germany)     & Binocular, $0.3^{\circ}$ accuracy   & USB (D-LAB) \\
    sEMG              & Myo Armband              & Thalmic Labs (Canada)        & 8-ch dry sEMG + 9-axis IMU          & BLE \\
    IMU               & WT9011DCL                & WitMotion (China)            & 9-axis (acc / gyro / mag)           & BLE 5.0 \\
    Finger-level pressure & GT-G0505M                & Guantuo Electronics (China)  & 25 points per hand (50 total)       & USB-serial \\
    \bottomrule
  \end{tabular}%
  }
\end{table}

\subsection{Data Format and File Organization}
\label{sec:app_format}

The released dataset is organized hierarchically by \emph{subject}~$\to$~\emph{scenario},
with one directory per recording. Each recording directory contains the five modality
streams as CSV files aligned to a shared 100\,Hz timebase, the Myo orientation streams,
the aligned skeleton export of the motion-capture system, and a JSON metadata file
documenting the alignment. Action-segment annotations are provided as one JSON file per
annotated recording and as a flat table. The layout of the public release is:

\begingroup\footnotesize
\begin{verbatim}
PULSE/
  README.md  DATASHEET.md  LICENSE  CODE_LICENSE  croissant.json
  data/v{id}/s{id}/
    aligned_mocap_100hz.csv      # marker CSV: time, Frame, Time + 52 markers x (X,Y,Z,Type)
    aligned_emg_100hz.csv        # 8-ch raw signed Myo EMG
    aligned_eyetrack_100hz.csv   # Dikablis export; 24 core gaze features
    aligned_imu_100hz.csv        # 180 cols: 10 units x (16 signals + temperature, battery)
    aligned_pressure_100hz.csv   # R1..R25, L1..L25 in grams
    aligned_myo_quat_100hz.csv   # Myo orientation quaternion
    aligned_myo_pose_100hz.csv
    alignment_metadata.json      # per-stream start/end times, aligned window
    raw/aligned_v{id}s{id}_s_Q.tsv  # 56-joint skeleton + 10 fingertips (422 dims)
    videos/scene_cam.mp4             # first-person scene camera, 1920x1080, 30 fps
    videos/eye_tracking_overlay.mp4  # same view with gaze overlay, recorded frame rate
  annotations/v{id}/s{id}.json   # 282 annotated recordings
  annotations_flat/segments.csv  # 7,356 segments, one row each
  metadata/recordings.csv        # per-recording modality availability
  metadata/modality_coverage.xlsx  metadata/batch_alignment_summary.json
\end{verbatim}
\endgroup

\noindent For 10 recordings the skeleton TSV lies at the recording root instead of in
\texttt{raw/}. The \texttt{split} column of \texttt{recordings.csv} is a legacy fixed
hold-out and is not used by any result in this paper. Of the 313 eye-tracking files, 291 hold
exactly the 24 core features, 20 add 4 or 10 further Dikablis export columns, and v1/s1
and v14/s8 contain one eye only (14 features).

\paragraph{Annotation file schema.}
Each per-recording file holds \texttt{\{"segments": [...]\}}; volunteer and scenario
are given by the path. Each segment contains \texttt{timestamp}
(\texttt{HH:MM:SS-HH:MM:SS}, 1\,s resolution) and \texttt{action\_annotation} with
\texttt{action\_name} (one of the 18 motor primitives),
\texttt{hand\_type} (\{\textit{left}, \textit{right}, \textit{both}\}),
\texttt{is\_lhand} and \texttt{is\_rhand} (binary hand-participation flags),
\texttt{object\_name} (English object name from the scenario whitelist),
\texttt{text\_key} (natural-language description), and
\texttt{text\_variants} (four paraphrased English variants preserving action, object, and hand).
The flat table \texttt{segments.csv} has one row per segment with the columns
\texttt{volunteer}, \texttt{scenario}, \texttt{recording\_id}, \texttt{segment\_idx},
\texttt{timestamp}, \texttt{start\_sec}, \texttt{end\_sec}, \texttt{duration\_sec},
\texttt{primitive}, \texttt{hand}, \texttt{is\_left\_hand}, \texttt{is\_right\_hand},
\texttt{object}, \texttt{description}, and \texttt{description\_variants}.

\paragraph{Typical per-recording footprint.}
The sensor and metadata files of one recording occupy about 59\,MB (median; 14--142\,MB
for 90\% of recordings), 22.0\,GB in total. The two first-person videos are the dominant
contributors (median $\sim$73\,MB per file, at most $\sim$0.5\,GB; 63.1\,GB in total);
users who need only the sensor streams can skip the \texttt{videos/} folders. Loaders that parse this layout (the \texttt{pulse.data}
package) are provided in the accompanying code repository.

\subsection{Sensor Donning and Placement Protocol}
\label{sec:app_donning}

Before calibration, each participant follows a fixed sensor-donning order to
minimize interference between devices and to keep body-worn electronics stable
throughout the session. The order is:
\begin{enumerate}[leftmargin=1.5em,itemsep=1pt,topsep=2pt]
  \item \textbf{Myo Armband first.} The Myo is placed on the right forearm
        before any optical markers or other wearable modules so that the armband
        sits directly against the skin and its channel orientation can be fixed.
  \item \textbf{MoCap markers second.} Body and full-finger reflective markers
        are then attached while the Myo position is already fixed, reducing the
        chance that later armband adjustment displaces nearby markers.
  \item \textbf{IMUs and finger-level pressure arrays third, in parallel.} The 10
        IMUs and the bilateral finger-level pressure arrays are mounted after the
        marker set. These two devices are installed at the same stage because
        they occupy different body regions and neither requires later access
        beneath the optical markers.
  \item \textbf{Dikablis Glasses~3 last.} The eye-tracking glasses are fitted
        after all body, hand, and cable routing steps are complete, preventing
        the headset from being disturbed while other sensors are attached.
\end{enumerate}
After all sensors are mounted, the experimenter checks cable slack, marker
visibility, pressure-array contact, wireless connectivity, and participant
comfort before proceeding to the calibration sequence in
Section~\ref{sec:app_calibration}.

\subsection{Calibration Procedures}
\label{sec:app_calibration}

Each session begins with a fixed calibration sequence that establishes reference
frames, zero points, and per-subject parameters for every sensor before activity
recording starts.

\paragraph{Motion capture (Qualisys).}
The optical camera system is calibrated at the start of each recording day using the
standard Qualisys wand (L-frame) procedure; residuals are checked in QTM to confirm
sub-millimetre reconstruction accuracy. Each subject additionally performs a brief
\emph{T-pose} and \emph{A-pose} at the beginning of the session for marker
identification and for hand / body segment labelling.

\paragraph{Eye tracking (Dikablis Glasses~3).}
Binocular gaze is calibrated using the D-LAB EYETRACKING software via a 5-point
fixation routine at known depths. Calibration is repeated at the end of the session
as a drift check.

\paragraph{Surface EMG (Myo Armband).}
The Myo is placed on the right forearm with a fixed channel-to-muscle-group
orientation (EMG channel~1 aligned with the extensor digitorum). A short
maximum-voluntary-contraction (MVC) reference trial (finger extension + forearm
supination, 3\,s) is recorded for optional per-subject amplitude normalisation.

\paragraph{IMU (WitMotion WT9011DCL).}
All ten IMUs are placed on the hands in a standardised bilateral pattern: five
per hand, with one sensor on the wrist and one on each of the thumb, index,
middle, and ring fingers. They are zeroed while the subject stands still in an
\emph{N-pose} for five seconds. A subsequent slow hand-and-wrist range-of-motion sweep records
bias-drift characteristics that can be used for sensor fusion in downstream analysis.

\paragraph{finger-level pressure (GT-G0505M).}
Zero-baseline values are recorded with both hands held relaxed in the air.
Participants are not asked to perform a quantitative force calibration;
reported pressure values therefore correspond to manufacturer-factory grams-force
units.

\paragraph{Cross-modal synchronisation.}
Beyond the per-sensor calibrations above, each acquisition system records its streams
with its own start timestamp. All streams are resampled onto a common 100\,Hz timebase
and placed on it from these recorded start timestamps; the aligned window of each
recording is the overlap of all available streams (Section~\ref{sec:protocol}).
Per-stream start and end times are documented in each recording's
\texttt{alignment\_metadata.json}.

\subsection{Annotations and Preprocessing}
\label{sec:app_annot_preproc}

\paragraph{Scenario labels.}
Each recording is assigned one of the eight scenario labels (S1--S8) defined in
Appendix~\ref{sec:app_scenarios}.

\paragraph{VLM-assisted annotation pipeline.}
Action-level annotations are produced by a \textbf{vision-language-model (VLM) assisted
pipeline}. For each recording, the compressed first-person eye-tracker video
($640\!\times\!H$ at 15\,fps, CRF~28) is passed to a VLM under a
scene-specific structured prompt. The prompt injects a fixed 18-primitive action
vocabulary (see below) and a per-scene object whitelist,
which together constrain the VLM output to a closed vocabulary and prevent object
hallucination. The model returns a JSON array of segments, each containing:

\begin{itemize}[leftmargin=1.5em,itemsep=1pt,topsep=2pt]
  \item \texttt{timestamp}: start/end time in \texttt{HH:MM:SS-HH:MM:SS} format (1\,s resolution, segments typically 1--3\,s);
  \item \texttt{action\_name}: the dominant motor primitive for the segment (one of 18);
  \item \texttt{hand\_type}: \{\textit{left}, \textit{right}, \textit{both}\};
  \item \texttt{is\_lhand}, \texttt{is\_rhand}: binary indicators of which hand participates;
  \item \texttt{object\_name}: the primary object affected (from the scene whitelist);
  \item \texttt{text\_key}: a natural-language description of the action (e.g., \texttt{``Grasp box with both hands.''});
  \item \texttt{text\_variants}: an array of four paraphrased variants of \texttt{text\_key}, preserving the hand/object/action.
\end{itemize}

The 18-primitive vocabulary is:
\textit{grasp, hold, move, push, pull, rotate, open, close, insert, remove, place,
pick\_up, put\_down, adjust, align, release, support, stabilize.} The prompt
template is reproduced in Appendix~\ref{sec:app_prompt}; the scenario-specific object
vocabularies are released with it in the code repository.
The vocabulary was fixed before annotation and organized by contact state:
establishing contact (grasp, pick\_up, hold); transport (move, push, pull);
release (place, put\_down, release); object-state change (open, close, rotate,
insert, remove); and stabilization (adjust, align, support, stabilize). We did not
adopt a published taxonomy wholesale, but this touch/untouch axis is the one
W\"org\"otter et al.~\cite{worgotter2013ontology} derive their manipulation
ontology from, at a granularity comparable to the 15 classes of Bullock
et al.~\cite{bullock2013handcentric}.

\paragraph{Human review of VLM annotations.}
Every annotated recording is then checked by a human segment by segment, and each
segment is confirmed or corrected, so all released segments have been confirmed or
corrected by a human. The review procedure and its outcome are described in
Appendix~\ref{sec:app_human_validation}.

\paragraph{MoCap feature engineering.}
Raw marker positions and quaternions are transformed into a \emph{hip-relative}
coordinate frame so that global translation is removed. First-order finite differences
are appended as velocity features, expanding the representation from 422 to \textbf{620
dimensions}.

\paragraph{Normalization.}
Each feature channel is independently z-score normalized using mean and standard
deviation computed on the training split only. Test and validation sets are normalized
using the training statistics to avoid information leakage.

\paragraph{Temporal resampling.}
All modalities are subsampled from 100\,Hz to \textbf{20\,Hz} by keeping every fifth
frame (no low-pass filter; T7 instead pools EMG by block RMS). This reduces sequence length by a factor of five while retaining features
relevant to daily activity recognition (which rarely exceeds 5\,Hz in effective bandwidth).

\paragraph{Sequence padding and masking.}
Recording lengths vary between 0.3 and 4.0 minutes (360--4{,}800 frames at 20\,Hz after
downsampling). Within each mini-batch, sequences are zero-padded to the length of the
longest element, and a binary attention mask is provided to all models so that padded
positions are ignored during loss computation and metric evaluation.

\subsection{Activity Scenario Descriptions}
\label{sec:app_scenarios}

This appendix details the eight activity scenarios in PULSE. For each scenario we
provide: (i) the \emph{instruction} given to volunteers verbatim, (ii) the initial \emph{scene setup} (object placement at session
start), and (iii) the available \emph{props}. Volunteers were free to use any subset of
the props in their preferred order, yielding natural inter-subject variation.

\paragraph{S1: Office desk organization (target duration $\sim$2\,min).}
\textit{Instruction:}
``You just arrived at the office and found your desk in disarray from yesterday. Tidy
up the workspace and prepare to start the day's work. The desk has assorted office
supplies, documents, and electronics. Please organize the desk according to your usual
habits and prepare to start working.''
\textit{Initial state (intentionally cluttered):}
5--6 documents scattered on the desk; an overturned pen holder with markers spread out;
a stapler, hole punch, and scissors placed haphazardly; a misaligned keyboard and mouse;
a water cup resting on a document; tangled charging cables; a wallet; loose tape and
small items.

\paragraph{S2: Package shipping preparation (target $\sim$2\,min).}
\textit{Instruction:}
``You need to ship important documents and small items to a client. Choose appropriate
packaging, pack the items safely, seal the package, and move it to a designated location
(the top shelf of a rack; note the rack is positioned to challenge occlusion-aware
recording).''
\textit{Available props:} 3--4 boxes/envelopes of varying sizes, foam wrap, bubble wrap,
documents, small items (USB drives, business-card holders, small gifts), tape, scissors,
address labels, shipping forms, markers, an electronic scale, and seal stickers.

\paragraph{S3: Kitchen spice organization (target $\sim$4\,min).}
\textit{Instruction:}
``Reorganize the kitchen spice rack. Reposition the bottles by frequency of use, wipe
down the bottle exteriors, label them with their contents, transfer loose spice into
sealed jars, organize the disposable utensils, and clean the countertop.''
\textit{Setup:} Five spice bottles, then re-arranged.
\textit{Available props:} 5 spice bottles, 1 sealed jar, kitchen paper, and a
cleaning cloth (occlusion-prone---volunteers were instructed on how to minimise marker
occlusion).

\paragraph{S4: Post-meal table cleaning (target $\sim$3\,min).}
\textit{Instruction:}
``You just finished eating and need to clean the dining table. Sort and collect the
dishes (bowls and plates separately), discard the residue from the plates into the trash,
wipe the table with a cloth to clean spilled liquids, organise the used cloths, gather
the placemats, and push the chairs back to their original positions.''
\textit{Available props:} Used dishes (3--4 rice bowls, 4--5 plates, 2 soup bowls, 3--4
water cups), simulated food residue (plastic fragments), a trash bin, paper towels and
napkins, wet wipes, two cloths (dry/wet), 3--4 condiment bottles (soy sauce, vinegar,
chilli oil), 4 placemats, a tray for collecting dishes, a tablecloth, cleaning spray
(water), and disposable gloves.

\paragraph{S5: Pre-meal table setting (target $\sim$1\,min~20\,s).}
\textit{Instruction:}
``It's almost dinner time, please set the dining table. Lay and smooth the tablecloth
(occlusion-prone), set 4 place settings (each with a bowl, plate, chopsticks, and spoon),
arrange the napkins and tissue box, place the condiment set in the centre of the table,
prepare communal chopsticks and a serving spoon, pour tea into the cups, place a
decorative vase, and adjust the chairs. Then bring out the food (a pot plus serving
dishes).''
\textit{Available props:} Tablecloth, 4 rice bowls, 8 plates (4 large, 4 small), 4 pairs
of chopsticks plus 2 communal pairs, 4 spoons plus 2 communal spoons, a large soup
ladle, 4 water/tea cups, 4 placemats, folded napkins, a tissue box, a condiment set on
a small tray, a teapot (filled with a small amount of water), a small flower vase, 4
chopstick rests, a Lazy Susan, and wet wipes.

\paragraph{S6: Business travel luggage packing (target $\sim$2\,min).}
\textit{Instruction:}
``Pack a 3-day business-trip suitcase. Pack the provided items into the suitcase
(opened flat to reduce side-wall occlusion). Protect electronics, fold garments
appropriately, place toiletries in a toiletry bag, keep documents accessible, and tidy
the charging cables. Heavy items go on the bottom, valuables in safe positions, and the
suitcase must close cleanly.''
\textit{Available props:}
\textit{Electronics}---a laptop with sleeve, a tablet with case, a portable SSD, a
wireless mouse, a USB hub, a foldable keyboard, a phone stand;
\textit{Charging accessories}---laptop / phone / tablet chargers, 2 power banks, 5--6
cables (Type-C, Lightning, USB), a universal travel adapter, a portable power strip, and
a cable-organisation pouch;
\textit{Garments}---2 business shirts, 2 T-shirts, 1 dress trousers / 1 casual trousers,
3 sets of underwear, 4 pairs of socks, 1 set of pyjamas, 1 light jacket, 1 tie;
\textit{Toiletries}---travel toiletry bag, toothbrush case, travel-size toiletry kit,
shaving kit, 1 towel, sample-size skincare;
\textit{Office items}---a document folder with documents, a notebook and markers, a
business-card holder, 2--3 USB drives, a portable stapler;
\textit{Other}---1 pair of leather shoes (with shoe bag), a folding umbrella, a
sunglasses case, a noise-cancelling headphone case, a passport sleeve / ID holder, a
common-medications case, a sleep mask, earplugs, and tissues.

\paragraph{S7: Coffee and beverage preparation (target $\sim$1\,min~15\,s).}
\textit{Instruction:}
``It's afternoon-tea time, please prepare a hot beverage for yourself. There are
instant coffee, milk-tea powder, cocoa powder, and other mixes on the table, plus the
relevant utensils. Choose a beverage and prepare it according to your usual taste,
including selecting an appropriate cup, controlling the water temperature, and adding
sugar / milk to taste.''
\textit{Available props:}
\textit{Mixes}---instant coffee (3 brands), milk-tea powder (original and chocolate),
instant cereal, honey-yuzu tea, tea bags (green / black / floral);
\textit{Utensils}---3--4 mugs of varying sizes, 2 glasses, 1 thermos, an electric
kettle (pre-filled with water), 3 stirring spoons;
\textit{Condiments}---white sugar (sachets and cubes), milk (carton), cream pods.

\paragraph{S8: Clothes hanging and folding (target $\sim$1\,min~10\,s).}
\textit{Instruction:}
``The clothes on the drying rack are dry. Please collect all the garments, fold them
by category (occlusion-prone), and place them into the wardrobe (a shelf is used in
place of a wardrobe; items must be placed on the top shelf only). Pay attention to
folding methods for different garment types and to wardrobe organisation.''
\textit{Available props:}
\textit{Garments on the drying rack}---3--4 T-shirts, 2 shirts, 2 pairs of long
trousers, 2 pairs of shorts, 2--3 towels, 5--6 pairs of socks, 1 jacket, 1 scarf;
\textit{Storage tools}---10--15 hangers, 3 storage baskets, 4 dividers;
\textit{Wardrobe configuration}---a hanging area (with a horizontal bar), a folded-clothes
area (3 shelves), 2 drawers, and a top storage compartment.

\section{Additional Activity-Scenario Recognition Ablations}
\label{sec:app_scene_extra}

Tables~\ref{tab:s_backbone} and~\ref{tab:s_fusion} use the T1 protocol (33-volunteer pool,
volunteer-stratified 5-fold CV, inner-validation model selection) with one seed per fold.

\begin{table}[h]
  \centering
  \caption{Backbone comparison for activity-scenario recognition (macro F1, mean
    $\pm$ SD over fold means; single modality; pool33 5-fold CV, one seed per fold).
    Best mean per modality in bold: the Transformer is best on IMU and EMG, BiLSTM on
    MoCap. Test recordings: IMU 248, MoCap
    247, EMG 259.}
  \label{tab:s_backbone}
  \footnotesize
  \begin{tabular}{lccc}
    \toprule
    Backbone & IMU F1 & MoCap F1 & EMG F1 \\
    \midrule
    CNN1D       & 0.259 $\pm$ 0.118 & 0.395 $\pm$ 0.139 & 0.180 $\pm$ 0.011 \\
    BiLSTM      & 0.286 $\pm$ 0.083 & \textbf{0.449 $\pm$ 0.057} & 0.104 $\pm$ 0.034 \\
    Transformer & \textbf{0.332 $\pm$ 0.093} & 0.426 $\pm$ 0.055 & \textbf{0.202 $\pm$ 0.034} \\
    \bottomrule
  \end{tabular}
\end{table}

\begin{table}[h]
  \centering
  \caption{Fusion strategy comparison for activity-scenario recognition (Transformer
    backbone, IMU + MoCap + EMG; pool33 5-fold CV, one seed per fold, 236
    test recordings; mean $\pm$ SD over fold means). $\Delta$: macro-F1 difference to
    late fusion with the paired $p$ over five folds. Late-family fusion has the highest
    means, but no strategy differs significantly from late fusion.}
  \label{tab:s_fusion}
  \footnotesize
  \setlength{\tabcolsep}{4pt}
  \begin{tabular}{lcccc}
    \toprule
    Fusion Method      & Macro F1       & Acc            & Params & $\Delta$ F1 ($p$) \\
    \midrule
    Early              & 0.337 $\pm$ 0.119 & 0.375 $\pm$ 0.123 & 0.5M & $-0.093$ (0.261) \\
    Late               & 0.430 $\pm$ 0.125 & 0.450 $\pm$ 0.099 & 1.0M & ref. \\
    Attention          & 0.353 $\pm$ 0.045 & 0.394 $\pm$ 0.060 & 1.1M & $-0.076$ (0.340) \\
    Weighted Late      & \textbf{0.438 $\pm$ 0.078} & \textbf{0.460 $\pm$ 0.064} & 1.0M & $+0.008$ (0.791) \\
    Gated Late         & 0.386 $\pm$ 0.038 & 0.410 $\pm$ 0.052 & 1.0M & $-0.044$ (0.545) \\
    Stacking           & 0.369 $\pm$ 0.058 & 0.411 $\pm$ 0.058 & 1.0M & $-0.061$ (0.399) \\
    Product of Experts & 0.400 $\pm$ 0.085 & 0.430 $\pm$ 0.086 & 1.0M & $-0.030$ (0.405) \\
    Mixture of Experts & 0.351 $\pm$ 0.050 & 0.393 $\pm$ 0.054 & 1.0M & $-0.078$ (0.158) \\
    Feature Concat     & 0.351 $\pm$ 0.070 & 0.387 $\pm$ 0.040 & 1.0M & $-0.078$ (0.195) \\
    \bottomrule
  \end{tabular}
\end{table}

\section{Detailed Task Descriptions}
\label{sec:app_tasks}

This appendix expands the benchmark task definitions summarised in Section~\ref{sec:tasks}.

\paragraph{T1: Activity-Scenario Recognition.}
Given a complete multi-modal recording, predict one of the eight scenario labels (S1--S8).
This is a \emph{sequence-level} classification task: a single label is assigned per
recording, requiring models to aggregate evidence across the full duration (0.3--4.0 minutes).
The full variable-length multi-modal sequence is zero-padded and masked at the batch level
(see Appendix~\ref{sec:app_annot_preproc}). Models are evaluated on the input configurations of Table~\ref{tab:pools}, from single modalities up
to all five. Macro-averaged F1 is the primary metric to account
for mild class imbalance across the eight scenarios.

\begin{table}[h]
  \centering
  \caption{Rollup of the 17 T2 fine verbs into the 6-class composite verb head
    (\texttt{verb\_composite}). \emph{support} belongs to the 18-primitive vocabulary
    but is not among the 17 T2 verbs. The six classes follow the contact-state grouping
    of Appendix~\ref{sec:app_annot_preproc}, with \emph{release} kept as its own class.}
  \label{tab:primitive_rollups}
  \footnotesize
  \setlength{\tabcolsep}{4pt}
  \begin{tabular}{@{}p{0.45\linewidth}p{0.40\linewidth}@{}}
    \toprule
    \textbf{Fine verb (17)} & \textbf{Composite (6)} \\
    \midrule
    grasp, pick\_up, hold           & grasp-family \\
    place, put\_down                & place-family \\
    move, pull, push                & transport \\
    adjust, align, stabilize        & adjust \\
    open, close, rotate, insert, remove & state-change \\
    release                         & release \\
    \bottomrule
  \end{tabular}
\end{table}

\paragraph{Evaluation protocol.}
All tasks are evaluated under a unified
\textbf{volunteer-stratified 5-fold cross-validation} protocol
(Section~\ref{sec:stats}) with two volunteer pools. \emph{pool33} contains the 33
volunteers with EMG and skeleton MoCap (T1, T4, T7); its test folds hold 7/7/7/6/6
volunteers, assigned round-robin by volunteer number. \emph{pool27} contains the 27
annotated volunteers with pressure data (T2, T3, T5, T6); its test folds hold 6/6/5/5/5
volunteers from a seed-42 random partition. Every volunteer of a pool appears in the
test set exactly once across the 5 folds, and within-fold training and
test recordings are always disjoint at the volunteer level. From each fold's training
volunteers, $\mathrm{round}(0.15\,n_{\mathrm{train}})$ inner-validation volunteers (4 in
pool33, 3 in pool27) are held out, chosen deterministically and stratified by modality
coverage; epoch selection, early stopping, and learning-rate scheduling use only them,
and each test fold is evaluated once. Main tables use seeds 42, 43, and 44: each fold's
metric is the mean over the seeds, and tables report mean $\pm$ SD (ddof\,=\,1) over the
five fold means (Tables~\ref{tab:s_backbone} and~\ref{tab:s_fusion}: seed 42 only).
Rows are compared on the common test set (per fold, the recordings that every compared
row loaded) with paired $t$-tests over the five fold means and the number of folds in
which a row is better. A configuration uses only recordings that contain all of its
input modalities; Table~\ref{tab:pools} lists the resulting test sets. No subject or
recording identity information is available to models at inference time.

\begin{table}[h]
  \centering
  \caption{Test recordings (summed over the five folds) and volunteers per input
    configuration.}
  \label{tab:pools}
  \footnotesize
  \setlength{\tabcolsep}{4pt}
  \begin{tabular}{@{}l l l r r@{}}
    \toprule
    Pool & Tasks & Input configuration & Test recordings & Volunteers \\
    \midrule
    pool33 & T1, Table~\ref{tab:s_backbone} & EMG                             & 259 & 33 \\
           & T1, Table~\ref{tab:s_backbone} & IMU                             & 248 & 33 \\
           & T1, T7                         & MoCap (T7: EMG + right-hand MoCap) & 247 & 33 \\
           & T1, Table~\ref{tab:s_fusion}   & MoCap + EMG + IMU               & 236 & 33 \\
           & T1, T4                         & MoCap + EMG + Eye + IMU         & 235 & 33 \\
           & T1, T4                         & all five (or gated on pressure) & 192 & 26 \\
    \midrule
    pool27 & T3, T5, T6a                    & EMG (+ pressure)                & 210 & 27 \\
           & T3                             & EyeTrack                        & 209 & 27 \\
           & T2, T3, T6b                    & IMU, MoCap, or EMG + IMU        & 198 & 26 \\
           & T2, T3, T5, T6a                & all five, or MoCap + EMG + Eye + IMU & 192 & 26 \\
    \bottomrule
  \end{tabular}
\end{table}

\noindent T2 rows with other input subsets use 197 (IMU + EMG
+ Eye) or 193 (MoCap + IMU + EMG) test recordings from 26 volunteers.
Baseline code, configuration files, the five-fold split definitions and the
entry point of every table are released alongside the dataset; with the
released data and the fixed seeds they regenerate every reported number.

\section{Fusion Strategies (Detailed Formulations)}
\label{sec:app_fusion}

Multi-modal fusion has a long line of prior work, from simple early/late-fusion baselines
to attention-based methods such as the Multimodal Transformer~\cite{tsai2019multimodal} and
Perceiver IO~\cite{jaegle2022perceiverio}, and gated / mixture-of-experts variants that allow
content-dependent modality weighting. Rather than proposing a new algorithm, we release
\emph{standardised implementations} of nine representative fusion strategies so that the
community can compare them on a single synchronised five-modality dataset under a
controlled protocol.

We implement nine multi-modal fusion strategies that combine the modality-specific
backbones described in Appendix~\ref{sec:app_backbones}.

\textbf{Early fusion} concatenates the raw feature vectors from all modalities along the
feature dimension before feeding into a single shared backbone, forcing the backbone to
learn joint representations from the outset.

\textbf{Late fusion} trains a separate backbone per modality and averages the resulting
class logits at inference time, allowing each modality to be processed independently.

\textbf{Weighted late fusion} extends late fusion with a set of learnable scalar weights
applied to each modality's logit vector before averaging, enabling the model to up-weight
more informative modalities.

\textbf{Gated late fusion} replaces the fixed scalar weights with an input-dependent gating
network: a small MLP takes the concatenated logits as input and produces per-modality
gates, allowing dynamic routing based on the content of each sample.

\textbf{Attention fusion} uses a cross-modal transformer that treats each modality's pooled
representation as a token, applies multi-head self-attention over the modality tokens, and
produces a fused representation via a learned aggregation head.

\textbf{Stacking} trains a shallow MLP meta-learner on top of the concatenated logit
vectors from all per-modality backbones, treating the first-stage outputs as features for a
second classifier.

\textbf{Product of Experts (PoE)} multiplies the softmax probability vectors from each
modality, normalises the result, and takes the argmax. PoE is appropriate when modalities
provide conditionally independent evidence.

\textbf{Mixture of Experts (MoE)} employs top-2 routing: a gating network assigns each
sample to two expert backbones, and the final logits are the weighted sum of the selected
experts' outputs.

\textbf{Feature concatenation} runs a separate backbone per modality, concatenates the
resulting feature vectors (before the classification head), and passes the joint embedding
through a shared MLP classifier.

\section{Backbone Architectures (Detailed)}
\label{sec:app_backbones}

We evaluate three sequence modelling architectures that serve as the building blocks for
both single-modality and multi-modality fusion experiments.

\paragraph{CNN1D.}
A three-layer 1D convolutional network with channel widths 64, 128, and
\texttt{hidden\_dim}, using kernel sizes 7, 5, and 3 respectively. Each convolution is
followed by batch normalisation and ReLU activation. The final representation is obtained
via masked global average pooling, which correctly handles variable-length sequences by
ignoring padded time steps.

\paragraph{BiLSTM.}
A two-layer bidirectional LSTM whose forward and backward hidden states are aggregated via
learned attention pooling. The attention mechanism computes a scalar weight over each
timestep, producing a fixed-dimensional summary vector that is robust to sequence length
variation.

\paragraph{Transformer.}
A linear projection maps raw features to a $d_{\text{model}}=128$ embedding space, to
which sinusoidal positional encodings are added. Two TransformerEncoder layers (number of
heads $=4$, feed-forward dimension $= 4 \times d_{\text{model}}$) process the sequence,
and the final representation is obtained via masked mean pooling over valid (non-padded)
positions.

\section{Training Protocol and Additional Baselines}
\label{sec:app_training}

\paragraph{Training protocol (T1, T4, and Tables~\ref{tab:s_backbone}--\ref{tab:s_fusion}).}
These models are trained with the Adam optimizer (learning rate $10^{-3}$, weight decay
$10^{-4}$) and batch size 16 (4 for ST-GCN and CTR-GCN). The learning rate is reduced on
plateau of the inner-validation loss with a factor of $0.5$ and patience of 7 epochs.
We use cross-entropy loss with class weights set to the inverse of
class frequency in the training set and label smoothing $0.1$. Gradients are clipped to
a maximum $\ell_2$ norm of $1.0$. Training stops early after 15 epochs without
improvement; the epoch is selected on inner-validation macro F1 for T1 (at most 80
epochs) and on the inner-validation loss for T4 and Tables~\ref{tab:s_backbone}
and~\ref{tab:s_fusion} (at most 100 epochs). All models are trained from scratch, except
the pretrained harnet5 encoder in Table~\ref{tab:t1-harnet}. T2 uses the
recipe of Appendix~\ref{subsec:action-results}; T3 and T5--T7 use the recipes below.

\paragraph{Training recipes of T3 and T5--T7.}
Table~\ref{tab:recipes} lists the optimisation settings of the remaining tasks, as recorded in the
configurations of the reported runs. All of them clip gradients at an $\ell_2$ norm of $1.0$ and use only
the inner-validation volunteers for scheduling, early stopping, and epoch selection. T3 uses
cross-entropy with label smoothing $0.1$; T5 uses cross-entropy with inverse-frequency class weights and
at most 10{,}000 training anchors per class; T6 minimises $0.2\,$Huber $+\,0.8\,(1-r)$ on
$\log(1+g)$ targets, averages the weights from epoch 30 on (SWA, learning rate $10^{-4}$, 5 annealing
epochs), and evaluates the final averaged model without checkpoint selection; T7 minimises the masked
MPJPE with the five fingertip joints weighted 2.

\begin{table}[h]
  \centering
  \caption{Training recipes of T3 and T5--T7. Plateau: ReduceLROnPlateau (factor, patience);
    ES: early-stopping patience in epochs. All monitoring and selection use the inner-validation
    volunteers; T7's mean-pose reference is not trained.}
  \label{tab:recipes}
  \footnotesize
  \setlength{\tabcolsep}{3pt}
  \resizebox{\textwidth}{!}{%
  \begin{tabular}{@{}l l c c c c l c l@{}}
    \toprule
    Task & Optimizer & lr & Weight decay & Batch & Max epochs & LR schedule & ES & Epoch selection \\
    \midrule
    T3                & Adam  & $5{\times}10^{-4}$ & $10^{-4}$ & 32 windows   & 50 & plateau (0.5, 5) on inner-val loss & 10 & inner-val ROC AUC \\
    T5 (DAF, FUTR)    & AdamW & $3{\times}10^{-4}$ & $10^{-3}$ & 64 anchors   & 30 & cosine to $0.05\times$ lr & 6 & inner-val macro F1 \\
    T5 (DeepConvLSTM) & AdamW & $10^{-4}$ & $10^{-3}$ & 64 anchors   & 50 & cosine to $0.05\times$ lr & 12 & inner-val macro F1 \\
    T6                & Adam  & $10^{-3}$ & $10^{-4}$ & 8 segments   & 60 & plateau (0.5, 7) on inner-val loss; SWA from epoch 30 & -- & none (final SWA model) \\
    T7                & AdamW & $10^{-3}$ & $10^{-4}$ & 8 recordings & 80 & plateau (0.5, 7) on inner-val MPJPE & 15 & inner-val MPJPE \\
    \bottomrule
  \end{tabular}}
\end{table}

\paragraph{Published baselines.}
To situate PULSE against the broader activity recognition community, we
reproduce published baselines on T1 (Table~\ref{tab:scene-published-ext}), seven
published baseline configurations on the headline action recognition task T2
covering all major design families relevant to sensor-based action recognition
(see Section~\ref{subsec:action-results}), and published regressors on T6 and T7.
Architectures (and architecture hyperparameters) follow each paper's original
recommendation, while optimization follows the protocol of the respective task.

\section{Motion-Primitive Recording Protocol (S9)}
\label{sec:app_primitives}

In addition to the 8 task scenarios (S1--S8), a dedicated \emph{motion-primitive
recording} protocol (denoted S9) was administered to 33 volunteers, yielding
approximately 2.8\,hours of supplementary data. Volunteers performed each primitive in
isolation (several repetitions each), followed by a catalogue of common task
compositions. This isolated primitive library provides a clean, segmentation-friendly
counterpart to the in-context scenarios and supports research on primitive-level
classification, pre-training, and motion-primitive transfer.

\paragraph{Motion primitives.}
\begin{itemize}[leftmargin=1.4em,itemsep=1pt,topsep=2pt]
  \item \textbf{Hand primitives (10):} reach (forward / up / down / lateral), grasp
        (pinch / power / key), release, rotate wrist, press (downward force), pull
        (toward self), push (away from self), slide (lateral translation), pinch
        (two / multi-finger), spread (fingers opening).
  \item \textbf{Arm primitives (6):} lift, lower, extend, bend (elbow flexion), swing,
        circle.
  \item \textbf{Body primitives (7):} bend at the waist, straighten, squat, stand up,
        turn, lean, step.
  \item \textbf{Fine-manipulation primitives (5):} insert, withdraw, fold, tear, wipe
        (reciprocal motion).
\end{itemize}

\paragraph{Basic task compositions.}
Each subject additionally executes a series of elementary task templates that combine
the above primitives:
\begin{itemize}[leftmargin=1.4em,itemsep=1pt,topsep=2pt]
  \item \emph{Pick-and-place:} reach $\to$ grasp $\to$ lift $\to$ move $\to$ lower
        $\to$ release.
  \item \emph{Container operation:} open lid (grasp + rotate / lift), pour (tilt +
        flow-control + reset), scoop (insert + lift + move).
  \item \emph{Open/close:} open drawer (grasp handle + pull), close drawer, press
        switch.
  \item \emph{Organise:} fold garment (unfold + fold + flatten), stack, align, sort
        by category.
  \item \emph{Clean:} wipe surface (grasp cloth + press + reciprocate), sweep and
        collect debris.
  \item \emph{Pack:} tear tape, attach label, wrap and seal a package.
  \item \emph{Write:} pick up pen, write, turn page.
  \item \emph{Assemble / connect:} plug power, connect cable, snap part.
  \item \emph{Measure:} weigh an object, measure a liquid, count items.
  \item \emph{Special:} hang garment, scoop litter, stir beverage, cut with scissors.
\end{itemize}

These primitive and task definitions were used to guide the annotation of the
in-context S1--S8 recordings.

\section{VLM Annotation Prompt Template}
\label{sec:app_prompt}

Below we reproduce the full English translation of the scene-constrained prompt given to
the VLMs during annotation. Scene-specific placeholders
(\texttt{\{scene\_name\}}, \texttt{\{object\_list\_zh\}}, \texttt{\{object\_translation\_table\}})
are substituted per scene before sending to the model. The prompt template texts are released
in the code repository as \path{prompts/vlm_annotation_prompt_s1_s8.txt} (S1--S8) and
\path{prompts/vlm_annotation_prompt_s9.txt} (S9).

\begin{quote}\small
\textbf{Role.} You are an expert in human--computer interaction and action analysis,
specialising in segmenting bimanual object manipulation in egocentric video into atomic
action clips.

\textbf{Objective.} Given the first-person video of a single manipulation scenario,
partition the recording into contiguous $\sim$1--3\,s segments aligned to action
transitions. Only emit segments in which at least one hand is actively interacting with an
object; skip walking, waiting, observing, or idle periods. If two hands perform different
actions, or a single action acts on two objects, emit them as separate segments.

\textbf{Scene context (strict).}
Scene name: \texttt{\{scene\_name\}}.
Allowed objects: \texttt{\{object\_list\_zh\}}.
The value of \texttt{object\_name} must come from the English translations
\texttt{\{object\_translation\_table\}}; unseen objects must be mapped to the closest
allowed entry.

\textbf{Gaze cues.} The video overlays the operator's gaze as a red crosshair ($\oplus$).
Prefer the gaze target when disambiguating the object of interaction.

\textbf{Required fields.} For each segment output a JSON object with exactly two top-level
fields: \texttt{timestamp} (\texttt{HH:MM:SS-HH:MM:SS}, 1\,s resolution) and
\texttt{action\_annotation} with sub-fields \texttt{action\_name} (one of:
\textit{grasp, hold, move, push, pull, rotate, open, close, insert, remove, place,
pick\_up, put\_down, adjust, align, release, support, stabilize}),
\texttt{hand\_type} (\textit{left}/\textit{right}/\textit{both}),
\texttt{is\_lhand}, \texttt{is\_rhand}, \texttt{text\_key}
(``\textit{\{Action\} \{object\} with \{hand\} hand(s).}''),
\texttt{text\_variants} (exactly 4 paraphrases preserving action, object, hand),
and \texttt{object\_name} (English name from the whitelist).

\textbf{Output.} Return a single JSON object \texttt{\{"segments": [...]\}}
with no Markdown, code fences, or surrounding text.
\end{quote}

\section{Human Review of Annotations}
\label{sec:app_human_review}
\label{sec:app_human_validation}

Before release, the VLM annotations (Appendix~\ref{sec:app_annot_preproc}) were
reviewed by a human. Following a written protocol, each recording was checked segment
by segment against its first-person video with gaze overlay, and every segment was
confirmed or corrected, editing the primitive, hand, object, or boundaries and adding
or deleting segments where needed. During review, nine
objects missing from the per-scenario lists were added to the object vocabulary; the
18-primitive vocabulary was not changed.

Recording v40/s3,
whose first-person view mostly misses the hands, was instead re-annotated by the
authors from third-person video and motion-capture hand trajectories. Recording v18/s2
carries no action labels because its first-person video froze; its sensor streams are
released. The released annotations therefore comprise 7{,}356 segments in 282
recordings, all of which have been confirmed or corrected by a human. Segments cover
about 73\% of the annotated recording time; intervals in which the hands or the
manipulated object are out of view are left unlabeled.

\section{Getting Started: Code Example}
\label{sec:app_getting_started}

Below is a minimal code example showing how to load and iterate over the dataset
using the data layer of the companion code (the \texttt{pulse} package), run from the root of a
checkout of the code repository.

\begingroup\footnotesize
\begin{verbatim}
# git clone https://huggingface.co/velvet-pine-22/PULSE-code && cd PULSE-code
# conda env create -f environment.yml && conda activate pulse
from pulse.paths import DataRoot
from pulse.splits import load_split
from pulse.data import load_recording
from pulse.data.annotations import load_segments
from pulse.data.availability import Availability
from pulse.data.manifest import expected_manifest

dr   = DataRoot.from_arg("/path/to/PULSE")         # or set $PULSE_ROOT
fold = load_split("splits/pool27.json").fold(0)     # volunteer-disjoint 5-fold CV
av   = Availability.load("manifests/availability_v1.1.json")
mods = ["mocap", "emg", "eyetrack"]
man  = expected_manifest(av, fold, mods, require_annotation=True)
print(man.summary())          # train 133 / inner_val 24 / test 40 recordings

rid = man.partitions["train"][0]                    # "v1s2"
rec = load_recording(dr, rid, mods, align_trimmed=True)
mocap, emg, gaze = (rec.arrays[m] for m in mods)    # (T,620) (T,8) (T,24), 100 Hz
for s in load_segments(dr, rid):                    # annotations/v{id}/s{id}.json
    print(s.start_sec, s.end_sec, s.verb, s.noun, s.hand)

# every benchmark run:
#   python -m pulse.tasks.<task> --root $PULSE_ROOT --split splits/<pool>.json \
#       --fold <k> --seed <s> --config configs/<...>.yaml
\end{verbatim}
\endgroup

\noindent
\texttt{scripts/run\_table.py} and \texttt{scripts/aggregate.py} regenerate whole tables (see the
repository README). Users who need only the sensor streams can download the release without the
\texttt{videos/} folders (22.0\,GB: 14.0\,GB of per-modality CSV files, 8.0\,GB of skeleton TSV files,
plus annotations and metadata), which suits small-compute environments.

\section{Action Recognition (T2): Full Per-Head Results}
\label{sec:app_action_full}

Table~\ref{tab:action-recognition-full} reports per-head top-1 accuracy, the joint
metrics, and the two non-headline heads (verb\_composite, 6 classes; hand, 3 classes)
on the same sliding-window runs as Table~\ref{tab:action-recognition}. The last row is
the accuracy of always predicting the most frequent class of the training folds
(training majority), evaluated on DAF's test windows.

\begin{table}[h]
  \centering
  \caption{Action recognition (T2): full per-head metrics, sliding-window protocol
    (pool27, 5 folds $\times$ 3 seeds, mean $\pm$ SD over fold means). The first six
    columns expand the three headline metrics of
    Table~\ref{tab:action-recognition} by reporting top-1 alongside top-5 for each;
    the last two columns are top-1 accuracy of the verb-composite and hand heads.}
  \label{tab:action-recognition-full}
  \footnotesize
  \setlength{\tabcolsep}{3pt}
  \resizebox{\textwidth}{!}{%
  \begin{tabular}{@{}lcccccccc@{}}
    \toprule
    Method & verb\_fine@1 & verb\_fine@5 & noun@1 & noun@5 & act\_vn@1 & act\_vn@5 & verb\_comp@1 & hand@1 \\
    \midrule
    DailyActFormer (ours)  & 0.292 $\pm$ 0.024 & 0.762 $\pm$ 0.013 & 0.261 $\pm$ 0.030 & 0.597 $\pm$ 0.034 & 0.097 $\pm$ 0.011 & 0.468 $\pm$ 0.026 & 0.388 $\pm$ 0.029 & 0.655 $\pm$ 0.024 \\
    HandFormer             & 0.289 $\pm$ 0.024 & 0.755 $\pm$ 0.023 & 0.261 $\pm$ 0.049 & 0.582 $\pm$ 0.072 & 0.095 $\pm$ 0.016 & 0.463 $\pm$ 0.057 & 0.405 $\pm$ 0.030 & 0.674 $\pm$ 0.031 \\
    AFFT                   & 0.279 $\pm$ 0.021 & 0.734 $\pm$ 0.015 & 0.233 $\pm$ 0.029 & 0.553 $\pm$ 0.037 & 0.083 $\pm$ 0.008 & 0.430 $\pm$ 0.017 & 0.379 $\pm$ 0.027 & 0.620 $\pm$ 0.040 \\
    FUTR                   & 0.268 $\pm$ 0.027 & 0.734 $\pm$ 0.026 & 0.239 $\pm$ 0.048 & 0.550 $\pm$ 0.070 & 0.081 $\pm$ 0.017 & 0.423 $\pm$ 0.048 & 0.365 $\pm$ 0.037 & 0.624 $\pm$ 0.035 \\
    DeepConvLSTM (3 mod)   & 0.269 $\pm$ 0.025 & 0.731 $\pm$ 0.029 & 0.218 $\pm$ 0.035 & 0.520 $\pm$ 0.069 & 0.084 $\pm$ 0.015 & 0.412 $\pm$ 0.052 & 0.380 $\pm$ 0.038 & 0.617 $\pm$ 0.030 \\
    RU-LSTM                & 0.265 $\pm$ 0.019 & 0.730 $\pm$ 0.019 & 0.219 $\pm$ 0.024 & 0.526 $\pm$ 0.055 & 0.075 $\pm$ 0.014 & 0.408 $\pm$ 0.033 & 0.370 $\pm$ 0.028 & 0.617 $\pm$ 0.057 \\
    ActionLLM (surrogate)  & 0.222 $\pm$ 0.014 & 0.701 $\pm$ 0.016 & 0.212 $\pm$ 0.026 & 0.501 $\pm$ 0.034 & 0.066 $\pm$ 0.005 & 0.366 $\pm$ 0.022 & 0.314 $\pm$ 0.021 & 0.583 $\pm$ 0.050 \\
    DeepConvLSTM (IMU)     & 0.212 $\pm$ 0.025 & 0.701 $\pm$ 0.018 & 0.209 $\pm$ 0.031 & 0.476 $\pm$ 0.044 & 0.064 $\pm$ 0.010 & 0.343 $\pm$ 0.034 & 0.304 $\pm$ 0.038 & 0.589 $\pm$ 0.070 \\
    \midrule
    Training majority      & 0.167 $\pm$ 0.018 & -- & 0.175 $\pm$ 0.035 & -- & -- & -- & 0.237 $\pm$ 0.026 & 0.597 $\pm$ 0.063 \\
    \bottomrule
  \end{tabular}%
  }
\end{table}

\paragraph{Notes.}
DAF has the highest or tied-highest mean in all six columns that expand the headline
metrics (noun@1 is tied with HandFormer). It does not lead on the two non-headline
heads: HandFormer has higher verb\_comp@1 and hand@1. Verb-composite accuracy is
clearly above the training majority for every model, while hand accuracy is close to it
(and slightly below it for the two models without MoCap). The joint $\mathrm{act\_vn}$ chance
levels are
$\mathrm{act\_vn@1}\!\approx\!1/561\!\approx\!0.18\%$ and
$\mathrm{act\_vn@5}\!\approx\!25/561\!\approx\!4.5\%$, so the absolute
joint values reflect both the difficulty of the 17$\times$33 joint label
space and the small per-fold training-pool size; per-head metrics on
verb\_fine and noun are correspondingly higher.

\section{T5 (TGSR): Per-backbone Cross-Validation Detail}
\label{sec:app_t5_cv}

The T5 tables of Appendix~\ref{sec:app_t5_tactile} (Tables~\ref{tab:tactile-online},
\ref{tab:tactile-priv}, \ref{tab:tactile-by-event}) report 5-fold
volunteer-stratified CV results aggregated across the three backbones.
This appendix expands the same runs to a per-backbone view.
The 27 volunteers of pool27 are partitioned (seed 42) into test folds of 6/6/5/5/5
volunteers; for fold $k$, that fold's volunteers are the test set, three of the
remaining volunteers are the inner-validation set, and the rest are the training set.
Each (backbone, condition) cell is trained with 3 seeds per fold
(3 backbones $\times$ 2 conditions $\times$ 5 folds $\times$ 3 seeds $= 90$ runs, plus
15 pressure-only DAF reference runs).
DeepConvLSTM uses $\mathit{lr}=10^{-4}$, at most 50 epochs, and early-stopping patience 12;
all other hyperparameters match the T5 recipe of Table~\ref{tab:recipes}.

\begin{table}[h]
  \centering
  \caption{T5 (TGSR) per-backbone view of the headline 5-fold CV (3 seeds per fold).
    Mean $\pm$ SD over the 5 fold means per backbone; the aggregate row averages the
    three backbones within each fold. The SustainedGrasp F1 lift is positive in every
    fold for every backbone, and the aggregate lift matches
    Table~\ref{tab:tactile-priv}.}
  \label{tab:tactile-cv}
  \footnotesize
  \setlength{\tabcolsep}{4pt}
  \resizebox{\textwidth}{!}{%
  \begin{tabular}{@{}lcccccc@{}}
    \toprule
    Backbone & macro F1 (EMG) & macro F1 (EMG+Pres) & macro lift &
    sust F1 (EMG) & sust F1 (EMG+Pres) & \textbf{sust lift} \\
    \midrule
    DAF              & 0.383 $\pm$ 0.024 & 0.473 $\pm$ 0.010 & $+0.090 \pm 0.024$ & 0.325 $\pm$ 0.060 & 0.516 $\pm$ 0.046 & $+0.192 \pm 0.015$ \\
    FUTR             & 0.382 $\pm$ 0.028 & 0.474 $\pm$ 0.011 & $+0.092 \pm 0.026$ & 0.335 $\pm$ 0.073 & 0.519 $\pm$ 0.046 & $+0.184 \pm 0.040$ \\
    DeepConvLSTM     & 0.374 $\pm$ 0.013 & 0.448 $\pm$ 0.021 & $+0.074 \pm 0.029$ & 0.304 $\pm$ 0.050 & 0.510 $\pm$ 0.050 & $+0.206 \pm 0.024$ \\
    \midrule
    3-backbone mean  & 0.380 $\pm$ 0.021 & 0.465 $\pm$ 0.012 & $+0.086 \pm 0.024$ & 0.321 $\pm$ 0.059 & 0.515 $\pm$ 0.047 & {\boldmath$+0.194 \pm 0.018$} \\
    \bottomrule
  \end{tabular}%
  }
\end{table}

DAF pressure-only reference across the 5 folds: macro F1
0.459 $\pm$ 0.018 (difference to DAF with EMG + pressure
$-0.014 \pm 0.012$, $p =$~0.066),
SustainedGrasp F1 0.519 $\pm$ 0.037 (difference
$+0.002 \pm 0.024$,
$p =$~0.853). Pressure alone thus performs about
as well as EMG plus pressure, so the lift is carried by the pressure channel rather than
by EMG.

Per-event SustainedGrasp F1 lift (Table~\ref{tab:tactile-by-event}) is largest in
\emph{steady-grip} ($+0.161 \pm 0.039$ within-event F1)
and smaller in \emph{pre-contact} ($+0.062 \pm 0.048$),
where it also varies across backbones (DAF +0.057,
FUTR +0.007, DeepConvLSTM
+0.123); the non-contact and release
events have no SustainedGrasp samples by construction.

\paragraph{Takeaway.} Across 27 volunteers, 5 fold rotations, and 3 seeds, the
SustainedGrasp F1 lift is $+0.194 \pm 0.018$
($p =$~$1.6\times 10^{-5}$ over the fold means), with per-backbone lifts
of +0.192, +0.184, and
+0.206; the magnitude is similar across
backbones. Pressure encodes SustainedGrasp information unavailable to forearm EMG, and
the conclusion is robust to test-volunteer choice.

\section{Compute Resources}
\label{sec:app_compute}

All benchmark experiments were run on a shared university cluster equipped with
NVIDIA A800 80\,GB GPUs, one GPU per run. According to the cluster accounting, the
1{,}515 GPU runs behind the reported T1--T7 tables (including the appendix ablations
and the matched-input comparisons) took \textbf{about 53 GPU-hours} in total; a single
run took a median of about 2 minutes and at most 22 minutes. GPU smoke tests before
the main runs took about 2 further GPU-hours, and the T6 seed ensembles and the F0 case
study ran on CPU. These figures exclude the F5 experiments and earlier exploratory
runs. Multi-fold and multi-seed experiments scale approximately linearly with the
number of folds and seeds.

\end{document}